%% file: main.tex
\documentclass[runningheads]{llncs}

\usepackage{eccv}

\usepackage{eccvabbrv}

\usepackage{graphicx}
\usepackage{booktabs}
\usepackage{algorithm}
\usepackage{algpseudocode}

\usepackage[accsupp]{axessibility}  
\usepackage{multirow}
\usepackage{bm}
\usepackage{pifont}
\usepackage{wrapfig}
\usepackage{placeins}
\usepackage{relsize}
\usepackage{setspace}
\usepackage{booktabs}
\usepackage{makecell}
\usepackage{bigstrut}
\usepackage{subcaption}
\usepackage[table]{xcolor}
\usepackage[utf8]{inputenc}
\usepackage{fontawesome5}
\usepackage{xcolor}
\usepackage{caption}
\usepackage{graphicx}
\usepackage{amssymb}
\usepackage{setspace}
\usepackage{array}
\usepackage{amsmath, amssymb}
\usepackage{threeparttable}
\usepackage[font=small,labelfont=bf,labelsep=space]{caption}
\usepackage[font=scriptsize,labelfont=bf]{subcaption}
\newcolumntype{M}{>{$}p{4.5cm}<{$}}

\newcommand{\fst}[1]{\cellcolor{red!25}\textbf{#1}}
\newcommand{\snd}[1]{\cellcolor{orange!25}#1}
\newcommand{\trd}[1]{\cellcolor{yellow!15}#1}

\newcommand{\red}[1]{{\color{red}#1}}

\newcommand{\newpara}[1]{\vspace{2pt}\noindent\textbf{#1}}
\newcommand{\cmark}{\textcolor{Green}{\checkmark}}
\newcommand{\xmark}{\textcolor{Red}{\ding{55}}}

\definecolor{blue}{rgb}{0.22, 0.22, 0.95}
\definecolor{bblue}{rgb}{0.12, 0.43, 0.84}
\definecolor{gray}{rgb}{0.29, 0.31, 0.31}
\definecolor{green}{rgb}{0.22, 0.7, 0.22}
\definecolor{lred}{rgb}{0.85, 0.27, 0.08}
\definecolor{tred}{rgb}{0.459,0.184,0.063}
\definecolor{orange}{rgb}{1.0, 0.4, 0}
\definecolor{headerblue}{RGB}{30, 80, 160}
\definecolor{rowgray}{RGB}{245, 247, 250}

\usepackage[breaklinks,colorlinks,citecolor=eccvblue]{hyperref}

\usepackage{orcidlink}

\begin{document}

\title{GeoNVS: Geometry Grounded Video Diffusion for Novel View Synthesis} 
\setstretch{0.97}

\titlerunning{GeoNVS: Geometry Grounded Video Diffusion for Novel View Synthesis}

\author{Minjun Kang\inst{1}\orcidlink{0000-0003-3102-7591} \and
Inkyu Shin\inst{2}\orcidlink{0009-0007-4314-9170} \and
Taeyeop Lee\inst{1}\orcidlink{0000-0002-3574-9172} \and
Myungchul Kim\inst{1}\orcidlink{0000-0003-2959-7405} \and \\
In So Kweon\inst{1}\orcidlink{0000-0001-9626-5983} \and
Kuk-Jin Yoon\inst{1}\orcidlink{0000-0002-1634-2756}}

\authorrunning{Kang et al.}

\institute{$^1$KAIST, South Korea \quad $^2$Luma AI, USA}

\maketitlewithvisual{We demonstrate \textbf{GeoNVS}, integrated into SEVA~\cite{zhou2025stable} and CameraCtrl~\cite{he2025cameractrl}, consistently enhancing geometric fidelity and camera controllability, while resolving photometric inconsistencies (\textbf{right-top}) across challenging large-scale indoor and outdoor scenes. Our approach consistently outperforms existing methods. \href{https://sites.google.com/view/minjun-kang/geonvs-eccv26}{\red{Project page}}.}
\vspace{-3mm}

\input{sec/0_abstract}
\input{sec/1_intro}

\vspace{-4mm}
\input{sec/2_relatedwork}
\vspace{-2mm}
\input{sec/3_method}
\input{sec/4_experiment}
\input{sec/5_conclusion}

\appendix
\input{Supplementary_Material/6_supple}

\clearpage  


%
%
\bibliographystyle{splncs04}
\bibliography{main}
\end{document}

%% file: sec/0_abstract.tex
\begin{abstract}
  Novel view synthesis requires strong 3D geometric consistency and the ability to generate visually coherent images across diverse viewpoints. While recent camera-controlled video diffusion models show promising results, they often suffer from geometric distortions and limited camera controllability. To overcome these challenges, we introduce \textbf{GeoNVS}, a geometry-grounded novel-view synthesizer that enhances both geometric fidelity and camera controllability through explicit 3D geometric guidance. Our key innovation is the Gaussian Splatting Feature Adapter (GS-Adapter), which lifts input-view diffusion features into 3D Gaussian representations, renders geometry-constrained novel-view features, and adaptively fuses them with diffusion features to correct geometrically inconsistent representations. Unlike prior methods that inject geometry at the input level, GS-Adapter operates in feature space, avoiding view-dependent color noise that degrades structural consistency. Its plug-and-play design enables zero-shot compatibility with diverse feed-forward geometry models without additional training, and can be adapted to other video diffusion backbones. Experiments across 9 scenes and 18 settings demonstrate state-of-the-art performance, achieving 11.3\% and 14.9\% improvements over SEVA and CameraCtrl, with up to $2\times$ reduction in translation error and $7\times$ in Chamfer Distance.
\end{abstract}

%% file: sec/1_intro.tex
\section{Introduction}
\label{sec:intro}
\vspace{-3mm}
Novel View Synthesis (NVS) aims to synthesize photorealistic images from novel viewpoints while preserving the geometric coherence of the observed scene. Neural Radiance Fields (NeRF)~\cite{mildenhall2021nerf} and 3D Gaussian Splatting (3D-GS)~\cite{kerbl3Dgaussians} are representative per-scene optimization solutions that require a large number of observations for training. These methods struggle with sparse-view inputs and are limited to novel views near input viewpoints. Recently, feed-forward geometry methods~\cite{charatan23pixelsplat,chen2024mvsplat,xu2025depthsplat,tang2024hisplat,wang2025vggt,wang2025pi3,lin2025depth} and generative approaches~\cite{gao2024cat3d,zhou2025stable,he2025cameractrl,he2025cameractrl2,wang2024motionctrl,zhang2025tora} have emerged as strong alternatives for such challenging scenarios.

\textbf{Feed-forward geometry} methods directly estimate explicit 3D representations from sparse-view input images in a single forward pass. One line of work~\cite{chen2024mvsplat,xu2025depthsplat,tang2024hisplat,lin2025depth} predicts 3D Gaussians directly from inputs and renders novel views via Gaussian splatting. Another line of work~\cite{wang2025vggt,wang2025pi3,keetha2026mapanything} first estimates point cloud maps and then requires optimization to fit 3D Gaussians~\cite{fan2024instantsplat}.
Despite their strong structural consistency in observed regions, these methods often fail to preserve fine-grained appearance details in low-overlap scenes.

\textbf{Generative novel-view synthesis} represents another powerful paradigm for handling sparse-view inputs, enabling dense and semantically rich scene synthesis. These methods incorporate camera~\cite{wang2024motionctrl,he2025cameractrl,zhou2025stable,gao2024cat3d,cao2025mvgenmaster,bai2025recammaster} or point trajectories~\cite{yin2023dragnuwa,wu2024draganything,geng2025motion} into video diffusion models to guide novel view generation. They exhibit remarkable capability even for scenes with low-overlap inputs, overcoming limitations of feed-forward geometry methods. However, these approaches still suffer from multi-view inconsistency, hallucinating structures that deviate from the input images, and limited camera controllability, failing to follow the intended camera trajectories as shown in~\cref{fig:teaser}. 
A natural remedy is to incorporate explicit geometry priors from feed-forward geometry models to enforce structural consistency between generated and input views.

Indeed, several studies~\cite{wu2024reconfusion,chan2023generative,liu20243dgs,chen2024mvsplat360,wu2025genfusion,wu2025difix3d+,yin2025gsfixer,ren2025gen3c} have explored this direction, combining geometry priors with generative models to achieve geometrically consistent and dense novel-view synthesis. These methods feed noisy rendered novel-view images~\cite{liu20243dgs,wu2025genfusion,wu2025difix3d+,yin2025gsfixer,ren2025gen3c} or features~\cite{wu2024reconfusion,chan2023generative,chen2024mvsplat360} into a diffusion model for refinement. However, we observe that these input-level fusion methods in~\cref{fig:model_overview}-(b) introduce geometric distortions and hallucinated structures in the final output, as noisy textures in the rendered images obscure structural information.

To overcome this limitation, we propose \textbf{GeoNVS}, which couples geometry priors with the diffusion model in feature space, enabling the diffusion model to leverage explicit 3D structural information during the denoising process. Our core component, the Gaussian Splatting Feature Adapter (\textbf{GS-Adapter}), bridges 3D Gaussians and the diffusion model through a feature embedding and rendering process, modulating geometrically inconsistent diffusion features. Specifically, GS-Adapter (1) embeds 2D input-view diffusion features into 3D Gaussians, (2) rasterizes geometry-aware novel-view features via Gaussian splatting, and (3) adaptively fuses them with novel-view diffusion features to produce geometrically corrected features. By anchoring diffusion features to explicit 3D structure, GS-Adapter ensures that generated pixels are geometrically aligned with the target camera viewpoint, leading to improved camera controllability.

\input{figure/model_overview}
Compared to previous methods that perform input-level fusion in~\cref{fig:model_overview}-(b), our approach in~\cref{fig:model_overview}-(c) leverages only the structural information encoded in 3D Gaussians, avoiding reliance on noisy view-dependent color information. Our modular design enables \textit{\textbf{plug-and-play}} integration with diverse feed-forward geometry models without additional training, while existing methods~\cite{chen2024mvsplat360,yin2025gsfixer,wu2025geometryforcing,li2025nvcomposer} depend on a specific geometry model, requiring retraining upon substitution.
To validate the effectiveness of \textbf{GeoNVS}, we integrate GS-Adapter into SEVA~\cite{zhou2025stable} and CameraCtrl~\cite{he2025cameractrl} and evaluate across indoor and outdoor scenes in both low- and large-overlap settings as well as long-trajectory generation, demonstrating consistent improvements in geometric consistency, camera controllability, and photorealistic quality.
In summary, our key contributions are as follows:
\begin{itemize}
\item We propose \textbf{GS-Adapter}, which couples 3D Gaussian priors with the diffusion model in feature space, improving geometric consistency and camera controllability in video diffusion-based novel-view synthesis.
\item Our modular design supports diverse combinations of feed-forward geometry models and video diffusion backbones without additional training, enabling flexible plug-and-play applicability.
\item \textbf{GeoNVS} achieves state-of-the-art performance over existing generative NVS methods across diverse indoor and outdoor benchmarks, evaluated on 9 scenes and 18 settings.
\end{itemize}

%% file: figure/model_overview.tex
\begin{figure}[t]
\centering
\includegraphics[width=0.99\linewidth]{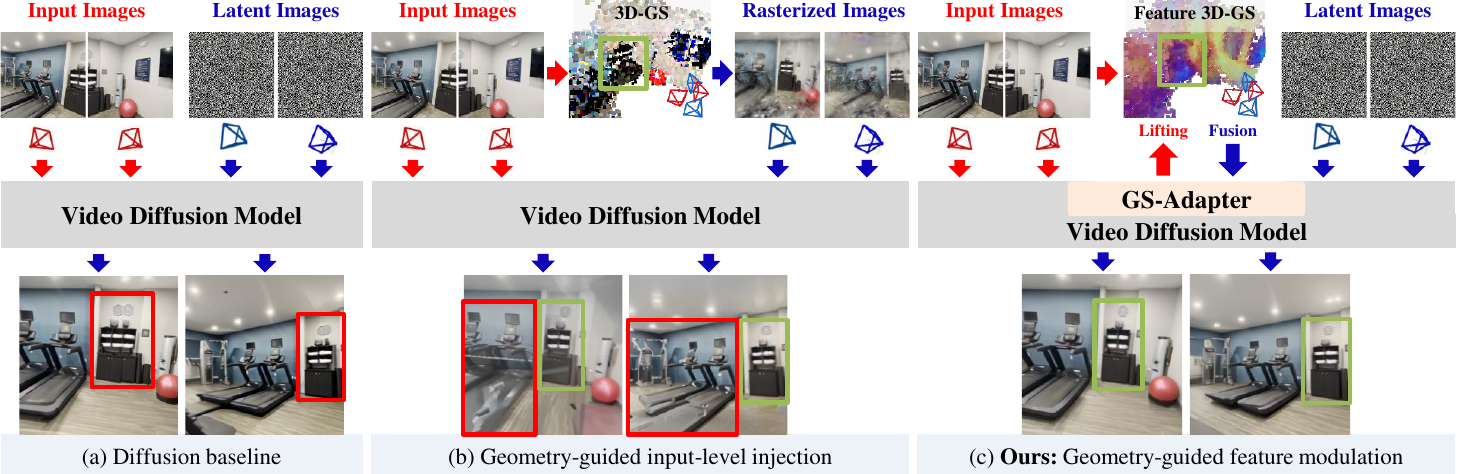}
\vspace{-2mm}
\caption{
\textbf{Geometry-guided generative NVS.} (a) Pure diffusion model produces view-inconsistent results. Given sparse \textcolor{red}{input-view} images, both (b) and (c) reconstruct 3D Gaussians from input views using a geometry prior. (b) Previous methods inject rasterized \textcolor{blue}{novel-view} images from 3D-GS as input, causing artifacts from noisy rasterized colors. (c) Our method modulates internal diffusion features via a GS-Adapter conditioned on 3D-GS, achieving superior geometry consistency and visual quality.}
\vspace{-6mm}
\label{fig:model_overview}
\end{figure}

%% file: sec/2_relatedwork.tex
\section{Related Work}
\label{sec:related_work}
\vspace{-1mm}
\noindent\textbf{Sparse-view Novel View Synthesis.}
NeRF~\cite{mildenhall2021nerf} and 3D-GS~\cite{kerbl3Dgaussians} achieve photorealistic novel-view synthesis given dense inputs, but overfit severely under sparse views. To resolve this, prior works introduce depth supervision~\cite{deng2022depth,wang2023sparsenerf,zhu2024fsgs,chung2024depth} or normal consistency~\cite{yu2022monosdf,seo2023flipnerf}, yet remain sensitive to constraint quality. Feed-forward geometry methods~\cite{charatan23pixelsplat,chen2024mvsplat,xu2025depthsplat,tang2024hisplat} directly predict 3D Gaussians using cost volumes~\cite{yao2018mvsnet,im2019dpsnet,gu2020cascade,duzceker2021deepvideomvs} or monocular depth~\cite{eftekhar2021omnidata,yang2024depth,yin2023metric3d}, and geometry foundation models~\cite{wang2025vggt,wang2025pi3,keetha2026mapanything,lin2025depth} further enhance reconstruction quality at scale. We leverage these priors to guide the diffusion model toward structural consistency.

\newpara{Generative Novel View Synthesis.}
Diffusion models have evolved into powerful task executors by incorporating controllable conditioning signals~\cite{zhang2023adding,li2024controlnet++}. Camera-controlled video diffusion methods embed camera poses~\cite{wang2024motionctrl}, Plücker coordinates~\cite{he2025cameractrl,he2025cameractrl2,zhou2025stable,gao2024cat3d,cao2025mvgenmaster,bai2025recammaster}, trajectory-based attention~\cite{zhang2025tora}, or projected 3D point maps~\cite{gu2025diffusion} to guide video generation. Recently, SEVA~\cite{zhou2025stable} demonstrated strong novel-view synthesis via a multi-view diffusion model trained on large-scale datasets. Building on SEVA and CameraCtrl, we enhance these baselines to achieve geometrically consistent generation and improved controllability.

\newpara{Feature field distillation for 3D-GS.}
Beyond radiance fields, embedding semantic features into 3D representations~\cite{kerr2023lerf,qin2024langsplat} has enabled various downstream applications~\cite{ji2024graspsplats,liang2024supergseg,huang2026openvoxel,lee2025mosaic3d,wang2025inpaint360gs}. Most existing methods~\cite{zhou2024feature,qin2024langsplat,li2025langsplatv2} distill semantic features~\cite{radford2021learning,kirillov2023segment,caron2021emerging,oquab2023dinov2} into 3D Gaussians via per-scene optimization. Recently, Dr.Splat~\cite{drsplat25} introduced feed-forward feature distillation into 3D Gaussians, enabling fast 3D localization without per-scene optimization. LUDVIG~\cite{marrie2025ludvig} extends this to integrate DINOv2~\cite{oquab2023dinov2} features into 3D scene geometry through graph diffusion for semantic segmentation.
Unlike prior methods that distill static semantic features, diffusion features vary across denoising timesteps, making per-scene optimization infeasible. We therefore adopt a direct feed-forward lifting approach, extending it to preserve pixel-level feature fidelity rather than coarse instance-level distinctions sufficient for segmentation.

%% file: sec/3_method.tex
\section{Method}
\label{sec:method}
\vspace{-2mm}
In this section, we first provide preliminaries on the problem formulation, geometry prior construction, and Gaussian splatting background in~\cref{subsec:overview}. We then introduce \textbf{GS-Adapter} in~\cref{subsec:gs_adapter}, our core module that leverages 3D Gaussian priors to correct geometrically inconsistent diffusion features, thereby enforcing geometric consistency and improving camera controllability. Finally, we describe the integration of GS-Adapter into an existing video diffusion framework via multi-scale feature aggregation in~\cref{subsec:geonvs}.

\vspace{-2mm}
\subsection{Preliminary}
\label{subsec:overview}
\input{figure/model_detail}
\noindent\textbf{Problem Formulation.}
Given $P$ reference images $\mathbf{I}_{\text{ref}} \in \mathbb{R}^{P \times hw \times 3}$ and their corresponding camera poses $\boldsymbol{\pi}^{\text{ref}}$, the NVS task is to synthesize $Q$ target views $\mathbf{I}_{\text{tgt}} \in \mathbb{R}^{Q \times hw \times 3}$ for the target camera poses $\boldsymbol{\pi}^{\text{tgt}}$. \textbf{GeoNVS} supports an arbitrary number of reference images, as both SEVA~\cite{zhou2025stable} and the geometry models natively handle monocular and multi-view inputs.

\newpara{Geometry Prior.} 
Our method requires 3D Gaussians $\mathcal{G}$ as structural guidance for the diffusion model. We utilize feed-forward geometry models~\cite{chen2024mvsplat,xu2025depthsplat,tang2024hisplat,wang2025vggt,wang2025pi3} to estimate this geometry prior from the reference images. 
For geometry models that estimate both camera poses and point clouds 
(VGGT~\cite{wang2025vggt} and Pi3~\cite{wang2025pi3}), we align the 
predicted camera scale $s^{*}$ to the target cameras by solving:
\begin{equation}
    s^{*} = \arg\min_{s} \left\| \mathbf{C}^{\text{gt}} - s\, \mathbf{C}^{\text{pred}} \right\|_{2}^{2}
    \label{eq:scale_optimize}
\end{equation}
where $\mathbf{C}^{\text{gt}}$ and $\mathbf{C}^{\text{pred}}$ denote 
the target and predicted camera centers, respectively. 
We then fit 3D Gaussians initialized from the estimated point clouds~\cite{fan2024instantsplat}.

\newpara{3D Gaussian Splatting.} 
3D Gaussian Splatting~\cite{kerbl3Dgaussians} represents a scene with $N$ Gaussian particles composed of position $\boldsymbol{\mu} \in \mathbb{R}^3$, rotation $\boldsymbol{r} \in \mathbb{R}^4$ via quaternions, scale $\boldsymbol{s} \in \mathbb{R}^3$, opacity $\boldsymbol{\sigma} \in \mathbb{R}^1$, and color represented by spherical harmonics. The color at pixel $p$ for view $d$ is obtained by summing contributions of Gaussians:
\vspace{-3mm}
\begin{equation}
C(d,p) = \sum_{i=1}^{N} c_i(d)\, w_i(d,p), \quad w_i(d,p) = \alpha_i(d,p) \prod_{j=1}^{i-1} (1 - \alpha_j(d,p))
\vspace{-3mm}
\label{eq:rendering_color}
\end{equation}
where $\alpha_i(d,p)$ is the product of the opacity $\sigma_i$ and the 2D projected Gaussian density $\mathcal{N}_i$ at pixel position $p$. Similarly, feature rasterization is represented as:
\vspace{-3mm}
\begin{equation}
\mathbf{F}(d,p) = \sum_{i=1}^{N} \mathrm{f}_i\, w_i(d,p)
\vspace{-4mm}
\label{eq:rendering_feat}
\end{equation}

\subsection{GS-Adapter: Geometry-Guided Diffusion Feature Adaptation}
\label{subsec:gs_adapter}
Given the diffusion features $\mathbf{F}^{t}$ at denoising timestep $t$ and 3D Gaussians prior $\mathcal{G}$, GS-Adapter is designed to obtain geometry-augmented features $\hat{\mathbf{F}}^{t}$, formulated as $\hat{\mathbf{F}}^{t} = \mathrm{\Phi}(\mathbf{F}^{t}, \mathcal{G})$. Diffusion features $\mathbf{F}^{t} = [\mathbf{F}_{\text{ref}}^{t};\mathbf{F}_{\text{tar}}^{t}]$ consist of reference-view (input-view) features $\mathbf{F}_{\text{ref}}^{t} \in \mathbb{R}^{P \times \frac{h}{n} \times \frac{w}{n} \times \text{C}}$ and novel-view features $\mathbf{F}_{\text{tar}}^{t} \in \mathbb{R}^{Q \times \frac{h}{n} \times \frac{w}{n} \times \text{C}}$, where $n$ denotes the downsampling factor of the Variational Autoencoder (VAE), and $\text{C}$ denotes the channel dimension of the diffusion features. For convenience, we omit the superscript $t$ hereafter. As illustrated in~\cref{fig:detail_geonvs}-(b),  GS-Adapter consists of three major steps: (1) \textbf{feature lifting:} projecting reference-view diffusion features onto 3D Gaussians, (2) \textbf{feature refinement:} refining the rendered novel-view features to recover fine-grained details, and (3) \textbf{feature fusion:} combining the geometry-guided features with the original diffusion features to obtain the revised diffusion features.

\newpara{Uplifting diffusion feature into 3D-GS.} 
Since the diffusion features vary across timesteps and our method requires obtaining geometry-augmented features at every denoising timestep, per-scene optimization-based feature embedding approaches~\cite{qin2024langsplat,li2025langsplatv2,zhou2024feature} are not feasible, as they cannot adapt to the continuously changing feature distribution during denoising.
Inspired by~\cite{drsplat25,marrie2025ludvig}, we obtain uplifted diffusion features for each Gaussian as a weighted average of 2D reference diffusion features $\mathbf{F}_{\text{ref}}$. Each 2D feature at the viewing direction $d$ and pixel $p$ contributes to the Gaussian feature $\mathbf{f}_i$ proportionally to the rendering weight $w_i(d,p)$ with normalization, where $\mathcal{I}_i$ denotes the set of view/pixel pairs contributing to each Gaussian. The uplifted feature $\hat{\mathbf{f}}_{i} \in \mathbb{R}^{\text{C}}$ is defined as:
\vspace{-2mm}
\begin{equation}
    \hat{\mathbf{f}}_{i}
= \frac{\mathbf{f}_{i}}{\lVert \mathbf{f}_{i} \rVert_{2}},
\qquad
\mathbf{f}_{i}
= \sum_{(d,p) \in \mathcal{I}_i}
\frac{w_{i}(d,p)}{\sum_{(d',p') \in \mathcal{I}_i} w_{i}(d',p')}
\mathbf{F}_{\text{ref}}(d,p)
\label{eq:uplifting}
\vspace{-3mm}
\end{equation}
This feed-forward uplifting enables explicit participation of 3D Gaussians at every denoising step, allowing end-to-end training with the diffusion model.

\input{figure/fusion_module}

\newpara{Feature refinement.}
After uplifting $\mathbf{F}_{\text{ref}}$ into the 3D Gaussians, we rasterize the geometry-constrained novel-view features $\mathbf{G}_{\text{tar}} \in \mathbb{R}^{Q \times \frac{h}{n} \times \frac{w}{n} \times \text{C}}$ using~\cref{eq:rendering_feat}. Our next goal is to fuse these geometry-aware features with the original novel-view diffusion features $\mathbf{F}_{\text{tar}}$. However, we observe that direct fusion alone is insufficient for recovering fine-grained details, as shown in~\cref{fig:feature_analysis}, consistent with the findings of~\cite{marrie2025ludvig}. 
To compensate for the information loss during lifting and rendering, we augment the rendered features $\mathbf{G}$ with Gaussian positional encoding (GS-PE) and refine them using a lightweight ResNet-based refinement network, $\mathcal{R}$. 
Specifically, for each pixel $(d, p)$, we identify the most contributing Gaussian $i^*(d,p) = \arg\max_i\, w_i(d,p)$, normalize its 3D center within the scene bounding box, and encode it alongside the rendering weight $w^*(d,p)$ via sinusoidal encoding $\gamma(\cdot)$ to obtain the positionally-encoded features $\mathbf{G}'$:
\vspace{-2mm}
\begin{equation}
    \mathbf{G}'(d,p) = \mathbf{G}(d,p) + 
    \left[\,\gamma(\tilde{x})\;\|\;\gamma(\tilde{y})\;\|\;
    \gamma(\tilde{z})\;\|\;\gamma(w^*)\,\right]
    \label{eq:gs_pe}
    \vspace{-2mm}
\end{equation}
where $\tilde{\mathbf{x}}(d,p) = (\tilde{x}, \tilde{y}, \tilde{z})$ denotes the 
normalized 3D center of $i^*(d,p)$, and 
$\gamma(v) = [\sin(\omega_k v),\, \cos(\omega_k v)]_{k=0}^{D'/2-1}$ with 
$\omega_k = \omega_0^{-2k/D'}$ and $D' = \text{C}/4$. 
This equips the refinement network with explicit 3D structural cues.
Afterwards, we get refined features $\hat{\mathbf{G}} = \mathcal{R}(\mathbf{G}')$.
Because the original input-view features $\mathbf{F}_{\text{ref}}$ serve as a reliable reference, we supervise $\mathcal{R}$ by aligning the refined input-view features $\hat{\mathbf{G}}_{\text{ref}} \in \mathbb{R}^{P \times \frac{h}{n} \times \frac{w}{n} \times \text{C}}$ with $\mathbf{F}_{\text{ref}}$ during training via a cosine similarity loss:
\vspace{-2mm}
\begin{equation}
\mathcal{L}_{\text{feat}}
= \left\| 1 - 
\frac{ \hat{\mathbf{G}}_{\text{ref}} \cdot \mathbf{F}_{\text{ref}}}
     { \| \hat{\mathbf{G}}_{\text{ref}} \|_2 \, \| \mathbf{F}_{\text{ref}} \|_2 }
\right\|_{2}
\label{eq:feature_loss}
\vspace{-3mm}
\end{equation}
\noindent\textbf{Adaptive feature fusion.}
Given the refined novel-view features $\hat{\mathbf{G}}_{\text{tar}}$, we fuse them with the diffusion features $\mathbf{F}_{\text{tar}}$ to obtain the geometry-augmented features $\hat{\mathbf{F}}_{\text{tar}}$. 
A naïve approach is to concatenate $\mathbf{F}_{\text{tar}}$ and $\hat{\mathbf{G}}_{\text{tar}}$, then project them back to the original channel dimension via MLPs:
\vspace{-3mm}
\begin{equation}
    \hat{\mathbf{F}}_{\text{tar}} = \mathrm{MLPs}\!\left(\mathbf{F}_{\text{tar}} \oplus 
    \hat{\mathbf{G}}_{\text{tar}}\right)
    \label{eq:naive_fusion}
\vspace{-3mm}
\end{equation}
However, this fusion relies entirely on the geometry prior and fails once it is corrupted (\cref{fig:fusion_module}). We instead propose an adaptive fusion that incorporates geometry based on its local reliability. 
Specifically, we first obtain the geometry-attended feature $\mathbf{F}_{\text{tar}}^{A}$ by cross-attending from $\mathbf{F}_{\text{tar}}$ (query) to $\hat{\mathbf{G}}_{\text{tar}}$ (key/value).
We then predict a pixel-wise confidence weight $\mathbf{W}_{\text{tar}} \in [-1, 1]$ via a gating MLP, and combine $\mathbf{F}_{\text{tar}}$ and $\mathbf{F}_{\text{tar}}^{A}$ weighted by $\mathbf{W}_{\text{tar}}$:
\vspace{-3mm}
\begin{equation}
    \mathbf{F}_{\text{tar}}^{A} = \mathrm{CrossAttn}\!\left(\mathbf{F}_{\text{tar}},\, 
    \hat{\mathbf{G}}_{\text{tar}}\right), \qquad
    \mathbf{W}_{\text{tar}} = \mathrm{Tanh}\!\left(\mathrm{MLP}\!\left(\mathbf{F}_{\text{tar}} 
    \oplus \mathbf{F}_{\text{tar}}^{A}\right)\right)
    \label{eq:gate}
\vspace{-3mm}
\end{equation}
\vspace{-2mm}
\begin{equation}
    \hat{\mathbf{F}}_{\text{tar}} = \mathbf{F}_{\text{tar}} + \mathbf{W}_{\text{tar}} \cdot 
    \mathbf{F}_{\text{tar}}^{A}
    \label{eq:adaptive_fusion}
\vspace{-1mm}
\end{equation}
To verify that $\mathbf{W}_{\text{tar}}$ behaves as intended, we measure the Pearson correlation between the absolute confidence weights, $\left| \mathbf{W}_{\text{tar}} \right|$ and 3D-GS uncertainty~\cite{hanson2025pup} across 3 reflective scenes (\cref{fig:fusion_module}). The consistent negative correlation ($r\approx\red{-0.30}$), which strengthens as denoising progresses and is most pronounced in reflective regions, supports our design claim: Adaptive Fusion downweights geometric guidance in regions where the geometry prior is unreliable.


\vspace{-3mm}
\subsection{Integration into Video Diffusion}
\label{subsec:geonvs}
\noindent\textbf{Integrating 3D-GS into video diffusion.} 
GeoNVS incorporates explicit 3D structural information into the denoising process by coupling 3D Gaussians with the diffusion model in feature space. At each denoising timestep $t$, GS-Adapter embeds $\mathbf{F}_{\text{ref}}^{t}$ into 3D Gaussians, rasterizes geometry-aware novel-view features, and adaptively injects them back into the diffusion model to correct geometrically inconsistent features throughout the entire denoising process.

\newpara{Multi-scale feature aggregation.} 
Multi-scale features play a crucial role in visual understanding~\cite{lin2017feature,wang2020deep,chen2017deeplab,wang2021pyramid,ranftl2021vision} and generation~\cite{rombach2022high,liu2025lumina}. 
To effectively combine geometry-guided features with diffusion features across different spatial scales, we leverage multi-scale encoder features via a DPT-style~\cite{ranftl2021vision} aggregation strategy, upsampling and hierarchically merging them to a unified feature of $\mathbf{F} \in \mathbb{R}^{(P+Q) \times \frac{h}{n} \times \frac{w}{n} \times \text{C}}$ before feeding into GS-Adapter (see~\cref{fig:detail_geonvs}). The geometry-corrected features are then downsampled to the original resolutions and added through skip connections to the decoder. We validate this design in \cref{subsec:ablation_studies}.

\newpara{Loss function.}
We train the multi-scale fusion module and GS-Adapter alongside trainable LoRA~\cite{hu2022lora} layers injected into the attention modules of the frozen video diffusion model. The total training objective is formulated by combining the latent space alignment loss~\cite{blattmann2023stable} and the feature alignment loss from~\cref{eq:feature_loss}:
\vspace{-2mm}
\begin{equation}
\mathcal{L} = \mathcal{L}_{\text{latent}} + 0.05 \mathcal{L}_{\text{feat}}
\label{eq:total_loss}
\vspace{-3mm}
\end{equation}

%% file: figure/model_detail.tex
\begin{figure*}[t]
\centering
\begin{subfigure}[t]{0.49\textwidth}
    \centering
    \includegraphics[width=\textwidth]{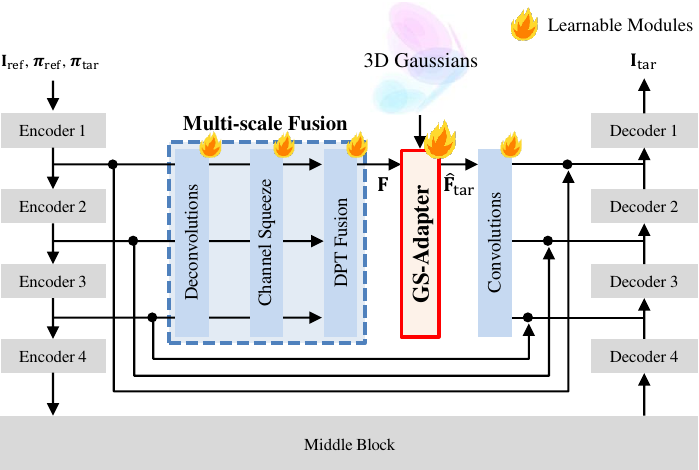}
    \caption{\textbf{Architecture Overview}}
    \label{fig:arch_overview}
\end{subfigure}
\hfill
\begin{subfigure}[t]{0.49\textwidth}
    \centering
    \includegraphics[width=\textwidth]{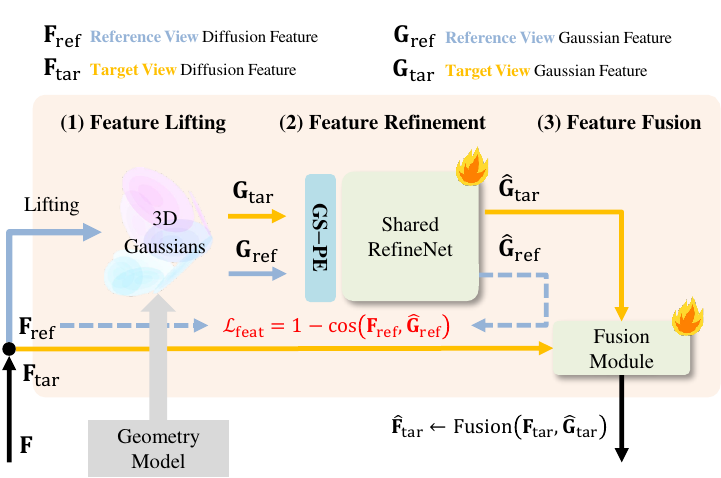}
    \caption{\textbf{GS-Adapter Pipeline}}
    \label{fig:adapter_pipeline}
\end{subfigure}

\vspace{-2mm}
\caption{\textbf{GeoNVS architecture.} 
(a) Overview of the integration with a video diffusion model. (b) The GS-Adapter pipeline for feature lifting, refinement, and fusion. All learnable modules (\textcolor{orange}{\faFire}) are trained with LoRA~\cite{hu2022lora}. During training, a consistency loss $\mathcal{L}_{\text{feat}}$ is applied to preserve geometric detail lost during feature lifting. Please refer to the supplementary material for details of the multi-scale fusion module and RefineNet.}
\vspace{-6mm}
\label{fig:detail_geonvs}
\end{figure*}

%% file: figure/fusion_module.tex
\begin{figure}[t]
\centering
\includegraphics[width=0.97\linewidth]{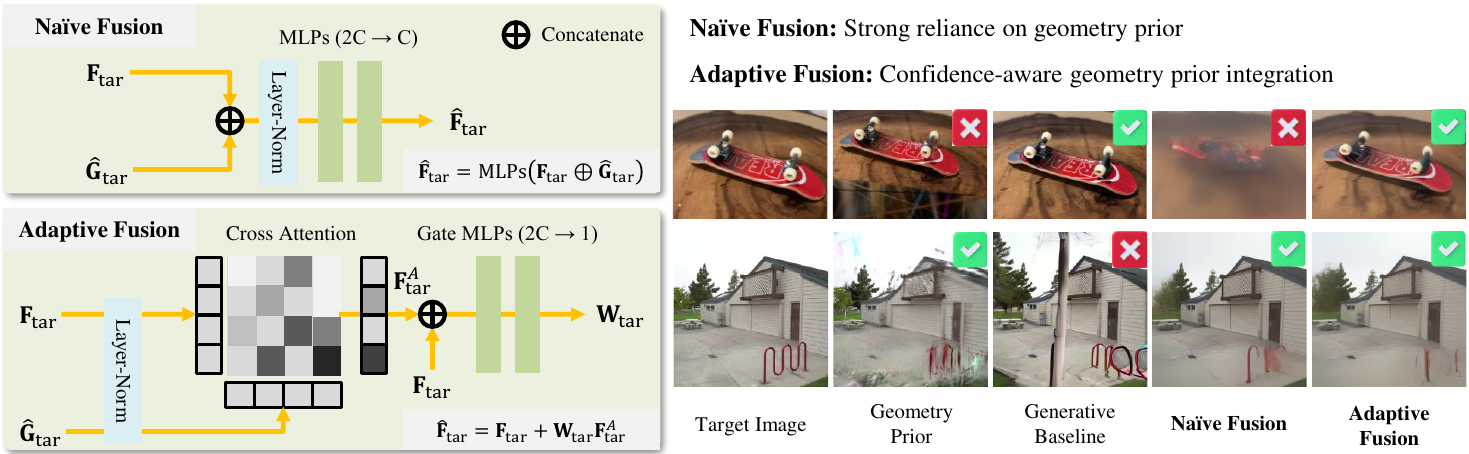} \\
\includegraphics[width=0.97\linewidth]{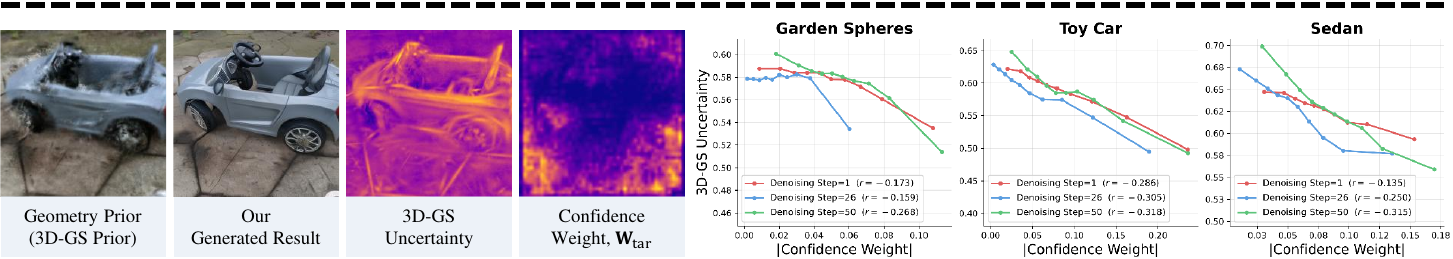}
\vspace{-3mm}
\caption{
\textbf{Feature fusion module of GS-Adapter.} Two fusion approaches are proposed to integrate the diffusion feature $\mathbf{F}_{\mathrm{tar}}$ and the geometry-aware feature $\hat{\mathbf{G}}_{\mathrm{tar}}$, producing the updated novel-view feature $\hat{\mathbf{F}}_{\mathrm{tar}}$. We adopt adaptive fusion as it remains effective even when either the geometry prior or the generative model fails.}
\vspace{-8mm}
\label{fig:fusion_module}
\end{figure}

%% file: sec/4_experiment.tex
\section{Experiment}
\label{sec:experiment}
\vspace{-2mm}
\subsection{Overview}
\label{subsec:overview_exp}
\vspace{-2mm}
We organize our experiments to validate three key claims of GeoNVS. First, GS-Adapter improves geometric consistency and camera controllability, quantified by camera pose error and Chamfer Distance, and further supported by a feature consistency metric (\cref{subsec:geo_exp,subsec:geo_feat_exp}). Second, GeoNVS achieves state-of-the-art photorealistic quality across diverse scenes, evaluated in~\cref{subsec:photo_exp} on 9 scenes under small/large-viewpoint and long-trajectory settings. Third, GS-Adapter supports plug-and-play integration with diverse geometry models without additional training, and is compatible with other diffusion backbones; we further show that its computational overhead can be substantially reduced via voxel-based Gaussian pruning with negligible performance degradation (\cref{subsec:ablation_studies}).
\vspace{-4mm}
\subsection{Implementation Details}
\label{subsec:impl_detail}
\vspace{-2mm}
We implement GeoNVS on two baselines, SEVA~\cite{zhou2025stable} and CameraCtrl~\cite{he2025cameractrl}, to demonstrate its compatibility. While the SEVA-based GeoNVS supports an arbitrary number of input and output views, the CameraCtrl-based GeoNVS is limited to a single reference view due to constraints of the baseline model.

\newpara{Training setup.}
We use the DL3DV-10K~\cite{ling2024dl3dv} dataset to train GeoNVS. Using VGGT~\cite{wang2025vggt} and InstantSplat~\cite{fan2024instantsplat}, we construct 3D Gaussians paired with 150K clips with varying reference-to-novel view splits $(P, Q) \in \{(1, 20),\allowbreak\ (3, 18),\allowbreak\ (6, 15),\allowbreak\ (9, 12),\allowbreak\ (12, 9)\}$ where the total number of views is fixed at 21. Please refer to Sec.~\red{E} of the supplementary material for details on dataset preprocessing. Both SEVA- and CameraCtrl-based GeoNVS are trained at resolutions $(h, w)$ of $384 \times 384$ and $320 \times 576$ with initial learning rates of $3 \times 10^{-5}$ and $1 \times 10^{-4}$, respectively. Our model is fine-tuned using LoRA~\cite{hu2022lora} on 8 NVIDIA A6000 GPUs (48GB). We set the rank and $\alpha$ to 16 for SEVA, and 4 for CameraCtrl. The training process takes approximately 80 hours with a total batch size of 8.

\newpara{Inference setup.}
We evaluate our method on the SEVA benchmark~\cite{zhou2025stable}, consisting 
of 9 scenes~\cite{jensen2014large,reizenstein2021common,xia2024rgbd,Knapitsch2017,
barron2022mip,ling2024dl3dv,zhou2018stereo,mildenhall2019local,wu2023omniobject3d} 
with input views $P$ varying from 1 to 9. We include Tanks and Temples~\cite{Knapitsch2017} under the CF-3DGS sampling protocol~\cite{fu2024colmap} for the view interpolation task (T\&T O split). For fair comparison, all combined methods (generative + geometry prior) use the same best-performing geometry prior per dataset. Note that MVSplat360~\cite{chen2024mvsplat360} is limited to MVSplat~\cite{chen2024mvsplat} by design. We evaluate all methods on a single A6000 GPU. Please refer to Sec.~\red{B} of the supplementary material for details, including evaluation dataset splits.

\newpara{Input-level injection of geometry prior.}
As a baseline for input-level geometry conditioning into the diffusion model in~\cref{fig:model_overview}-(b), we implement an input-level injection approach following prior work~\cite{wu2025difix3d+,wu2025genfusion,meng2021sdedit}.
Given the VAE-encoded latents $\mathbf{z}_\text{tar}$ of 3D-GS rendered novel-view images, we initialize the noisy latents as $\mathbf{x}_\text{tar} \leftarrow \mathbf{x} \cdot \sigma_{t_0} + \mathbf{z}_\text{tar}$, where $\sigma_{t_0}$ is the noise scale at the start timestep $t_0 = \lfloor T \cdot s \rfloor$ determined by a strength $s \in [0, 1]$. A larger $s$ allows the diffusion model more freedom to deviate from the geometry prior, while a smaller $s$ enforces stronger adherence to the rendered geometry; we set $s = 0.2$. We compare this approach against our GS-Adapter, which modulates internal diffusion features with a geometry prior rather than relying on rasterized images.

\input{table/maintable}
\vspace{-4mm}
\subsection{Perceptual Quality}
\label{subsec:photo_exp}
\vspace{-1mm}
\noindent\textbf{Metrics.}
We evaluate perceptual quality using PSNR, SSIM, and LPIPS — standard metrics in NVS — and report PSNR and SSIM as the primary metrics in the main manuscript; full results are in Sec.~\red{F} of the supplementary material.

\newpara{Robustness to viewpoint overlap.}
Following SEVA~\cite{zhou2025stable}, we evaluate on Set NVS, which generates a set of target views in arbitrary order, decomposed into small- and large-viewpoint settings based on the CLIP~\cite{radford2021learning} distance between reference and target views (threshold: 0.11). As shown in~\cref{tab:method_comparison,tab:large_viewpoint,fig:qual1,fig:qual3}, our method consistently outperforms comparison methods across most datasets, improving over our baseline SEVA~\cite{zhou2025stable} on all benchmarks and demonstrating robustness to both small and large viewpoint changes.

\input{table/subtables}
\input{table/geometry_metrics}
\newpara{Long trajectory generation.}
Trajectory NVS organizes target views as a smooth camera trajectory forming a video sequence. Leveraging the geometry prior from reference views, we adopt the two-pass sampling method from SEVA~\cite{zhou2025stable} to generate consistent, non-flickering long-term videos. As shown in~\cref{tab:long_traj_psnr,tab:long_traj_geo}, our method 
consistently improves over baseline SEVA by a large margin, with performance scaling favorably with more reference views.

\vspace{-4mm}
\subsection{Geometric Consistency and Camera Controllability}
\label{subsec:geo_exp}
\vspace{-1mm}
\noindent\textbf{Metrics.}
To validate geometric fidelity, we reconstruct point clouds and estimate camera trajectories from generated videos using ViPE~\cite{huang2025vipe}.
Camera controllability is evaluated by the translation error $T_{\textit{err}}$[cm] and rotation error $R_{\textit{err}}$[deg] between estimated and ground-truth camera poses following~\cite{he2025cameractrl}. Geometric consistency is quantified by mean Chamfer Distance ($CD$) between point clouds reconstructed from generated and ground-truth videos~\cite{chan2023generative,fan2017point}. We exclude scenes where ViPE~\cite{huang2025vipe} fails to reconstruct the ground-truth video ($T_{\textit{err}}$ or $R_{\textit{err}}$ $>$ 50).

\newpara{Geometric consistency.}
As shown in \cref{tab:long_traj_geo}, GeoNVS substantially improves geometric consistency over the SEVA baseline across all datasets. This is further corroborated by \cref{tab:geometric_consistency}, where our method consistently outperforms input-level geometry injection approaches — Difix3D\cite{wu2025difix3d+} and SEVA with input-level injection — by a large margin, demonstrating that feature-space modulation via GS-Adapter is more effective than injecting geometry at the input level.

\newpara{Camera controllability.}
GeoNVS substantially improves camera controllability over the SEVA baseline. As shown in~\cref{tab:long_traj_geo,fig:qual4}, the gains in synthesis quality stem primarily from improved camera controllability and geometry enhancement, with our method being especially effective in reducing translation deviation.
\cref{tab:geometric_consistency} further supports this: GeoNVS achieves consistently lower pose errors than the SEVA baseline across all three datasets, while competing methods that inject geometry priors at the input level suffer catastrophic degradation in controllability, with translation and rotation errors far exceeding those of the baseline. These results indicate that anchoring diffusion features to explicit 3D structure is key to reliable camera controllability.

\newpara{Geometry prior extrapolates to unseen regions.}
To evaluate whether geometry priors improve synthesis beyond observed regions, we construct co-visibility masks using VGGT-derived~\cite{wang2025vggt} point clouds following the co-visible mask construction in~\cite{fan2024instantsplat}. 
Specifically, we project only the input-view 3D points back onto each novel view, yielding per-pixel co-visibility masks; we measure PSNR$_\text{V}$ on co-visible and PSNR$_\text{U}$ on non-co-visible (unseen) regions. 
As shown in Table~\ref{tab:geometric_consistency}, our method consistently outperforms baselines on PSNR$_\text{U}$ across all datasets, with notable gains of \textbf{+1.06} on Mip360 and \textbf{+2.08} on DL3DV over SEVA. This demonstrates that our geometry guidance, explicitly conditioned on visible regions, facilitates geometry-consistent extrapolation into unseen areas.

\input{figure/feature_analysis}
\vspace{-4mm}
\subsection{Feature analysis of GS-Adapter}
\label{subsec:geo_feat_exp}
\vspace{-1mm}
To qualitatively validate GS-Adapter, we visualize intermediate features throughout the denoising process in~\cref{fig:feature_analysis}. The initial diffusion features $\mathbf{F}_\text{tar}^0$ are spatially incoherent due to high noise levels, whereas the rendered Gaussian features $\mathbf{G}_\text{tar}$ consistently exhibit sharp structures derived from the 3D geometry prior. After refinement, $\hat{\mathbf{G}}_\text{tar}$ recovers fine-grained details lost during lifting and rendering. Upon fusion, the geometry-corrected diffusion features progressively align with the underlying 3D structure as denoising advances, ultimately yielding photorealistic novel-view images with improved geometric consistency. These observations confirm that GS-Adapter effectively anchors diffusion features to explicit 3D geometry throughout the entire denoising process.


\vspace{-5mm}
\subsection{Ablation Studies}
\label{subsec:ablation_studies}
\vspace{-1mm}
\input{table/ablation_geometry_priors}

\noindent\textbf{Compatibility with geometry priors.}
A key advantage of GeoNVS is its plug-and-play compatibility with diverse feed-forward geometry models without requiring any additional training. To validate this, we integrate GS-Adapter with four different geometry priors~\cite{chen2024mvsplat,xu2025depthsplat,wang2025vggt,wang2025pi3} and evaluate using SEVA~\cite{zhou2025stable} as the generative backbone in~\cref{tab:ablation_lrms1}.
Both fusion variants consistently improve over the SEVA baseline regardless of the geometry prior used. 
However, Naïve Fusion remains sensitive to prior quality: while VGGT yields strong gains on T\&T (+3.37 dB), Pi3 degrades below the baseline on DTU (14.83 vs. 17.03), as observed in~\cref{fig:fusion_module}. 
Adaptive Fusion substantially mitigates this sensitivity, achieving robust and consistently superior results across all priors (19.02–19.18 dB on DTU; 19.22–19.26 dB on CO3D), and enabling flexible substitution of the geometry backbone without performance degradation.

\newpara{GeoNVS with CameraCtrl.}
To demonstrate that GS-Adapter generalizes beyond SEVA, we integrate it into CameraCtrl~\cite{he2025cameractrl} in~\cref{tab:ablation_lrms2}. Standalone geometry priors consistently underperform the CameraCtrl baseline, suggesting that geometry alone is insufficient in sparse-view settings. GeoNVS with Naïve Fusion improves over both the baseline and the standalone priors across most settings, achieving up to \textbf{+1.66} PSNR gain on average over the baseline, suggesting that GS-Adapter provides complementary benefit even on top of a strong geometry prior and is applicable beyond a single diffusion backbone.
\input{table/ablation_model}

\newpara{Analysis of model components.}
Table~\ref{tab:ablation_model} validates the contribution of each design choice in GS-Adapter.
Both the feature consistency loss $\mathcal{L}_\text{feat}$ and Gaussian positional encoding (GS-PE) contribute to performance, as removing either component leads to consistent drops across benchmarks, indicating that explicit 3D structural cues and fine-grained supervision are both necessary for recovering details lost during the lifting-and-rendering process.
RefineNet consistently yields performance gains, demonstrating its independent contribution beyond 3D-GS splatting.
Finally, using multi-scale features together in GeoNVS provides consistent improvements, indicating that geometry-guided features across multiple spatial resolutions are complementary.

\newpara{Comparison with input-level injection of geometry.}
As shown in \cref{tab:geometric_consistency}, input-level injection of geometry priors causes severe degradation in camera controllability and geometric consistency, performing even worse than SEVA, a purely generative baseline. This failure stems from the fact that rasterized images introduce view-dependent color noise that corrupts the structural signal rather than reinforcing it. In contrast, the proposed GS-Adapter leverages only the structural information encoded in 3D Gaussians to modulate internal diffusion features, consistently reducing camera pose errors and Chamfer Distance across all datasets. \cref{tab:ablation_model} further reinforces this finding: input-level injection yields synthesis quality gains only when the geometry prior is near-perfect, whereas GS-Adapter consistently improves PSNR regardless of prior quality. These results collectively demonstrate that feature-level geometry modulation is both more robust and more effective than image-level injection.

\input{figure/qualitative1}
\input{figure/qualitative3}
\input{figure/gaussian_pruning}

\newpara{Pose-free experiment.}
We extend GeoNVS to unknown pose scenarios by integrating ViPE~\cite{huang2025vipe}, 
which outperforms VGGT~\cite{wang2025vggt} in robustness on complex scenes. Given poses estimated by ViPE from input images, we optimize 3D-GS using the VGGT point cloud, with scale aligned to ViPE poses solved via~\cref{eq:scale_optimize}.
\setlength{\columnsep}{4pt}
\begin{wraptable}{r}{0.45\columnwidth}
\vspace{-7mm}
\raggedleft
\tiny
\setlength{\tabcolsep}{0.45pt}
\renewcommand{\arraystretch}{1.0}
\begin{tabular}{l c c c c c c}
\toprule
Dataset ($P$)
& \multicolumn{3}{c}{DL3DV (6)}
& \multicolumn{3}{c}{T\&T (3)} \\
\cmidrule(lr){1-1}
\cmidrule(lr){2-4}
\cmidrule(lr){5-7}
Metric
& $T_{err}$ & $R_{err}$ & PSNR
& $T_{err}$ & $R_{err}$ & PSNR \\
\midrule
\multicolumn{5}{l}{\textbf{w/ GT Pose}} \\
SEVA~\cite{zhou2025stable}
& 19.68 & 2.80 & 15.64
& 6.08 & 2.87 & 18.63 \\
\rowcolor[gray]{0.9} \textbf{Ours}
& \textbf{13.27} & \textbf{2.01} & \textbf{17.99}
& 3.56 & \textbf{1.52} & 22.24 \\
\midrule
\multicolumn{5}{l}{\textbf{w/ ViPE~\cite{huang2025vipe} Pose}} \\
SEVA~\cite{zhou2025stable}
& 25.11 & 3.54 & 15.38
& 2.73 & 4.87 & 19.24 \\
\rowcolor[gray]{0.9} \textbf{Ours}
& 20.64 & 2.98 & 17.87
& \textbf{1.05} & 2.40 & \textbf{22.31} \\
\bottomrule
\end{tabular}
\captionsetup{labelformat=empty, skip=0pt}
\caption{}
\label{tab:pose_free}
\vspace{-\baselineskip}
\vspace{-5mm}
\end{wraptable}
As shown in the \hyperref[tab:pose_free]{right table}, GeoNVS maintains high synthesis quality without ground-truth intrinsics or extrinsics. On T\&T~\cite{Knapitsch2017}, the ViPE pose setting slightly surpasses the GT setting, likely because ViPE estimates poses more robustly than the noisy COLMAP~\cite{schoenberger2016sfm,schoenberger2016mvs}-derived poses.
\\
\vspace{-5mm}

\newpara{Runtime and memory efficiency.}
We measure inference time, peak GPU memory, and PSNR across 10 scenes from DL3DV~\cite{ling2024dl3dv} to analyze the computational overhead relative to SEVA (\cref{fig:performance}). The primary bottleneck is the number of Gaussians from the geometry prior, which raises inference time to 4.19 sec/frame and peak memory to 18.03 GB. Applying voxel-based Gaussian pruning~\cite{jiang2025anysplat} reduces these to 1.89 sec/frame (2.22$\times$ speedup) and 12.93 GB, on par with SEVA's 12.81 GB, while maintaining a PSNR advantage over SEVA.

\input{figure/qualitative_geo}
\input{figure/limitation}

%% file: table/maintable.tex
\vspace{-1mm}
\begin{table}[!htbp]
\centering
\caption{\textbf{Perceptual metrics on small-viewpoint set NVS.} 
The small-viewpoint set contains scenes with high view overlap between reference and target views.
$P$ denotes the number of reference images. 
For $P$=1, we sweep the unit length for camera normalization~\cite{zhou2025stable} to resolve scale ambiguity. 
All combined methods (generative $+$ geometry) use the best feed-forward geometry model per dataset as the geometry prior.} 
\label{tab:method_comparison}
\vspace{-6mm}
  \begin{subtable}[t]{\textwidth}
    \centering
    \caption{\textbf{PSNR $\uparrow$ on small-viewpoint set NVS.}}
    \vspace{-2mm}
    \label{tab:method_comparison_psnr}
    \resizebox{\columnwidth}{!}{%
    \setlength{\tabcolsep}{1.1pt}
    \renewcommand{\arraystretch}{1.4}
    \setlength{\aboverulesep}{2pt}
    \setlength{\belowrulesep}{2pt}
    \begin{tabular}{l *{22}{c}}
    \toprule
    \multirow{3}{*}{Method}
    & dataset
    & OO3D~\cite{wu2023omniobject3d}
    & \multicolumn{4}{c}{RE10K~\cite{zhou2018stereo}} 
    & \multicolumn{2}{c}{LLFF~\cite{mildenhall2019local}} 
    & \multicolumn{2}{c}{DTU~\cite{jensen2014large}}
    & \multicolumn{2}{c}{CO3D~\cite{reizenstein2021common}} 
    & \multicolumn{2}{c}{WRGBD~\cite{xia2024rgbd}} 
    & Mip360~\cite{barron2022mip} 
    & \multicolumn{2}{c}{DL3DV~\cite{ling2024dl3dv}} 
    & \multicolumn{2}{c}{T\&T~\cite{Knapitsch2017}} \\
    \cmidrule(lr){2-2}
    \cmidrule(lr){3-3}
    \cmidrule(lr){4-7}
    \cmidrule(lr){8-9}
    \cmidrule(lr){10-11}
    \cmidrule(lr){12-13}
    \cmidrule(lr){14-15}
    \cmidrule(lr){16-16}
    \cmidrule(lr){17-18}
    \cmidrule(lr){19-20}
    & split
    & S 
      & D & P & \multicolumn{2}{c}{R}
      & \multicolumn{2}{c}{R} 
      & \multicolumn{2}{c}{R} 
      & V & R 
      & S$_e$ & S$_h$ 
      & R 
      & S & L
      & \multicolumn{2}{c}{O} & Average \\
    \cmidrule(lr){2-2}
    \cmidrule(lr){3-3}
    \cmidrule(lr){4-4}
    \cmidrule(lr){5-5}
    \cmidrule(lr){6-7}
    \cmidrule(lr){8-9}
    \cmidrule(lr){10-11}
    \cmidrule(lr){12-12}
    \cmidrule(lr){13-13}
    \cmidrule(lr){14-14}
    \cmidrule(lr){15-15}
    \cmidrule(lr){16-16}
    \cmidrule(lr){17-17}
    \cmidrule(lr){18-18}
    \cmidrule(lr){19-20}
    & $P$
      & 3
      & 1 & 2 & 1 & 3
      & 1 & 3
      & 1 & 3
      & 1 & 3 
      & 3 & 6 
      & 6 
      & 6 & 9 
      & 2 & 3 & \\
    \midrule
    \multicolumn{20}{l}{\textbf{Feed-forward geometry models}} \\
    \multicolumn{2}{l}{MVSplat~\cite{chen2024mvsplat}} & 21.82 & 10.68 & \snd{25.56} & 11.26 & 23.35 & 5.69 & 13.68 & 7.47 & 11.35 & 10.12 & 12.04 & 12.89 & 12.38 & 13.45 & 13.89 & 14.07 & 12.52 & 13.45 & 13.65  \\
    \multicolumn{2}{l}{HiSplat~\cite{tang2024hisplat}} & 21.98 & 12.38 & 25.50 & 13.89 & 25.81 & 5.75 & 15.16 & 7.11 & 12.37 & 10.59 & 12.55 & 13.49 & 12.48 & 13.90 & 14.04 & 14.20 & 12.38 & 15.57 & 14.40 \\
    \multicolumn{2}{l}{DepthSplat~\cite{xu2025depthsplat}} & 22.66 & 13.02 & \fst{25.57} & 14.49 & 26.99 & 5.91 & 15.02 & 8.79 & 14.19 & 11.29 & 15.08 & 13.92 & 13.50 & 15.04 & 15.87 & 16.88 & 15.76 & 17.99 & 15.67 \\
    \multicolumn{2}{l}{VGGT~\cite{wang2025vggt}} & 22.12 & 13.73 & 23.98 & 13.95 & \snd{27.48} & 10.89 & \fst{20.08} & 7.05 & \trd{17.15} & 10.48 & \trd{15.39} & 11.09 & 12.56 & \snd{15.49} & \snd{17.09} & \snd{18.31} & \fst{20.92} & \fst{23.96} & 16.76 \\
    \multicolumn{2}{l}{Pi3~\cite{wang2025pi3}} & 18.48 & 10.56 & 24.35 & 11.00 & \fst{27.66} & 8.80 & 15.54 & 8.82 & 7.81 & 12.08 & 12.60 & 10.54 & 10.46 & 13.16 & 16.03 & 17.23 & 14.46 & 16.98 & 14.25 \\
    \midrule
    \multicolumn{20}{l}{\textbf{Generative models}} \\
    \multicolumn{2}{l}{MotionCtrl~\cite{wang2024motionctrl}} & - & 13.10 & - & 13.86 & - & 11.40 & - & 8.41 & - & 13.64 & - & - & - & - & - & - & - & - & 12.08 \\
    \multicolumn{2}{l}{CameraCtrl~\cite{he2025cameractrl}} & - & 13.08 & - & 14.09 & - & 11.88 & - & 8.72 & - & 13.70 & - & - & - & - & - & - & - & - & 12.29 \\
    \multicolumn{2}{l}{ViewCrafter~\cite{yu2024viewcrafter}} & 18.56 & 14.37 & - & \snd{17.44} & 15.83 & 11.79 & 11.39 & 9.22 & 8.49 & \trd{15.87} & 9.89 & 9.23 & 9.37 & 10.81 & 10.51 & 9.75 & 10.74 & 11.13 & 12.02 \\
    \multicolumn{2}{l}{SEVA~\cite{zhou2025stable}} & \snd{26.28} & \trd{14.51} & 22.77 & 15.25 & 24.44 & \trd{12.22} & 17.12 & \snd{10.54} & 17.03 & \snd{16.29} & \snd{17.22} & \snd{15.96} & \snd{15.71} & \trd{15.38} & 15.84 & 16.87 & 16.35 & 18.85 & \snd{17.15} \\
    \midrule
    \multicolumn{20}{l}{\textbf{Generative model + Geometry prior}} \\
    \multicolumn{2}{l}{MVSplat360~\cite{chen2024mvsplat360}} & 21.45 & 13.68 & 20.20 & 13.44 & 20.97 & \snd{12.79} & 15.31 & \trd{9.10} & 11.38 & 12.69 & 14.29 & \trd{14.34} & \trd{13.22} & 14.15 & 15.09 & 15.26 & 14.22 & 15.84 & 14.86 \\
    \multicolumn{2}{l}{GenFusion~\cite{wu2025genfusion}} & 20.90 & \snd{14.87} & 22.48 & \trd{15.88} & 22.69 & 11.39 & 17.87 & 8.36 & 13.75 & 12.15 & 12.99 & 13.38 & 12.47 & 11.99 & 13.11 & 14.78 & 18.25 & 20.14 & 15.30 \\
    \multicolumn{2}{l}{Difix3D~\cite{wu2025difix3d+}} & \trd{23.05} & 12.92 & \trd{25.23} & 14.37 & \trd{27.14} & 10.85 & \snd{19.59} & 8.79 & \snd{17.22} & 11.93 & 15.23 & 12.60 & 12.67 & 14.95 & \trd{16.33} & \trd{17.33} & \snd{20.68} & \snd{23.36} & \trd{16.89} \\
    \rowcolor[gray]{0.9}\multicolumn{2}{l}{GeoNVS (\textbf{Ours})} & \fst{26.58} & \fst{16.66} & 24.78 & \fst{17.62} & 27.11 & \fst{14.10} & \trd{18.29} & \fst{12.12} & \fst{19.18} & \fst{17.14} & \fst{19.26} & \fst{18.23} & \fst{17.12} & \fst{16.57} & \fst{18.11} & \fst{19.66} & \trd{18.76} & \trd{22.24} & \fst{19.09} \\
    \bottomrule
    \end{tabular}
    }
  \end{subtable}

  \vspace{2.5pt}

  \begin{subtable}[t]{\textwidth}
    \centering
    \caption{\textbf{SSIM $\uparrow$ on small-viewpoint set NVS.}}
    \vspace{-2mm}
    \label{tab:method_comparison_ssim}
    \resizebox{\columnwidth}{!}{%
    \setlength{\tabcolsep}{1.1pt}
    \renewcommand{\arraystretch}{1.4}
    \setlength{\aboverulesep}{2pt}
    \setlength{\belowrulesep}{2pt}
    \begin{tabular}{l *{22}{c}}
    \toprule
    \multirow{3}{*}{Method}
    & dataset
    & OO3D~\cite{wu2023omniobject3d}
    & \multicolumn{4}{c}{RE10K~\cite{zhou2018stereo}} 
    & \multicolumn{2}{c}{LLFF~\cite{mildenhall2019local}} 
    & \multicolumn{2}{c}{DTU~\cite{jensen2014large}}
    & \multicolumn{2}{c}{CO3D~\cite{reizenstein2021common}} 
    & \multicolumn{2}{c}{WRGBD~\cite{xia2024rgbd}} 
    & Mip360~\cite{barron2022mip} 
    & \multicolumn{2}{c}{DL3DV~\cite{ling2024dl3dv}} 
    & \multicolumn{2}{c}{T\&T~\cite{Knapitsch2017}} \\
    \cmidrule(lr){2-2}
    \cmidrule(lr){3-3}
    \cmidrule(lr){4-7}
    \cmidrule(lr){8-9}
    \cmidrule(lr){10-11}
    \cmidrule(lr){12-13}
    \cmidrule(lr){14-15}
    \cmidrule(lr){16-16}
    \cmidrule(lr){17-18}
    \cmidrule(lr){19-20}
    & split
    & S 
      & D & P & \multicolumn{2}{c}{R}
      & \multicolumn{2}{c}{R} 
      & \multicolumn{2}{c}{R} 
      & V & R 
      & S$_e$ & S$_h$ 
      & R 
      & S & L
      & \multicolumn{2}{c}{O} & Average \\
    \cmidrule(lr){2-2}
    \cmidrule(lr){3-3}
    \cmidrule(lr){4-4}
    \cmidrule(lr){5-5}
    \cmidrule(lr){6-7}
    \cmidrule(lr){8-9}
    \cmidrule(lr){10-11}
    \cmidrule(lr){12-12}
    \cmidrule(lr){13-13}
    \cmidrule(lr){14-14}
    \cmidrule(lr){15-15}
    \cmidrule(lr){16-16}
    \cmidrule(lr){17-17}
    \cmidrule(lr){18-18}
    \cmidrule(lr){19-20}
    & $P$
      & 3
      & 1 & 2 & 1 & 3
      & 1 & 3
      & 1 & 3
      & 1 & 3 
      & 3 & 6 
      & 6 
      & 6 & 9 
      & 2 & 3 & \\
    \midrule
    \multicolumn{20}{l}{\textbf{Feed-forward geometry models}} \\
    \multicolumn{2}{l}{HiSplat~\cite{tang2024hisplat}} & 0.877 & 0.444 & \fst{0.872} & 0.540 & 0.902 & 0.01 & 0.384 & 0.149 & 0.467 & 0.286 & 0.434 & 0.434 & 0.419 & 0.290 & 0.419 & 0.457 & 0.360 & 0.468 & 0.456 \\
    \multicolumn{2}{l}{DepthSplat~\cite{xu2025depthsplat}} & \trd{0.897} & 0.468 & \snd{0.868} & 0.529 & \fst{0.918} & 0.02 & 0.332 & 0.262 & 0.485 & 0.337 & 0.492 & 0.441 & 0.434 & 0.303 & 0.465 & \trd{0.528} & 0.472 & 0.535 & 0.488 \\
    \multicolumn{2}{l}{VGGT~\cite{wang2025vggt} + InstantSplat~\cite{fan2024instantsplat}} & 0.892 & 0.499 & 0.798 & 0.561 & \snd{0.914} & 0.234 & \fst{0.672} & 0.215 & \trd{0.640} & 0.297 & 0.524 & 0.409 & 0.432 & \snd{0.362} & \snd{0.518} & \snd{0.571} & \fst{0.716} & \fst{0.807} & \trd{0.559} \\
    \midrule
    \multicolumn{20}{l}{\textbf{Generative models}} \\
    \multicolumn{2}{l}{CameraCtrl~\cite{he2025cameractrl}} & - & 0.491 & - & 0.578 & - & \trd{0.294} & - & 0.400 & - & 0.451 & - & - & - & - & - & - & - & - & 0.443 \\
    \multicolumn{2}{l}{ViewCrafter~\cite{yu2024viewcrafter}} & 0.861 & \trd{0.551} & - & \fst{0.693} & 0.647 & 0.283 & 0.253 & 0.370 & 0.274 & \snd{0.520} & 0.300 & 0.233 & 0.215 & 0.194 & 0.241 & 0.230 & 0.286 & 0.300 & 0.379 \\
    \multicolumn{2}{l}{SEVA~\cite{zhou2025stable}} & \fst{0.926} & 0.534 & 0.777 & 0.616 & 0.842 & \trd{0.294} & 0.467 & \snd{0.430} & \snd{0.659} & \trd{0.507} & \trd{0.551} & \snd{0.527} & \snd{0.522} & \trd{0.339} & 0.450 & 0.513 & 0.521 & 0.608 & \snd{0.560} \\
    \midrule
    \multicolumn{20}{l}{\textbf{Generative model + Geometry prior}} \\
    \multicolumn{2}{l}{GenFusion~\cite{wu2025genfusion}} & 0.876 & \snd{0.559} & 0.818 & \trd{0.618} & 0.864 & 0.271 & \trd{0.554} & 0.355 & 0.556 & 0.438 & 0.449 & 0.453 & 0.425 & 0.282 & 0.394 & 0.492 & \trd{0.630} & 0.689 & 0.540 \\
    \multicolumn{2}{l}{Difix3D~\cite{wu2025difix3d+}} & \trd{0.898} & 0.463 & \trd{0.857} & 0.528 & 0.903 & 0.230 & \snd{0.638} & 0.342 & 0.637 & 0.375 & 0.479 & 0.350 & 0.333 & 0.301 & \trd{0.481} & 0.519 & \snd{0.696} & \snd{0.779} & 0.545 \\
    \rowcolor[gray]{0.9}\multicolumn{2}{l}{GeoNVS (\textbf{Ours})} & \snd{0.904} & \fst{0.612} & 0.826 & \snd{0.680} & 0.895 & \fst{0.346} & 0.514 & \fst{0.480} & \fst{0.691} & \fst{0.550} & \fst{0.627} & \fst{0.607} & \fst{0.577} & \fst{0.383} & \fst{0.562} & \fst{0.635} & 0.603 & \trd{0.695} & \fst{0.622} \\
    \bottomrule
    \end{tabular}
    }
  \end{subtable}

  \begin{subtable}[t]{\textwidth}
    \centering
    \caption{\textbf{LPIPS $\downarrow$ on small-viewpoint set NVS.}}
    \vspace{-2mm}
    \label{tab:method_comparison_lpips}
    \resizebox{\columnwidth}{!}{%
    \setlength{\tabcolsep}{1.1pt}
    \renewcommand{\arraystretch}{1.4}
    \setlength{\aboverulesep}{2pt}
    \setlength{\belowrulesep}{2pt}
    \begin{tabular}{l *{22}{c}}
    \toprule
    \multirow{3}{*}{Method}
    & dataset
    & OO3D~\cite{wu2023omniobject3d}
    & \multicolumn{4}{c}{RE10K~\cite{zhou2018stereo}} 
    & \multicolumn{2}{c}{LLFF~\cite{mildenhall2019local}} 
    & \multicolumn{2}{c}{DTU~\cite{jensen2014large}}
    & \multicolumn{2}{c}{CO3D~\cite{reizenstein2021common}} 
    & \multicolumn{2}{c}{WRGBD~\cite{xia2024rgbd}} 
    & Mip360~\cite{barron2022mip} 
    & \multicolumn{2}{c}{DL3DV~\cite{ling2024dl3dv}} 
    & \multicolumn{2}{c}{T\&T~\cite{Knapitsch2017}} \\
    \cmidrule(lr){2-2}
    \cmidrule(lr){3-3}
    \cmidrule(lr){4-7}
    \cmidrule(lr){8-9}
    \cmidrule(lr){10-11}
    \cmidrule(lr){12-13}
    \cmidrule(lr){14-15}
    \cmidrule(lr){16-16}
    \cmidrule(lr){17-18}
    \cmidrule(lr){19-20}
    & split
    & S 
      & D & P & \multicolumn{2}{c}{R}
      & \multicolumn{2}{c}{R} 
      & \multicolumn{2}{c}{R} 
      & V & R 
      & S$_e$ & S$_h$ 
      & R 
      & S & L
      & \multicolumn{2}{c}{O} & Average \\
    \cmidrule(lr){2-2}
    \cmidrule(lr){3-3}
    \cmidrule(lr){4-4}
    \cmidrule(lr){5-5}
    \cmidrule(lr){6-7}
    \cmidrule(lr){8-9}
    \cmidrule(lr){10-11}
    \cmidrule(lr){12-12}
    \cmidrule(lr){13-13}
    \cmidrule(lr){14-14}
    \cmidrule(lr){15-15}
    \cmidrule(lr){16-16}
    \cmidrule(lr){17-17}
    \cmidrule(lr){18-18}
    \cmidrule(lr){19-20}
    & $P$
      & 3
      & 1 & 2 & 1 & 3
      & 1 & 3
      & 1 & 3
      & 1 & 3 
      & 3 & 6 
      & 6 
      & 6 & 9 
      & 2 & 3 & \\
    \midrule
    \multicolumn{20}{l}{\textbf{Feed-forward geometry models}} \\
    \multicolumn{2}{l}{HiSplat~\cite{tang2024hisplat}} & 0.155 & 0.522 & 0.110 & 0.484 & 0.094 & 1.02 & 0.479 & 0.783 & 0.483 & 0.645 & 0.581 & 0.496 & 0.610 & 0.691 & 0.627 & 0.650 & 0.612 & 0.474 & 0.529 \\
    \multicolumn{2}{l}{DepthSplat~\cite{xu2025depthsplat}} & 0.134 & 0.444 & \trd{0.100} & 0.414 & \snd{0.074} & 0.979 & 0.351 & 0.550 & 0.362 & 0.553 & 0.498 & \trd{0.431} & 0.493 & 0.552 & 0.487 & 0.477 & 0.332 & 0.235 & 0.415 \\
    \multicolumn{2}{l}{VGGT~\cite{wang2025vggt} + InstantSplat~\cite{fan2024instantsplat}} & 0.172 & 0.441 & 0.138 & 0.454 & \trd{0.078} & \trd{0.517} & \fst{0.153} & 0.779 & 0.340 & 0.608 & 0.530 & 0.696 & 0.623 & 0.618 & 0.466 & 0.497 & \snd{0.194} & \snd{0.127} & 0.413 \\
    \midrule
    \multicolumn{20}{l}{\textbf{Generative models}} \\
    \multicolumn{2}{l}{CameraCtrl~\cite{he2025cameractrl}} & - & 0.514 & - & 0.492 & - & 0.610 & - & \trd{0.611} & - & 0.540 & - & - & - & - & - & - & - & - & 0.553 \\
    \multicolumn{2}{l}{ViewCrafter~\cite{yu2024viewcrafter}} & 0.211 & 0.405 & - & \snd{0.338} & 0.385 & 0.527 & 0.562 & 0.634 & 0.690 & \trd{0.425} & 0.727 & 0.715 & 0.727 & 0.675 & 0.649 & 0.680 & 0.542 & 0.516 & 0.553 \\
    \multicolumn{2}{l}{SEVA~\cite{zhou2025stable}} & \fst{0.062} & 0.400 & 0.131 & 0.402 & 0.095 & \snd{0.469} & \trd{0.212} & \snd{0.524} & \snd{0.210} & \snd{0.399} & \snd{0.333} & \snd{0.339} & \snd{0.331} & \snd{0.367} & \snd{0.305} & \snd{0.288} & 0.238 & 0.164 & \snd{0.293} \\
    \midrule
    \multicolumn{20}{l}{\textbf{Generative model + Geometry prior}} \\
    \multicolumn{2}{l}{GenFusion~\cite{wu2025genfusion}} & 0.158 & 0.408 & 0.183 & \trd{0.384} & 0.159 & 0.554 & 0.301 & 0.702 & 0.430 & 0.621 & 0.616 & 0.507 & 0.577 & 0.655 & 0.537 & 0.520 & 0.278 & 0.223 & 0.434 \\
    \multicolumn{2}{l}{Difix3D~\cite{wu2025difix3d+}} & \trd{0.106} & 0.447 & \fst{0.097} & 0.422 & 0.083 & 0.520 & \fst{0.153} & 0.648 & \trd{0.240} & 0.544 & \trd{0.456} & 0.508 & 0.549 & \trd{0.425} & \trd{0.313} & \trd{0.326} & \fst{0.158} & \fst{0.101} & \trd{0.339} \\
    \rowcolor[gray]{0.9}\multicolumn{2}{l}{GeoNVS (\textbf{Ours})} & \snd{0.074} & \fst{0.325} & 0.113 & \fst{0.319} & \fst{0.071} & \fst{0.412} & \snd{0.208} & \fst{0.449} & \fst{0.174} & \fst{0.393} & \fst{0.279} & \fst{0.275} & \fst{0.290} & \fst{0.329} & \fst{0.245} & \fst{0.235} & \trd{0.225} & \trd{0.138} & \fst{0.253} \\
    \bottomrule
    \end{tabular}
    }
  \end{subtable}

\vspace{-6mm}
\end{table}

%% file: table/subtables.tex
\begin{table}[t]
\centering
\begin{minipage}[t]{0.5\textwidth}
\centering
\captionsetup{hypcap=false}
\captionof{table}{\textbf{Large-viewpoint set NVS.} The large-viewpoint set contains scenes with low view overlap between reference and target views. $P$ denotes the number of reference images. We use DepthSplat~\cite{xu2025depthsplat} as the geometry prior for both combined methods.}
\vspace{-5mm}
\begin{subtable}[t]{0.99\textwidth}
\centering
\caption{\textbf{PSNR $\uparrow$ results.}}
\vspace{-3mm}
\label{tab:large_viewpoint}
\resizebox{\linewidth}{!}{%
\setlength{\tabcolsep}{1.0pt}
\renewcommand{\arraystretch}{1.3}
\setlength{\aboverulesep}{2pt}
\setlength{\belowrulesep}{2pt}
\begin{tabular}{l c c c c c c c c c c c}
\toprule
\multirow{3}{*}{Method}
& dataset
& CO3D~\cite{reizenstein2021common}
& \multicolumn{2}{c}{WRGBD~\cite{xia2024rgbd}}
& \multicolumn{2}{c}{Mip360~\cite{barron2022mip}} 
& \multicolumn{2}{c}{DL3DV~\cite{ling2024dl3dv}} \\
\cmidrule(lr){2-2}
\cmidrule(lr){3-3}
\cmidrule(lr){4-5}
\cmidrule(lr){6-7}
\cmidrule(lr){8-9}
& split & R & \multicolumn{2}{c}{S$_h$} & \multicolumn{2}{c}{R} & \multicolumn{2}{c}{S} & Average \\
\cmidrule(lr){2-2}
\cmidrule(lr){3-3}
\cmidrule(lr){4-5}
\cmidrule(lr){6-7}
\cmidrule(lr){8-9}
& $P$ & 1 & 1 & 3 & 1 & 3 & 1 & 3 \\
\midrule
\multicolumn{6}{l}{\textbf{Geometry prior}} \\
\multicolumn{2}{l}{DepthSplat~\cite{xu2025depthsplat}} & 8.54 & 7.31 & \trd{11.94} & 9.20 & \trd{14.22} & 7.37 & \snd{14.67} & 10.46 \\
\midrule
\multicolumn{6}{l}{\textbf{Generative models}} \\
\multicolumn{2}{l}{MotionCtrl~\cite{wang2024motionctrl}} & \trd{11.77} & 10.33 & - & \trd{11.04} & - & 10.99 & - & \trd{11.03} \\
\multicolumn{2}{l}{CameraCtrl~\cite{he2025cameractrl}} & 11.71 & \trd{10.38} & - & 10.94 & - & 11.05 & - & 11.02 \\
\multicolumn{2}{l}{ViewCrafter~\cite{yu2024viewcrafter}} & 11.11 & - & 9.32 & 10.79 & 10.38 & \trd{11.18} & 10.20 & 10.50 \\
\multicolumn{2}{l}{SEVA~\cite{zhou2025stable}} & \snd{12.27} & \snd{11.27} & \snd{13.72} & \snd{11.19} & \snd{14.30} & \snd{11.48} & \trd{14.27} & \snd{12.64} \\
\midrule
\multicolumn{6}{l}{\textbf{Generative model + Geometry prior}} \\
\multicolumn{2}{l}{GenFusion~\cite{wu2025genfusion}} & 10.21 & 9.60 & 11.37 & 9.84 & 11.88 & 10.11 & 13.33 & 10.91 \\
\multicolumn{2}{l}{Difix3D~\cite{wu2025difix3d+}} & 8.56 & 7.33 & 11.81 & 9.30 & 13.83 & 7.48 & 14.24 & 10.36 \\
\rowcolor[gray]{0.9}\multicolumn{2}{l}{GeoNVS (\textbf{Ours})} & \fst{14.66} & \fst{12.72} & \fst{15.08} & \fst{12.34} & \fst{15.47} & \fst{13.07} & \fst{16.09} & \fst{14.20} \\
\bottomrule
\end{tabular}
}
\end{subtable}
\end{minipage}
\hfill
\begin{minipage}[t]{0.49\textwidth}
\centering
\captionsetup{hypcap=false}
\captionof{table}{\textbf{Long trajectory NVS.} The long trajectory NVS measures the long-term generation ability of generative NVS. $P$ denotes the number of reference images. We use VGGT~\cite{wang2025vggt} as the geometry prior for our method.}
\vspace{-5mm}
\begin{subtable}[t]{1.0\textwidth}
    \centering
    \caption{\textbf{PSNR $\uparrow$ results.}}
    \vspace{-3mm}
    \label{tab:long_traj_psnr}
    \resizebox{\linewidth}{!}{%
    \setlength{\tabcolsep}{1.1pt}
    \renewcommand{\arraystretch}{1.3}
    \begin{tabular}{l *{10}{c}}
    \toprule
    \multirow{2}{*}{Method} & dataset & \multicolumn{3}{c}{Mip360~\cite{barron2022mip}} & \multicolumn{3}{c}{DL3DV~\cite{ling2024dl3dv}} & \multicolumn{3}{c}{T\&T~\cite{Knapitsch2017}} \\
    \cmidrule(lr){2-2}\cmidrule(lr){3-5}\cmidrule(lr){6-8}\cmidrule(lr){9-11}
    & $P$ & 3 & 6 & 9 & 3 & 6 & 9 & 2 & 3 & 6 \\
    \midrule
    \multicolumn{2}{l}{DepthSplat~\cite{xu2025depthsplat}} & \snd{13.22} & 13.91 & 14.57 & \trd{14.79} & \trd{15.90} & 16.68 & 15.69 & 17.51 & 18.11 \\
    \multicolumn{2}{l}{VGGT~\cite{wang2025vggt}} & 12.16 & \snd{14.45} & \snd{15.80} & \snd{15.43} & \snd{16.99} & \snd{18.11} & \fst{20.99} & \fst{24.06} & \fst{26.84} \\
    \multicolumn{2}{l}{SEVA~\cite{zhou2025stable}} & \trd{13.01} & \trd{14.32} & \trd{15.16} & 14.11 & 15.65 & \trd{16.73} & \trd{16.59} & \trd{19.27} & \trd{23.09} \\
    \rowcolor[gray]{0.9}\multicolumn{2}{l}{\textbf{Ours}} & \fst{14.16} & \fst{15.52} & \fst{16.45} & \fst{16.03} & \fst{17.99} & \fst{19.32} & \snd{19.26} & \snd{22.31} & \snd{24.52} \\
    \bottomrule
    \end{tabular}
    }
    \end{subtable}
    
    \vspace{0.8mm}
    \begin{subtable}[t]{1.0\textwidth}

    \caption{\textbf{Pose error and chamfer distance $\downarrow$.}}
    \vspace{-3mm}
    \label{tab:long_traj_geo}
    \resizebox{\linewidth}{!}{%
    \setlength{\tabcolsep}{1.1pt}
    \renewcommand{\arraystretch}{1.3}
    \begin{tabular}{l *{9}{c}}
    \toprule
    Dataset ($P$)
    & \multicolumn{3}{c}{RE10K~\cite{zhou2018stereo} (3)}
    & \multicolumn{3}{c}{DL3DV~\cite{ling2024dl3dv} (3)}
    & \multicolumn{3}{c}{T\&T~\cite{Knapitsch2017} (3)} \\
    \cmidrule(lr){1-1}\cmidrule(lr){2-4}\cmidrule(lr){5-7}\cmidrule(lr){8-10}
    Metric ($\downarrow$)
    & $T_{\textit{err}}$ & $R_{\textit{err}}$ & $CD$
    & $T_{\textit{err}}$ & $R_{\textit{err}}$ & $CD$
    & $T_{\textit{err}}$ & $R_{\textit{err}}$ & $CD$ \\
    \midrule
    SEVA~\cite{zhou2025stable}
    & 1.62 & 5.84 & 1.35
    & 35.0 & 6.44 & 1.03
    & 6.08 & 2.87 & 1.42 \\
    \rowcolor[gray]{0.9} \textbf{Ours}
    & \textbf{0.76} & \textbf{4.83} & \textbf{0.19}
    & \textbf{28.3} & \textbf{5.13} & \textbf{0.62}
    & \textbf{3.56} & \textbf{1.52} & \textbf{0.92} \\
    \bottomrule
    \end{tabular}
    }
\end{subtable}
\end{minipage}
\vspace{-3mm}
\end{table}

%% file: table/geometry_metrics.tex
\begin{table*}[t]
\centering
\caption{\textbf{Geometry Fidelity of GeoNVS.} $T_\text{err}$, $R_\text{err}$, 
and $CD$ measure geometric fidelity. PSNR$_\text{V}$ and PSNR$_\text{U}$ denote 
synthesis quality on co-visible and non-co-visible regions w.r.t.\ the reference 
views. All the results are under the Trajectory NVS setting. ``Input-level injection'' denotes input-level integration of 3D Gaussian priors.}
\vspace{-3mm}
\label{tab:geometric_consistency}
\setlength{\tabcolsep}{5pt}
\renewcommand{\arraystretch}{1.2}
\resizebox{\textwidth}{!}{
\begin{tabular}{l ccccc ccccc ccccc}
\toprule
\multirow{3}{*}{Method}
& \multicolumn{5}{c}{Mip360 Dataset~\cite{barron2022mip}}
& \multicolumn{5}{c}{DL3DV Dataset~\cite{ling2024dl3dv}}
& \multicolumn{5}{c}{WRGBD-S$_h$ Dataset~\cite{xia2024rgbd}} \\
& \multicolumn{5}{c}{(3 reference views)}
& \multicolumn{5}{c}{(6 reference views)}
& \multicolumn{5}{c}{(3 reference views)} \\
\cmidrule(lr){2-6} \cmidrule(lr){7-11} \cmidrule(lr){12-16}
& $T_\text{err}$$\downarrow$ & $R_\text{err}$$\downarrow$ & $CD$$\downarrow$ & PSNR$_\text{V}$$\uparrow$ & PSNR$_\text{U}$$\uparrow$
& $T_\text{err}$$\downarrow$ & $R_\text{err}$$\downarrow$ & $CD$$\downarrow$ & PSNR$_\text{V}$$\uparrow$ & PSNR$_\text{U}$$\uparrow$
& $T_\text{err}$$\downarrow$ & $R_\text{err}$$\downarrow$ & $CD$$\downarrow$ & PSNR$_\text{V}$$\uparrow$ & PSNR$_\text{U}$$\uparrow$ \\
\midrule
\multicolumn{9}{l}{\textbf{Baselines}} \\
SEVA~\cite{zhou2025stable}
& \snd{18.76} & \snd{3.29} & \fst{0.07} & \snd{13.85} & \snd{13.01}
& \snd{19.68} & \snd{2.80} & \snd{0.57} & 15.68 & 16.90
& \snd{8.10} & \snd{8.86} & \snd{1.58} & \snd{15.53} & \snd{15.79} \\
\midrule
\multicolumn{9}{l}{\textbf{Generative model + VGGT~\cite{wang2025vggt} prior}} \\
Difix3D~\cite{wu2025difix3d+}
& \trd{219.39} & \trd{51.64} & \trd{0.68} & 12.92 & 11.82
& \trd{64.4} & \trd{9.66} & \trd{3.76} & \trd{16.11} & \trd{17.29}
& \trd{19.59} & \trd{32.48} & 10.11 & 11.07 & 11.36 \\
SEVA $+$ Input-level injection
& 233.03 & 99.79 & 1.02 & \trd{13.14} & \trd{12.03}
& 81.14 & 17.19 & 3.82 & \snd{16.76} & \snd{17.76}
& 29.27 & 52.51 & \trd{2.01} & \trd{11.36} & \trd{11.42} \\
\rowcolor[gray]{0.9}\textbf{Ours}
& \fst{16.24} & \fst{2.98} & \fst{0.07} & \fst{15.11} & \fst{14.07}\,{\small\color{red}(\textbf{+1.06})}
& \fst{13.27} & \fst{2.01} & \fst{0.41} & \fst{18.11} & \fst{18.98}\,{\small\color{red}(\textbf{+2.08})}
& \fst{4.00} & \fst{4.56} & \fst{1.29} & \fst{17.33} & \fst{17.51}\,{\small\color{red}(\textbf{+1.72})} \\
\bottomrule
\end{tabular}
}
\vspace{-6mm}
\end{table*}

%% file: figure/feature_analysis.tex
\begin{figure}[t]
\centering
\includegraphics[width=0.99\linewidth]{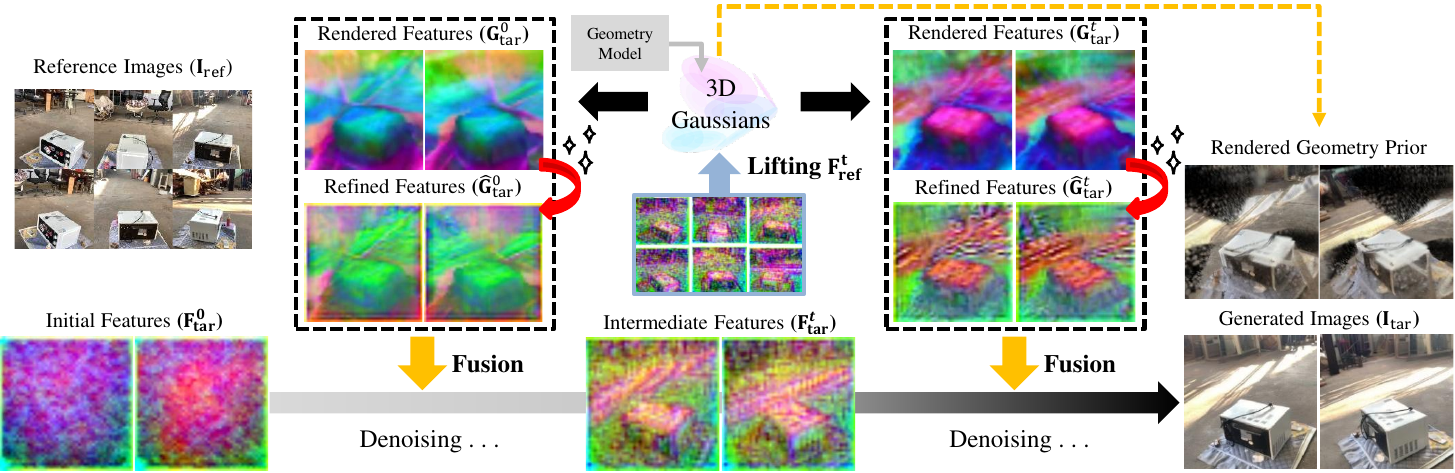}
\vspace{-3mm}
\caption{
\textbf{Feature modulation by GS-Adapter.} We visualize intermediate diffusion features during the denoising process. GS-Adapter consists of three stages: (1) \textbf{lifting} reference-view features $\mathbf{F}_\text{ref}^t$ into 3D Gaussians, (2) \textbf{refining} the novel-view features $\mathbf{G}_\text{tar}$ into $\hat{\mathbf{G}}_\text{tar}$, and (3) \textbf{fusing} $\hat{\mathbf{G}}_\text{tar}$ with $\mathbf{F}_\text{tar}^t$ to generate geometry-corrected outputs.}
\vspace{-3mm}
\label{fig:feature_analysis}
\end{figure}

%% file: table/ablation_geometry_priors.tex
\begin{table}[t]
\centering
\begin{minipage}[t]{0.49\textwidth}
\centering
\captionof{table}{\textbf{GeoNVS with SEVA~\cite{zhou2025stable}.}  We extend GeoNVS with various geometry priors in a zero-shot manner. All methods are evaluated with 3 reference images.}
\vspace{-2mm}
\centering
\label{tab:ablation_lrms1}
\resizebox{\linewidth}{!}{%
\setlength{\tabcolsep}{0.8pt}
\renewcommand{\arraystretch}{1.1}
\setlength{\aboverulesep}{2pt}
\setlength{\belowrulesep}{2pt}
\begin{tabular}{l c c c c c c c c}
\toprule
\multirow{2}{*}{Method}
& \multicolumn{2}{c}{DTU}
& \multicolumn{2}{c}{CO3D}
& \multicolumn{2}{c}{WRGBD-S$_e$}
& \multicolumn{2}{c}{T\&T} \\
\cmidrule(lr){2-3}
\cmidrule(lr){4-5}
\cmidrule(lr){6-7}
\cmidrule(lr){8-9}
& PSNR & SSIM & PSNR & SSIM & PSNR & SSIM & PSNR & SSIM \\
\midrule
\multicolumn{9}{l}{\textbf{Baseline}} \\
SEVA~\cite{zhou2025stable} & 17.03 & 0.66 & 17.22 & 0.55 & 15.96 & 0.53 & 18.85 & 0.61 \\
\midrule
\multicolumn{9}{l}{\textbf{Geometry priors}} \\
MVSplat~\cite{chen2024mvsplat} & \trd{11.35} & \trd{0.41} & 12.04 & 0.39 & \snd{12.89} & \snd{0.41} & 13.45 & 0.38 \\
DepthSplat~\cite{xu2025depthsplat} & \snd{14.19} & \snd{0.49} & \snd{15.08} & \snd{0.49} & \fst{13.92} & \fst{0.44} &\snd{17.99} & \snd{0.54} \\
VGGT~\cite{wang2025vggt} & \fst{17.15} & \fst{0.64} & \fst{15.39} & \fst{0.52} & \trd{11.09} & \trd{0.41} & \fst{23.96} & \fst{0.81} \\
Pi3~\cite{wang2025pi3} & 7.81 & 0.31 & \trd{12.60} & \trd{0.47} & 10.54 & 0.37 & \trd{16.98} & \trd{0.48} \\
\midrule
\multicolumn{9}{l}{\textbf{Ours w/ Naïve Fusion}} \\
$+$ MVSplat & \trd{17.04} & \trd{0.65} & \snd{18.79} & \snd{0.62} & \snd{17.62} & \trd{0.59} & 21.18 & 0.68 \\
$+$ DepthSplat & \snd{17.94} & \snd{0.68} & \fst{19.00} & \fst{0.63} & \fst{17.97} & \fst{0.61} & \snd{22.05} & \fst{0.71} \\
$+$ VGGT & \fst{18.89} & \fst{0.69} & \trd{18.60} & 0.60 & \trd{17.60} & \snd{0.59} & \fst{22.22} & \snd{0.70} \\
$+$ Pi3 & 14.83 & 0.59 & 18.59 & \trd{0.61} & 17.38 & 0.58 & \trd{21.92} & \trd{0.69} \\
\midrule
\multicolumn{9}{l}{\textbf{Ours w/ Adaptive Fusion}} \\
$+$ MVSplat & \snd{19.15} & 0.69 & \trd{19.23} & 0.63 & \fst{18.22} & 0.61 & \fst{22.24} & 0.69 \\
$+$ DepthSplat & \trd{19.14} & 0.69 & \snd{19.24} & 0.63 & \fst{18.22} & 0.61 & \fst{22.24} & 0.69 \\
$+$ VGGT & \fst{19.18} & 0.69 & \fst{19.26} & 0.63 & \trd{18.21} & 0.61 & \fst{22.24} & 0.69 \\
$+$ Pi3 & 19.02 & 0.69 & 19.22 & 0.63 & 18.19 & 0.61 & 22.22 & 0.69 \\
\bottomrule
\end{tabular}
}
\end{minipage}
\hfill
\begin{minipage}[t]{0.49\textwidth}
\centering
\captionof{table}{\textbf{PSNR $\uparrow$ on GeoNVS with CameraCtrl~\cite{he2025cameractrl}.} GeoNVS built on CameraCtrl generalizes to various geometry priors in a zero-shot manner, without additional training. Following the original CameraCtrl setting, all methods are evaluated using a single reference image.}
\vspace{-2mm}
\label{tab:ablation_lrms2}
\resizebox{\linewidth}{!}{%
\setlength{\tabcolsep}{0.9pt}
\renewcommand{\arraystretch}{1.2}
\setlength{\aboverulesep}{2pt}
\setlength{\belowrulesep}{2pt}
\begin{tabular}{l c c c c c c c c c c c c}
\toprule
\multirow{2}{*}{Method}
& dataset
& \multicolumn{2}{c}{CO3D}
& DTU
& RE10K
& Mip360
& WRGBD & \multirow{2}{*}{Average} \\
\cmidrule(lr){2-2}
\cmidrule(lr){3-4}
\cmidrule(lr){5-5}
\cmidrule(lr){6-6}
\cmidrule(lr){7-7}
\cmidrule(lr){8-8}
& split & V & R & R & R & R & S$_h$ \\
\midrule
\multicolumn{9}{l}{\textbf{Baseline}} \\  
\multicolumn{2}{l}{CameraCtrl~\cite{he2025cameractrl}} & 13.23 & 11.71 & 8.72 & 14.09 & 10.94 & 10.38 & 11.14 \\
\midrule
\multicolumn{9}{l}{\textbf{Geometry priors}} \\  
\multicolumn{2}{l}{MVSplat~\cite{chen2024mvsplat}} & 10.12 & \snd{9.14} & \trd{7.37} & \trd{11.26} & 8.55 & \trd{6.48} & \trd{8.82} \\
\multicolumn{2}{l}{DepthSplat~\cite{xu2025depthsplat}} & \snd{11.29} & \trd{8.54} & \snd{8.79} & \fst{14.49} & \snd{9.20} & \snd{7.31} & \snd{9.94} \\
\multicolumn{2}{l}{VGGT~\cite{wang2025vggt}} & \trd{10.48} & 8.43 & 7.04 & \snd{13.94} & \trd{8.89} & 6.27 & 8.67 \\
\multicolumn{2}{l}{Pi3~\cite{wang2025pi3}} & \fst{12.08} & \fst{11.80} & \fst{8.82} & 11.00 & \fst{10.60} & \fst{8.86} & \fst{10.16} \\
\midrule
\multicolumn{9}{l}{\textbf{Ours w/ Naïve Fusion}} \\  
\multicolumn{2}{l}{$+$ MVSplat} & \snd{13.76} & \snd{12.69} & 7.71 & \snd{14.44} & \trd{11.19} & \trd{10.63} & \trd{11.74} \\
\multicolumn{2}{l}{$+$ DepthSplat} & \fst{14.64} & \trd{12.15} & \fst{11.00} & \fst{16.85} & \snd{11.27} & \snd{10.88} & \fst{12.80} \\
\multicolumn{2}{l}{$+$ VGGT} & 13.30 & 11.69 & \trd{8.68} & \trd{14.08} & \fst{11.30} & 10.55 & 11.60 \\
\multicolumn{2}{l}{$+$ Pi3} & \trd{13.60} & \fst{12.99} & \snd{8.99} & 13.35 & \fst{11.30} & \fst{10.97} & \snd{11.87} \\
\bottomrule
\end{tabular}
}  
\end{minipage}
\vspace{-6mm}
\end{table}

%% file: table/ablation_model.tex
\begin{table*}[t]
\centering
\scriptsize
\caption{\textbf{Model ablation study of GeoNVS.} We ablate the key components of GS-Adapter, including feature refinement, feature fusion strategy, and multi-scale fusion. All experiments use VGGT~\cite{wang2025vggt} as the geometry prior. Here, feats abbreviates features.}
\vspace{-3mm}
\label{tab:ablation_model}
\renewcommand{\arraystretch}{0.8}
\resizebox{\textwidth}{!}{
\begin{tabular}{lcccccccccc}
\toprule
\multirow{3}{*}{\textbf{Method}} & \multicolumn{4}{c}{\textbf{GS-Adapter}} & \multicolumn{2}{c}{\textbf{Mip360}} & \multicolumn{2}{c}{\textbf{DL3DV}} & \multicolumn{2}{c}{\textbf{T\&T}} \\
\cmidrule(lr){2-5} \cmidrule(lr){6-7} \cmidrule(lr){8-9} \cmidrule(lr){10-11}
 & \multicolumn{2}{c}{Refinement} & \multicolumn{2}{c}{Fusion} & \multicolumn{2}{c}{6-view} & \multicolumn{2}{c}{6-view} & \multicolumn{2}{c}{3-view} \\
\cmidrule(lr){2-3} \cmidrule(lr){4-5} \cmidrule(lr){6-7} \cmidrule(lr){8-9} \cmidrule(lr){10-11} & $\mathcal{L}_\text{feat}$ & GS-PE & Na\"{i}ve & Adaptive & PSNR$\uparrow$ & SSIM$\uparrow$ & PSNR$\uparrow$ & SSIM$\uparrow$ & PSNR$\uparrow$ & SSIM$\uparrow$ \\
\midrule
\multicolumn{5}{l}{\textbf{Geometry prior}} \\
VGGT~\cite{wang2025vggt} & & & & & 15.49 & 0.362 & 17.09 & 0.52 & 23.96 & 0.81 \\
\midrule
\multicolumn{5}{l}{\textbf{Baseline}} \\
SEVA~\cite{zhou2025stable} & & & & & 15.38 & 0.339 & 15.84 & 0.45 & 18.85 & 0.61 \\
\midrule
$+$ Input-level injection & & & &  & 15.40 & 0.373 & 16.73 & 0.52 & \fst{23.05} & \fst{0.74}  \\
$+$ LoRA only       &            &            &            &            & 15.55 & 0.360 & 17.35 & \trd{0.54} & 21.41 & 0.68 \\
\midrule
\multirow{3}{*}{$+$ GS-Adapter (Encoder-1,2,3 feats)} & \cmark & & \cmark & & 16.36 & \snd{0.378} & 17.80 & \snd{0.55} & 21.67 & \trd{0.69} \\
                               & & \cmark & \cmark & & \trd{16.47} & \snd{0.378} & \trd{17.88} & \snd{0.55} & 22.11 & \trd{0.69} \\
                               & \cmark & \cmark & \cmark & & \snd{16.48} & \fst{0.383}
                               & \snd{17.95} & \fst{0.56} & \trd{22.22} & \snd{0.70} \\
\midrule
\multicolumn{5}{l}{\textbf{Multi-scale fusion}} \\
$+$ GS-Adapter (Encoder-1 feat) & \cmark & \cmark & \cmark & & 16.28 & \trd{0.376} & 17.49 & 0.53 & 21.36 & 0.68 \\
$+$ GS-Adapter (Encoder-1,2 feats) & \cmark & \cmark & \cmark & & 16.31 & 0.370 & 17.53 & 0.53 & 21.73 & 0.68 \\
$+$ GS-Adapter (Encoder-1,2,3 feats) & \cmark & \cmark & \cmark & & \snd{16.48} & \fst{0.383}
                               & \snd{17.95} & \fst{0.56} & \trd{22.22} & \snd{0.70} \\
\midrule
\multicolumn{5}{l}{\textbf{Feature Refinement}} \\
$-$ RefineNet, $\mathcal{R}$ &  &  &  & \cmark & 16.23 & 0.376 & 17.71 & 0.45 & 21.91 & \trd{0.69} \\
\midrule
\rowcolor[gray]{0.9}\textbf{Ours (Encoder-1,2,3 feats)} & \cmark & \cmark && \cmark & \fst{16.57} & \fst{0.383} & \fst{18.11} & \fst{0.56} & \snd{22.24} & \trd{0.69} \\
\bottomrule
\end{tabular}
}
\vspace{-9mm}
\end{table*}

%% file: figure/qualitative1.tex
\begin{figure}[t]
\centering
\includegraphics[width=0.97\linewidth]{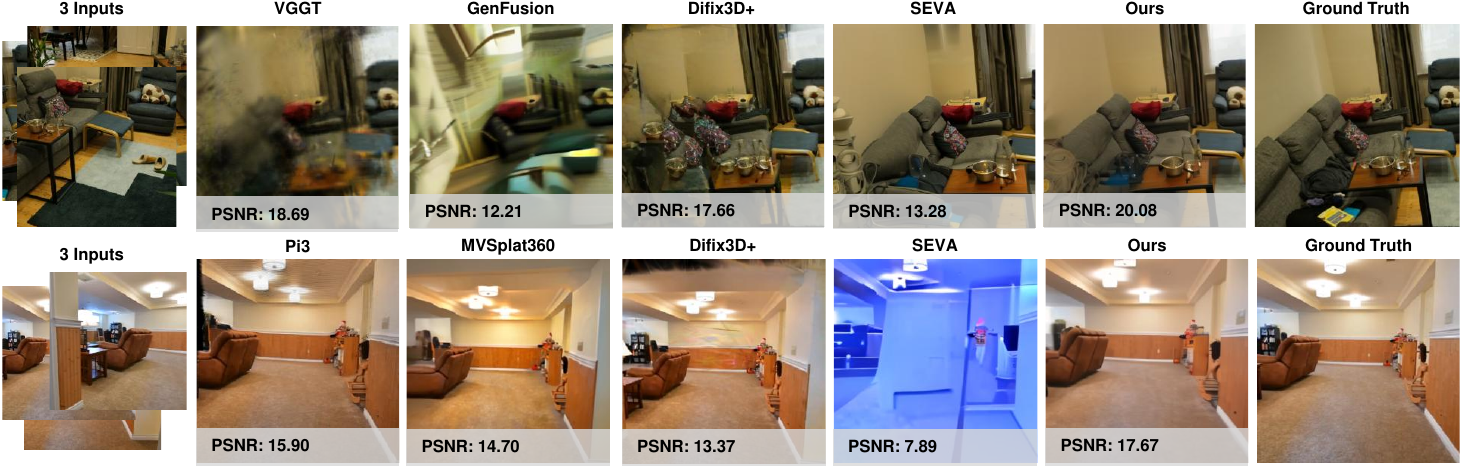}
\vspace{-3mm}
\caption{
\textbf{Qualitative results of GeoNVS with SEVA~\cite{zhou2025stable}.}}
\vspace{-4mm}
\label{fig:qual1}
\end{figure}

%% file: figure/qualitative3.tex
\begin{figure}[t]
\centering
\includegraphics[width=0.97\linewidth]{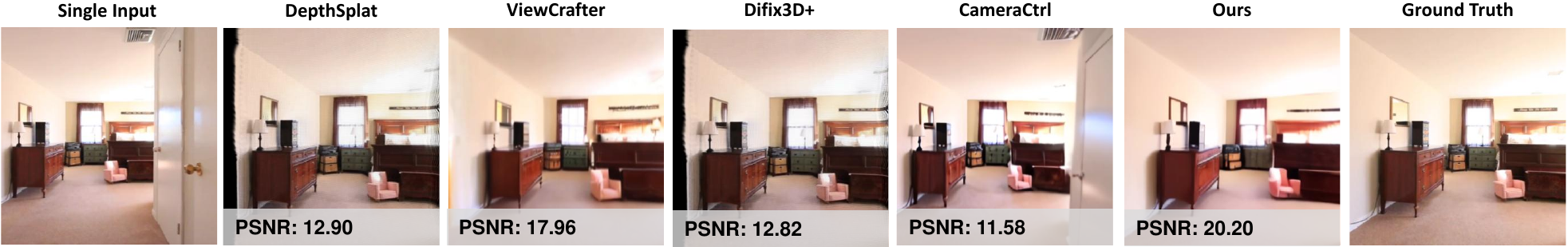}
\vspace{-2mm}
\caption{
\textbf{Qualitative results of GeoNVS with CameraCtrl~\cite{he2025cameractrl}}}
\vspace{-5mm}
\label{fig:qual3}
\end{figure}

%% file: figure/gaussian_pruning.tex
\begin{figure}[!htbp]
\centering
\begin{subfigure}[t]{0.49\textwidth}
    \centering
    \includegraphics[width=\textwidth]{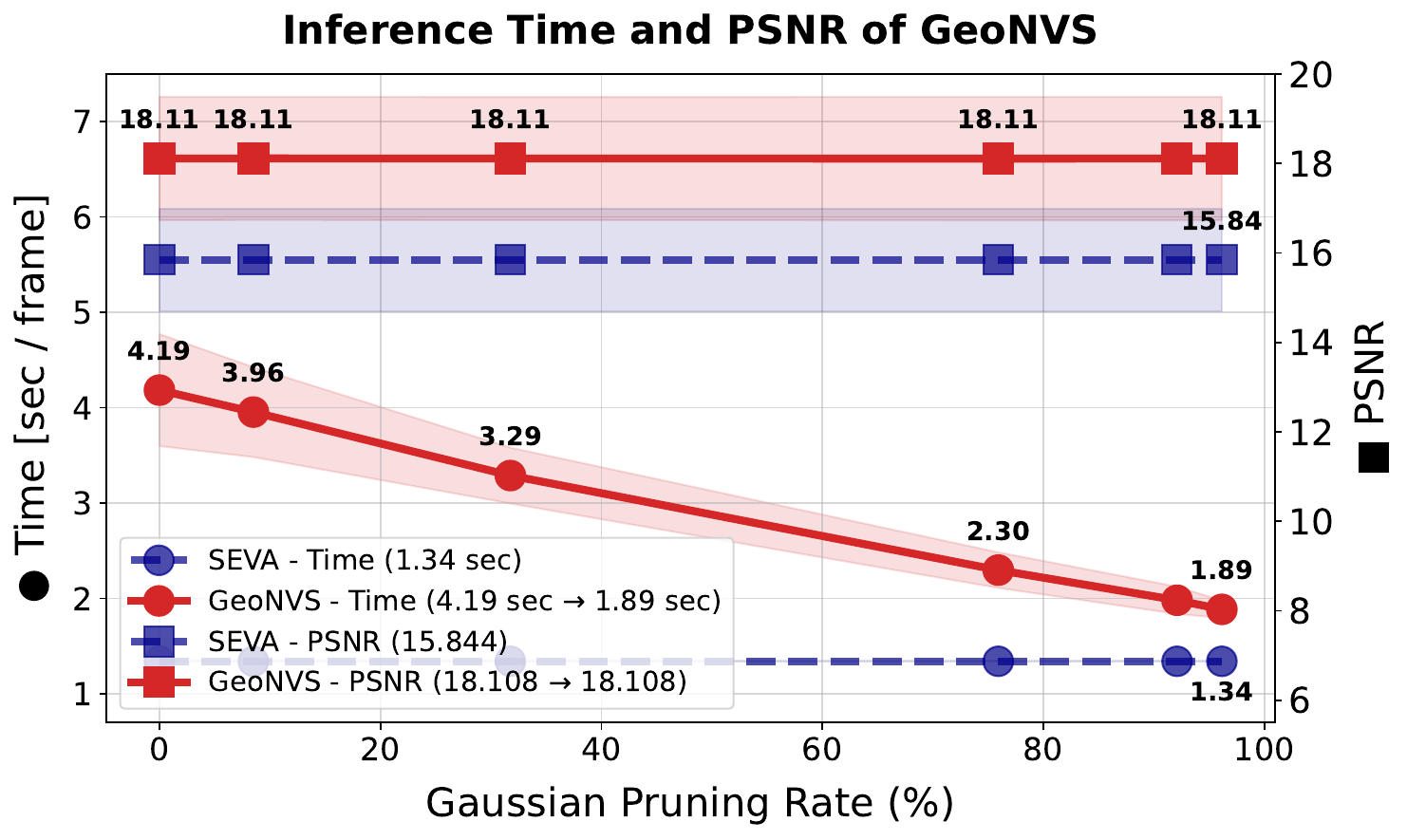}
    \label{fig:runtime_performance}
\end{subfigure}
\begin{subfigure}[t]{0.49\textwidth}
    \centering
    \includegraphics[width=\textwidth]{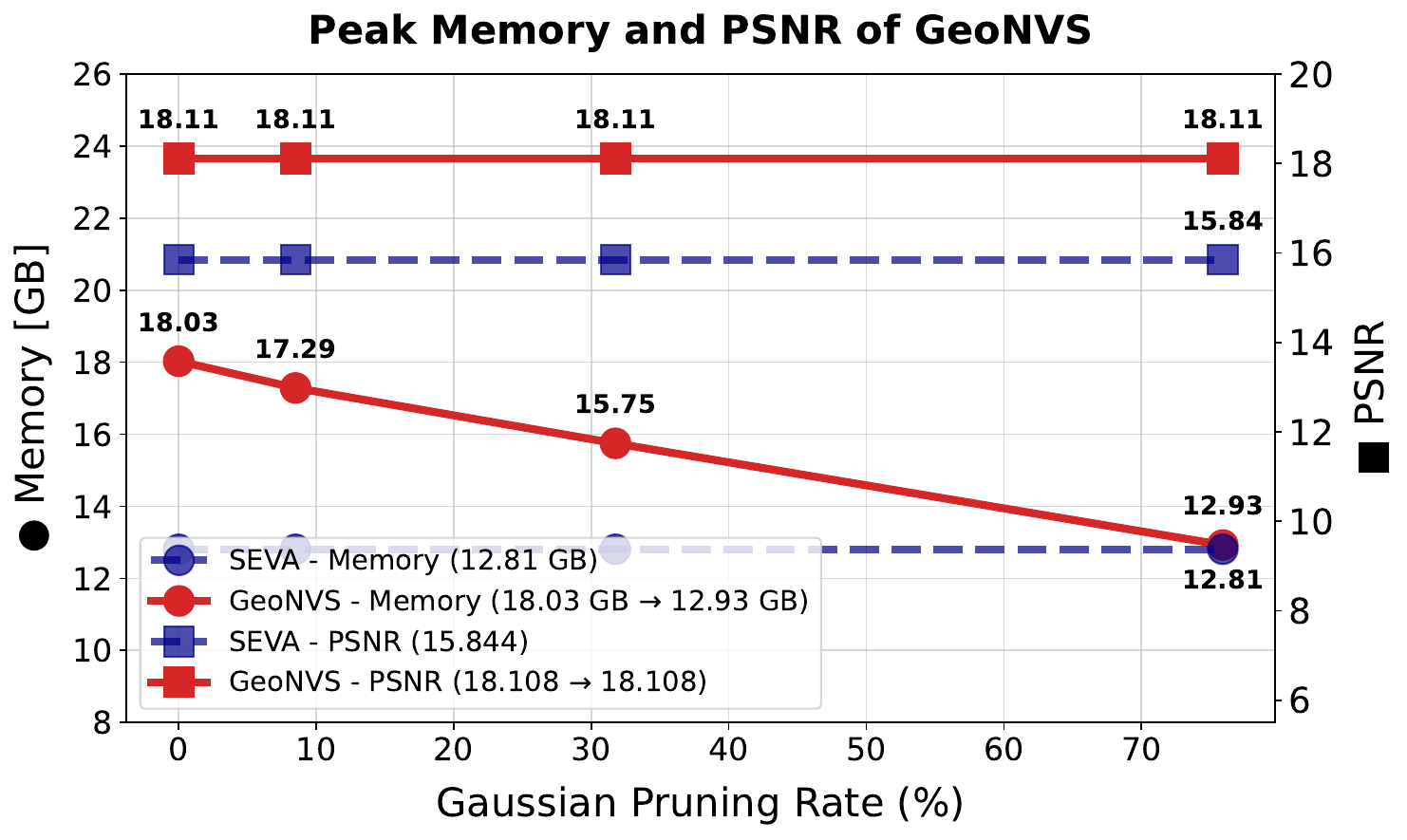}
    \label{fig:memory_performance}
\end{subfigure}

\vspace{-7mm}
\caption{\textbf{Runtime and memory analysis under Gaussian pruning.} 
Inference time, peak GPU memory, and PSNR of GeoNVS versus Gaussian pruning rate across 10 DL3DV~\cite{ling2024dl3dv} scenes. Voxel-based pruning~\cite{jiang2025anysplat} yields a 2.22$\times$ speedup at 96.2\% pruning. Peak memory drops to 12.93 GB, close to SEVA~\cite{zhou2025stable}, while PSNR is preserved.}
\vspace{-8mm}
\label{fig:performance}
\end{figure}

%% file: figure/qualitative_geo.tex
\begin{figure}[t]
\centering
\includegraphics[width=0.99\linewidth]{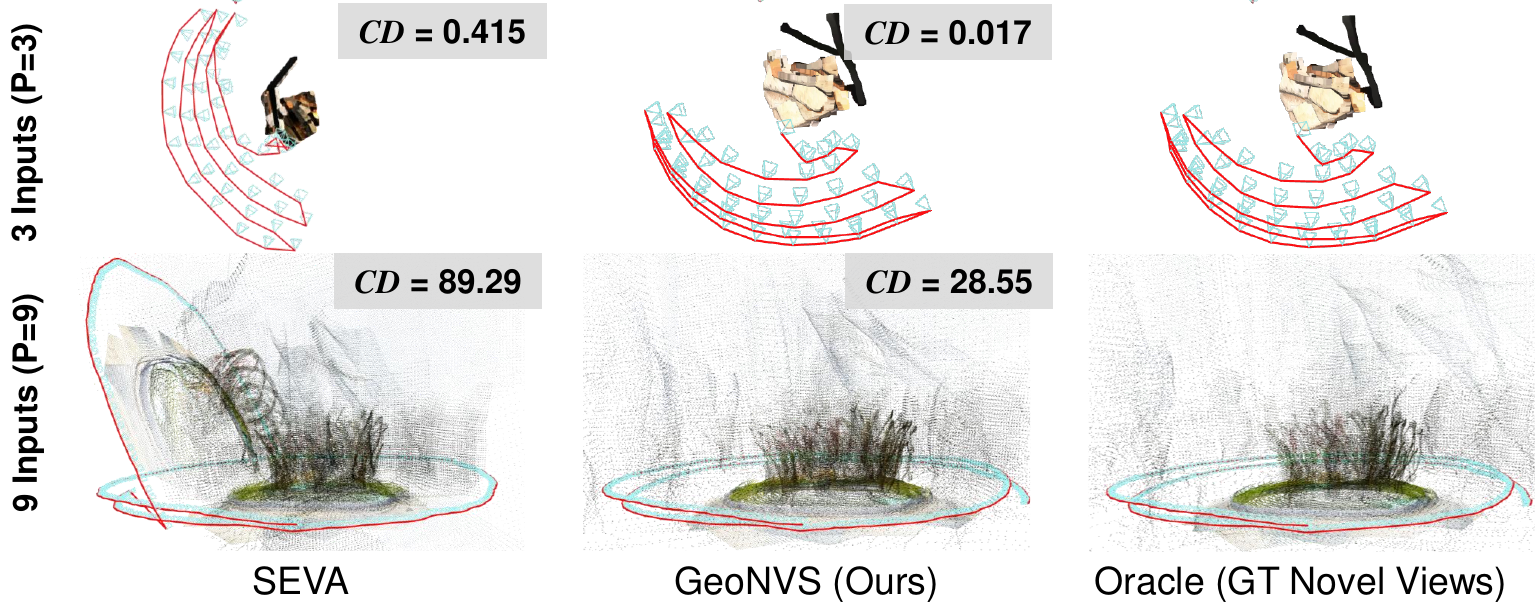}
\vspace{-3mm}
\caption{
\textbf{Qualitative results of GeoNVS.} 3D reconstruction and {\color{SkyBlue} camera trajectory} estimated from the generated video, showing the improved geometric fidelity of GeoNVS.}
\vspace{-3mm}
\label{fig:qual4}
\end{figure}

%% file: figure/limitation.tex
\begin{figure}[t]
\centering
\includegraphics[width=0.99\linewidth]{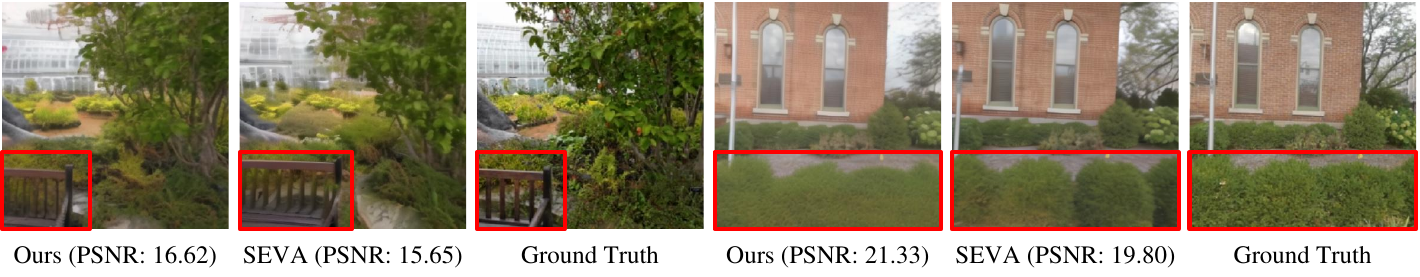}
\vspace{-3mm}
\caption{\textbf{Limitation: blurry textures.}
Our method preserves structure well but occasionally produces blurry textures in fine-detail regions (\textcolor{red}{red boxes}).}
\vspace{-5mm}
\label{fig:limitation_cases}
\end{figure}

%% file: sec/5_conclusion.tex
\section{Conclusions}
\label{sec:conclusion}
\vspace{-3mm}

We present GeoNVS, a geometry-grounded novel-view synthesizer that enhances geometric fidelity and camera controllability using GS-Adapter. Unlike prior methods that inject geometry at the input level, GS-Adapter modulates internal diffusion features using 3D Gaussian priors in a plug-and-play manner, enabling zero-shot compatibility with any geometry model and diffusion backbone. Extensive experiments across 9 scenes and 18 settings demonstrate state-of-the-art performance, achieving 11.3\% and 14.9\% improvements over SEVA and CameraCtrl, with up to $2\times$ reduction in translation error and $7\times$ in Chamfer Distance. GeoNVS improves synthesis in non-co-visible regions by up to 2.08 dB, confirming that feature-level 3D grounding facilitates extrapolation beyond observed regions. Finally, GeoNVS supports pose-free scenarios via SfM methods~\cite{huang2025vipe,fu2024colmap} and reduces inference overhead by $2.2\times$ with negligible performance degradation. 

\noindent\textbf{Limitations and future work.}
While our proposed method peforms well in sparse-view and long-range trajectory NVS, its generation ability degrades in regions distant from input views. We attribute this to increasingly uncertain feature rendering in GS-Adapter as the distance from input views grows. Our method occasionally produces blurry textures due to information loss during feature lifting, which can reduce perceptual sharpness despite well-preserved structure (\cref{fig:limitation_cases}). We leave addressing these limitations to future work.

\noindent\textbf{Acknowledgements.} This work was supported by the InnoCORE program of the Ministry of Science and ICT (26-InnoCORE-01), and by the National Research Foundation of Korea (NRF) grant funded by the Korea government (MSIT) (RS-2026-25473963).

%% file: Supplementary_Material/6_supple.tex
\newpage
\section*{Supplementary Material}
\label{sec:supple_overview}
\vspace{-2mm}
This supplementary material is organized as follows. 
\vspace{-2mm}
\begin{itemize}
    \item \cref{sec:supple_notation} summarizes the mathematical notation used throughout the paper.
    \item \cref{sec:supple_eval_dataset} provides details on dataset splits and comparison methods.
    \item \cref{sec:supple_additional_exp} presents additional experiments, including further analysis and ablations on input-level injection strength, feature uplifting, feature refinement, and adaptive fusion gating.
    \item \cref{sec:supple_arch} describes the detailed architectures of the multi-scale fusion module and RefineNet.
    \item \cref{sec:supple_data_preprocess} explains the training dataset preprocessing pipeline.
    \item \cref{sec:supple_additional_quant} provides additional quantitative results in SSIM and LPIPS.
    \item \cref{sec:supple_qualitative} presents additional qualitative results across diverse settings.
    \item \cref{sec:supple_limitations} discusses limitations and failure cases of GeoNVS.
\end{itemize}
\vspace{-7mm}

\section{\normalsize Notation Table}
\label{sec:supple_notation}
\vspace{-3mm}
~\cref{tab:notation} summarizes the mathematical notation used in the paper by category.

\vspace{-4mm}
\section{\normalsize Detail of Evaluation Benchmark}
\label{sec:supple_eval_dataset}
\vspace{-3mm}
We evaluate GeoNVS across 9 scenes and 18 settings drawn from 9 public 
benchmarks following SEVA~\cite{zhou2025stable}: OmniObject3D~\cite{wu2023omniobject3d}, 
RealEstate10K~\cite{zhou2018stereo}, LLFF~\cite{mildenhall2019local}, 
DTU~\cite{jensen2014large}, CO3D~\cite{reizenstein2021common}, 
WRGBD~\cite{xia2024rgbd}, Mip-NeRF 360~\cite{barron2022mip}, 
DL3DV~\cite{ling2024dl3dv}, and Tanks and Temples~\cite{Knapitsch2017}. 
We detail the dataset splits and evaluation protocols of comparison methods 
in~\cref{subsec:supple_dataset_split,subsec:supple_evaluation_method}.

\vspace{-5mm}
\subsection{\normalsize Dataset Splits}
\label{subsec:supple_dataset_split}
\vspace{-2mm}
\begin{sloppypar}
The data splits used in the evaluation tables include 
ReconFusion~\cite{wu2024reconfusion} (R), 
4DiM~\cite{watson2024controlling} (D), 
SEVA~\cite{zhou2025stable} (S), 
pixelSplat~\cite{charatan2024pixelsplat} (P), 
ViewCrafter~\cite{yu2024viewcrafter} (V), 
LongLRM~\cite{ziwen2024llrm} (L), 
and our own split adapted from CF-3DGS~\cite{fu2024colmap} (O).
\end{sloppypar}

\vspace{-5mm}
\subsection{\normalsize Evaluation Details of Comparison Methods}
\label{subsec:supple_evaluation_method}
\vspace{-1mm}
\noindent\textbf{MVSplat~\cite{chen2024mvsplat}, HiSplat~\cite{tang2024hisplat}, DepthSplat~\cite{xu2025depthsplat}.}
We use these feed-forward methods to directly regress Gaussian splats $\mathcal{G}$ from the input-view images $\mathbf{I}_{\text{ref}}$. These methods require posed images (images with corresponding poses $\boldsymbol{\pi}$).

\newpara{VGGT~\cite{wang2025vggt}, Pi3~\cite{wang2025pi3}.}
Using the point cloud and camera parameters predicted by the feed-forward methods, we initialize the Gaussian splats $\mathcal{G}$ directly from the point cloud. We then apply the co-visibility-based redundancy elimination strategy of InstantSplat~\cite{fan2024instantsplat} to prune overlapping regions, and further optimize the 3D-GS representation \cite{kerbl3Dgaussians} with the reference view images $\mathbf{I}_{\text{ref}}$. The scale ambiguity between the predicted camera poses and the ground-truth point cloud is resolved using~\cref{eq:scale_optimize}.

\newpara{ViewCrafter~\cite{yu2024viewcrafter}.}
In the original implementation, ViewCrafter operates in an exploratory manner without any knowledge of target viewpoints, relying on dust3r~\cite{wang2024dust3r} to estimate camera parameters and iteratively expanding the point cloud with generated frames. For evaluation, we instead provide ground-truth camera poses and use $P$ reference views alongside 5 additional viewpoints for geometry estimation, then render along the ground-truth trajectory before passing to the diffusion model. Note that this setting provides an advantage over the original by eliminating camera estimation error, and thus can be regarded as an upper-bound approximation of ViewCrafter's performance.

\newpara{MotionCtrl~\cite{wang2024motionctrl}.}
We employ only the CMCM module from MotionCtrl for camera control, following the original inference protocol with 25 denoising steps and the same classifier-free guidance scale. All videos are generated at $576 \times 1024$ resolution and resized to $384 \times 384$ for evaluation.

\newpara{CameraCtrl~\cite{he2025cameractrl}.}
We use the SVD~\cite{blattmann2023stable} based CameraCtrl officially provided by the author. The inference process follows the original paper, using 25 steps and the same conditional-guidance scale. All videos are generated at a resolution of $320 \times 576$ and subsequently resized to $384 \times 384$ for evaluation.

\newpara{SEVA~\cite{zhou2025stable}.}
We adopt the officially released v1.0 model provided by the authors; our GeoNVS framework is similarly built upon SEVA v1.0. Inference is performed following the settings of the original paper, using 50 denoising steps with a CFG scale of 2.0. All videos are rendered at a resolution of $384 \times 384$.

\newpara{MVSplat360~\cite{chen2024mvsplat360}.}
We first render novel-view features from MVSplat and use these features as input to generate novel-view images using SVD, following the original implementation.

\newpara{GenFusion~\cite{wu2025genfusion}.}
Rather than adopting GenFusion’s direct 3D reconstruction pipeline, we first leverage geometry priors from a regression-based model to render novel-view RGB, depth (RGBD) images. These noisy RGBD predictions are then fed into GenFusion for refinement. All videos are generated at a resolution of $320 \times 512$ and subsequently resized to $384 \times 384$ for evaluation.

\newpara{Difix3D~\cite{wu2025difix3d+}.}
We utilize the Difix3D model by providing a reference image together with the novel-view image rendered by the geometry prior (regression-based model). The reference image is chosen from the input-view set $\textbf{I}_{\text{ref}}$ based on the camera-center distance to the target novel-view viewpoint.

\subsection{\normalsize Geometry prior used for combined methods}
\label{subsec:supple_geometry_prior}
\vspace{-2mm}
For fair comparison, all combined methods~\cite{wu2025genfusion, wu2025difix3d+} and ours use the best-performing geometry prior per dataset in~\cref{tab:method_comparison}.~\cref{tab:geometry_prior_smallnvs} lists the geometry priors used in each benchmark for small-viewpoint NVS. Note that MVSplat360~\cite{chen2024mvsplat360} is an exception, as it is limited to MVSplat~\cite{chen2024mvsplat} by design.
\vspace{-4mm}
\begin{table}[ht]
\centering
\scriptsize
\caption{Geometry prior used for combined methods of small-viewpoint NVS in~\cref{tab:method_comparison}.}
\vspace{-2mm}
\label{tab:geometry_prior_smallnvs}
\begin{tabular}{lcc}
\toprule
\textbf{Benchmark} & \textbf{Input number ($P$)} & \textbf{Geometry prior} \\
\midrule
OO3D~\cite{wu2023omniobject3d}-S & 3 & DepthSplat~\cite{xu2025depthsplat} \\
RE10K~\cite{zhou2018stereo}-D & 1 & DepthSplat~\cite{xu2025depthsplat} \\
RE10K~\cite{zhou2018stereo}-P & 2 & DepthSplat~\cite{xu2025depthsplat} \\
RE10K~\cite{zhou2018stereo}-R & 1 & DepthSplat~\cite{xu2025depthsplat} \\
RE10K~\cite{zhou2018stereo}-R & 3 & Pi3~\cite{wang2025pi3} \\
LLFF~\cite{mildenhall2019local}-R & 1, 3 & VGGT~\cite{wang2025vggt} \\
DTU~\cite{jensen2014large}-R & 1 & Pi3~\cite{wang2025pi3} \\
DTU~\cite{jensen2014large}-R & 3 & VGGT~\cite{wang2025vggt} \\
CO3D~\cite{reizenstein2021common}-V & 1 & Pi3~\cite{wang2025pi3} \\
CO3D~\cite{reizenstein2021common}-R & 3 & VGGT~\cite{wang2025vggt} \\
WRGBD~\cite{xia2024rgbd}-S$_e$ & 3 & DepthSplat~\cite{xu2025depthsplat} \\
WRGBD~\cite{xia2024rgbd}-S$_h$ & 6 & DepthSplat~\cite{xu2025depthsplat} \\
Mip360~\cite{barron2022mip}-R & 6 & VGGT~\cite{wang2025vggt} \\
DL3DV~\cite{ling2024dl3dv}-S & 6 & VGGT~\cite{wang2025vggt} \\
DL3DV~\cite{ling2024dl3dv}-L & 9 & VGGT~\cite{wang2025vggt} \\
T\&T~\cite{Knapitsch2017}-O & 2, 3 & VGGT~\cite{wang2025vggt} \\
\bottomrule
\end{tabular}
\end{table}
\vspace{-4mm}

\newpage


\section{\normalsize Additional Experiments}
\label{sec:supple_additional_exp}
\vspace{-1mm}
\subsection{\normalsize Injection Strength in Input-Level Integration}
\label{subsec:supple_strength}
\vspace{-2mm}
To ensure a fair comparison with input-level injection approaches, we analyze performance across different inpaint strength levels $s$, where smaller $s$ enforces stronger adherence to the geometry prior and larger $s$ approaches the purely \setlength{\columnsep}{6pt}
\begin{wrapfigure}{r}{0.7\columnwidth}
\vspace{-9mm}
\centering
\includegraphics[width=0.7\columnwidth]{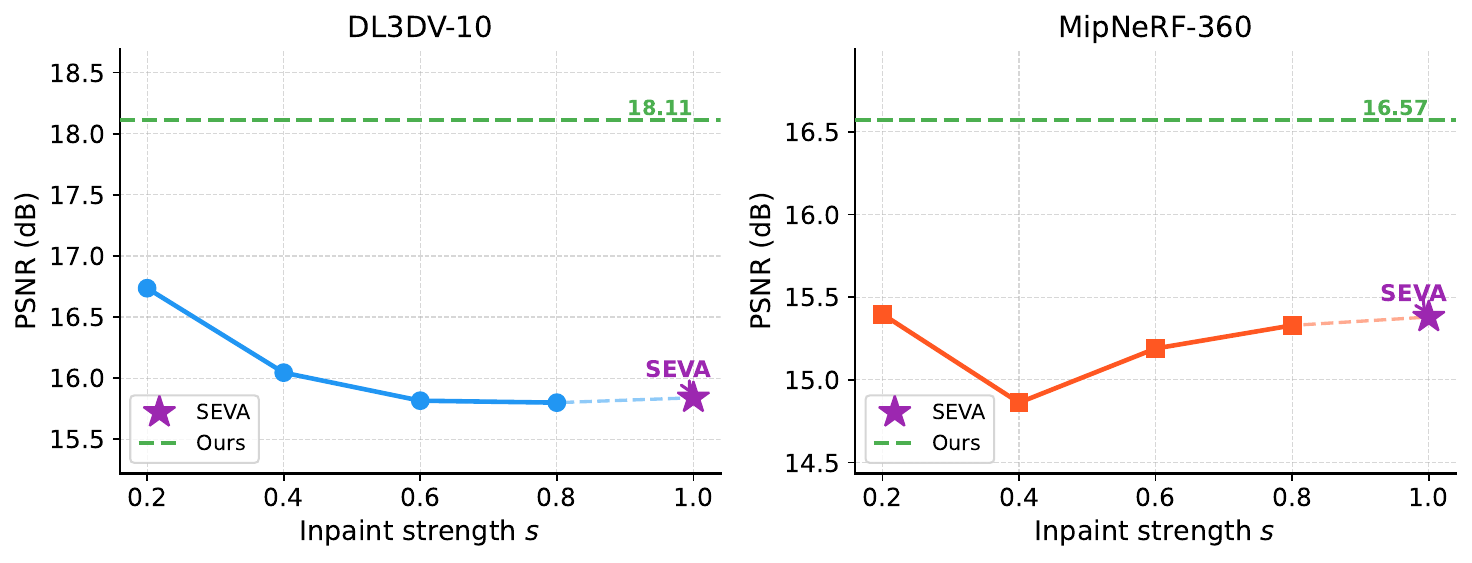}
\captionsetup{labelformat=empty, skip=0pt}
\caption{}
\label{fig:inpaint_ablation}
\vspace{-\baselineskip}
\vspace{-10mm}
\end{wrapfigure}generative baseline~\cite{zhou2025stable}.
As shown in Fig.~\ref{fig:inpaint_ablation}, input-level injection performs best at small $s$ on DL3DV-10K, where the geometry prior dominates, but degrades monotonically as $s$ increases toward the generative baseline. On MipNeRF-360, performance is unstable across all $s$ values, suggesting that noisy rasterized colors corrupt the structural signal regardless of inpaint strength. In contrast, GS-Adapter consistently outperforms input-level injection across all $s$ values on both datasets, demonstrating that feature-level modulation yields more stable and consistently superior results regardless of the geometry prior strength.

\vspace{-5mm}
\subsection{\normalsize Ablation on Feature Uplifting: Point Cloud vs 3D-GS}
\label{subsec:supple_ablation_featup}
\vspace{-2mm}
In the feature uplifting process, restricting each ray to consider only the single Gaussian with the highest rendering weight reduces the process to a \textbf{hard assignment} scheme, in which each pixel's feature is attributed solely to the most opaque surface point along that ray, which is mathematically equivalent to projecting features onto a point cloud. In contrast, the \textbf{soft assignment} scheme in~\cref{eq:uplifting} aggregates all Gaussians along a ray as a rendering-weight-proportional weighted sum, following the same principle as alpha-compositing to continuously accumulate features across the volume. This formulation mitigates the fragility of hard assignment in regions where multiple Gaussians partially contribute to a single pixel, as a single dominant Gaussian may fail to capture the full structural signal of the scene. 
Since Gaussians with moderate rendering weights are entirely discarded under hard assignment, soft assignment ensures \setlength{\columnsep}{6pt}
\begin{wrapfigure}{r}{0.6\columnwidth}
\vspace{-8mm}
\centering
\includegraphics[width=0.6\columnwidth]{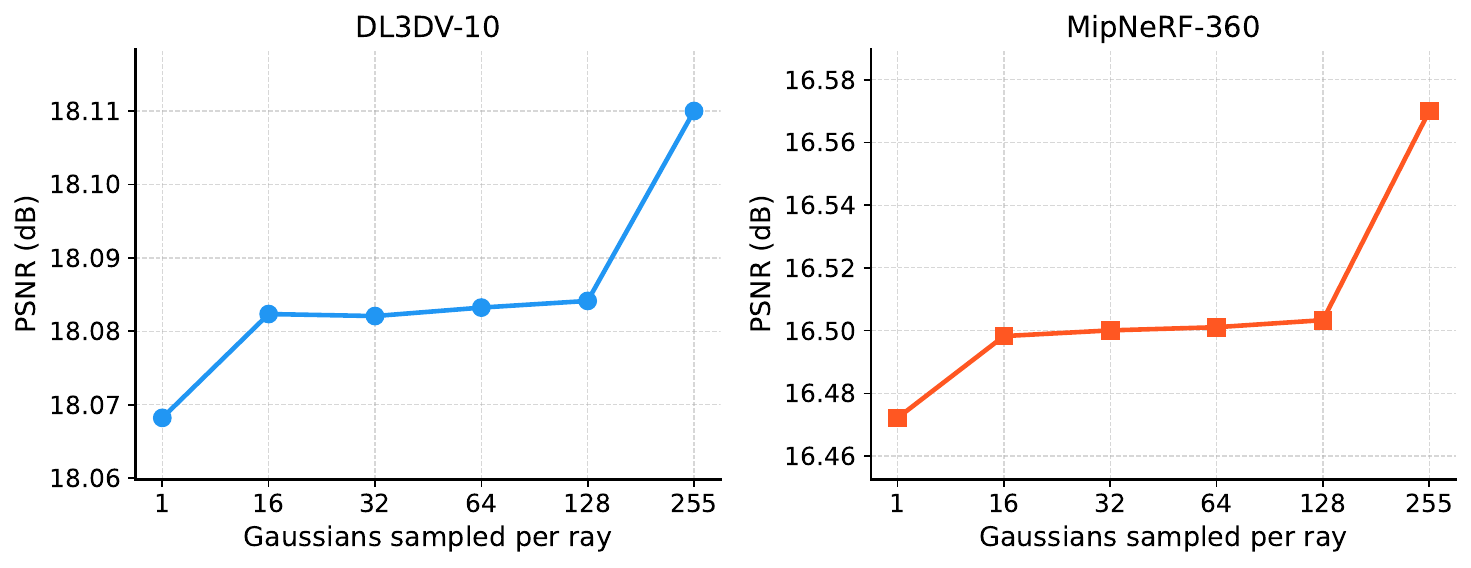}
\captionsetup{labelformat=empty, skip=0pt}
\caption{}
\label{fig:top-k}
\vspace{-\baselineskip}
\vspace{-8mm}
\end{wrapfigure}that all contributing Gaussians accumulate feature information proportional to their geometric relevance, yielding a smoother and more consistent feature field upon rasterization into novel views and ultimately improving both geometric consistency and synthesis quality. We validate this in \hyperref[fig:top-k]{Figure above} by varying $k \in \{1, 16, 32, 64, 128, 255\}$ on Mip-NeRF~360 and DL3DV-10K, where $k{=}1$ and $k{=}255$ correspond to hard and full soft assignment, respectively. PSNR consistently improves with increasing $k$ on both datasets, confirming the effectiveness of soft assignment.

\vspace{-4mm}
\subsection{\normalsize Validity of Feature Splatting and Refinement}
\label{subsec:supple_ablation_refinement}
\vspace{-1mm}
Feature splatting in the latent domain~\cite{zhou2024feature} implicitly assumes that diffusion latent features share the linear additive properties of RGB colors. However, certain feature channels may not correlate with geometric opacity, leading to a potential gap in the rendered features. Unlike prior works~\cite{drsplat25,marrie2025ludvig} that apply feature splatting to discrete tasks such as segmentation, novel-view synthesis is particularly sensitive to this gap, as any feature distortion directly degrades perceptual output quality. To mitigate this, RefineNet is trained with the feature refinement loss (\cref{eq:feature_loss}) to recover fidelity lost during compositing.
Beyond the ablation in~\cref{tab:ablation_model}, we further evaluate the effect of refinement by comparing \textbf{(a) lifting-rendering (w/o refinement)} against \textbf{(b) lifting-rendering-refinement}, with per-scene optimization applied to the VAE-encoded diffusion latent space. We report PSNR and SSIM between the decoded novel-view images and the ground truth across 540 randomly sampled scenes from the DL3DV~\cite{ling2024dl3dv} test set. As shown in the~\hyperref[fig:supple_refinement]{figure below}, refinement consistently improves both metrics across all input-view settings, with gains widening as the number of input views increases (+1.37 dB PSNR at 9 views), demonstrating that it recovers feature fidelity lost during compositing.

\begingroup
\centering
\scriptsize
\includegraphics[width=0.98\linewidth]{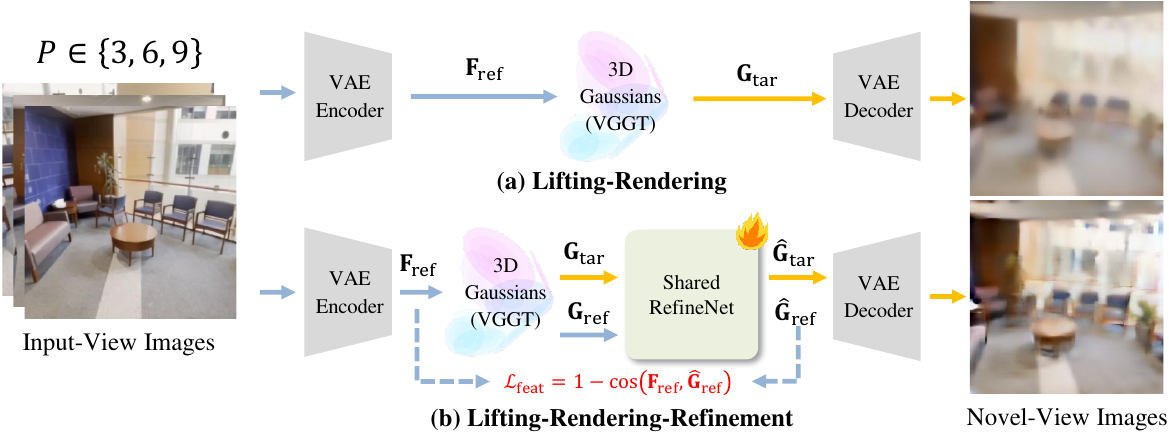}

\renewcommand{\arraystretch}{0.66}
\begin{center}
\begin{tabular}{cccccccc}
\toprule
Method & Refinement & \multicolumn{2}{c}{3 Input Views} & \multicolumn{2}{c}{6 Input Views} & \multicolumn{2}{c}{9 Input Views} \\
\cmidrule(lr){3-4} \cmidrule(lr){5-6} \cmidrule(lr){7-8}
 & & PSNR$\uparrow$ & SSIM$\uparrow$ & PSNR$\uparrow$ & SSIM$\uparrow$ & PSNR$\uparrow$ & SSIM$\uparrow$ \\
\midrule
\textbf{(a)} & \xmark & \snd{16.35} & \snd{0.56} & \snd{17.04} & \snd{0.58} & \snd{17.38} & \snd{0.59} \\
\textbf{(b)} & \cmark & \fst{16.59} & \fst{0.57} & \fst{18.07} & \fst{0.60} & \fst{18.75} & \fst{0.62} \\
\bottomrule
\end{tabular}
\end{center}
\refstepcounter{figure}\label{fig:supple_refinement}
\endgroup
\vspace{-3mm}


\vspace{-3mm}
\subsection{\normalsize Sensitivity to Classifier-free Guidance Scale.}
\label{subsec:supple_cfg}
SEVA~\cite{zhou2025stable} is sensitive to the classifier-free guidance (CFG) scale, which can induce color shifts or structural distortions regardless of viewpoint difficulty. As shown in the figure below, these artifacts appear in SEVA across varying CFG scales, whereas our method remains robust: the geometric conditioning provided by GS-Adapter stabilizes the generation and suppresses such distortions.
\begin{center}
\vspace{-2mm}
\includegraphics[width=0.99\linewidth]{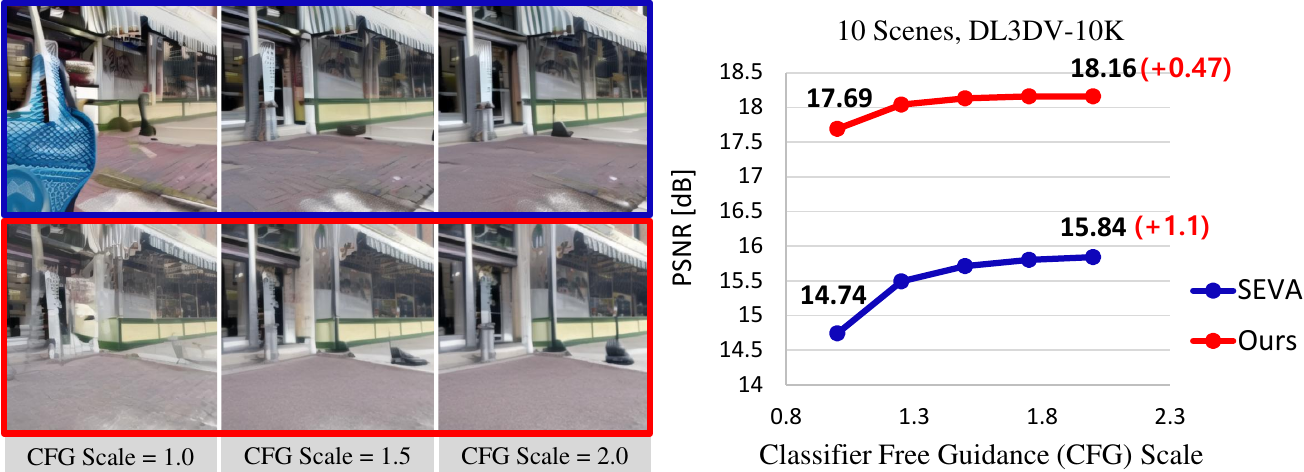}
\vspace{-3mm}
\end{center}

\newpage

\section{\normalsize Model Architecture Details}
\label{sec:supple_arch}
Detailed descriptions of the multi-scale fusion module and the RefineNet $\mathcal{R}$ used in the feature refinement of GS-Adapter are provided in \cref{subsec:geonvs} and \cref{subsec:gs_adapter}.

\subsection{\normalsize Multi-Scale Fusion}
\label{subsec:supple_multiscale}
\vspace{-2mm}
The detailed architecture of the multi-scale fusion module in GeoNVS is illustrated in~\cref{fig:multi_scale_fusion}. This module aggregates multi-scale features from the diffusion module to produce a single feature map at a unified resolution.

\subsection{\normalsize RefineNet Architecture}
\label{subsec:supple_refinenet}
\vspace{-2mm}
The detailed architecture of the RefineNet module in GS-Adapter is illustrated in~\cref{fig:refinenet_module}. This module takes the rasterized Gaussian feature $\mathbf{G}$ as input and produces a refined feature representation $\hat{\mathbf{G}}$.

\input{Supplementary_Material/figure/fusionblock}
\input{Supplementary_Material/figure/refinenet}

\newpage


\vspace{-1mm}
\section{\normalsize Training Dataset Preprocessing}
\label{sec:supple_data_preprocess}
\vspace{-2mm}
We construct our training dataset from DL3DV-10K~\cite{ling2024dl3dv} by preprocessing each scene into input--target view pairs annotated with pretrained 3D Gaussian representations (\cref{fig:training_dataset_pipeline}). Images are resized to $480 \times 270$ or $960 \times 540$ resolution, and camera poses are converted to world-to-camera extrinsics with normalized intrinsics. For structured view selection in~\cref{alg:view_selection}, we construct a pose graph over all frames 
using pairwise combined pose distances~\cite{duzceker2021deepvideomvs} and connect frames within a dynamic distance threshold that guarantees at least $N_{\text{group}}=21$ near neighbors per frame. Anchor nodes are sampled proportionally to their graph degree, yielding $K=3$ independent view groups per scene. Within each group, input views ($P \in \{1, 3, 6, 9, 12\}$) are selected via K-means clustering on camera position and orientation to maximize spatial diversity. The remaining frames serve as novel-view candidates and are divided into \textit{easy} and \textit{hard} subsets based on their minimum CLIP cosine distance~\cite{radford2021learning} to the input views, where the bottom 60\% by distance constitute easy samples. Both subsets are filtered by a frustum IoU threshold of $\tau_{\text{IoU}} = 0.4$ to ensure sufficient overlap between input and target cameras. Each resulting input--target pair is then processed by VGGT~\cite{wang2025vggt} to obtain an initial point cloud, which is further optimized into 3D Gaussians 
given the reference images, parameterized by position $\boldsymbol{\mu}$, rotation quaternion $\mathbf{q}$, scale $\mathbf{s}$, opacity $\alpha$, and spherical harmonic coefficients.

\input{Supplementary_Material/figure/dataset_preprocess}
\vspace{-3mm}
\begin{algorithm}
\scriptsize
\caption{View Selection for Novel View Synthesis}
\label{alg:view_selection}
\begin{algorithmic}[1]
\Require Images $\mathcal{I}$ with camera poses $\{\boldsymbol{\pi}_i\}_{i=1}^{P+Q}$, 
         number of anchors $V$, group size $N_g$, input view count $P$,
         frustum IoU threshold $\tau$
\Ensure Set of (input, target) view pairs $\mathcal{P}$

\State \textbf{Step 1: Build pose-distance graph}
\State Compute pairwise pose distance for all $(i,j)$:
\[
  d(i,j) = \sqrt{\|\mathbf{t}_{ij}\|^2 + 2\!\left(1 - \tfrac{\mathrm{tr}(\mathbf{R}_{ij})}{3}\right)}
\]
\State Set threshold $\delta$ s.t.\ each node has at least $N_g$ neighbors
\State Construct graph $G {=} (\mathcal{V}, \mathcal{E})$ where $(i,j) {\in} \mathcal{E}$ iff $d(i,j) {<} \delta$

\State \textbf{Step 2: Sample anchor nodes}
\State Sample $V$ anchors weighted by graph degree

\State \textbf{Step 3: Generate (input, target) pairs}
\State $\mathcal{P} \leftarrow \emptyset$
\For{each anchor $a_k$}
    \State $\Omega_k \leftarrow \{a_k\} \cup \mathcal{N}_G(a_k)$
    \State Select input views $\mathcal{S}_\text{ref}$ via K-Means on $[\mathbf{c}_i \| \hat{\mathbf{d}}_i]$
    \State Partition remaining views into $\mathcal{C}_\text{easy}$ / $\mathcal{C}_\text{hard}$ by CLIP distance to $\mathcal{S}_\text{ref}$
    \State Select easy targets via K-Means; select hard targets via random sampling
    \State Filter both by $\mathrm{FrustumIoU}(\cdot, \mathcal{S}_\text{ref}) \geq \tau$
    \State $\mathcal{P} \leftarrow \mathcal{P} \cup \{(\mathcal{S}_\text{ref},\; \mathcal{S}_\text{tar})\}$
\EndFor
\State \Return $\mathcal{P}$
\end{algorithmic}
\end{algorithm}

\newpage
\vspace{-2mm}
\section{\normalsize Additional Quantitative Results}
\label{sec:supple_additional_quant}
\vspace{-2mm}
We provide additional SSIM and LPIPS results:
\begin{itemize}
    \item \textbf{Small-viewpoint set NVS} in~\cref{tab:supple_smallnvs}, corresponding to~\cref{tab:method_comparison}.
    \item \textbf{Large-viewpoint set NVS} in~\cref{tab:supple_large_viewpoint}, corresponding to~\cref{tab:large_viewpoint}.
    \item \textbf{Long trajectory NVS} in~\cref{tab:supple_long_traj}, corresponding to~\cref{tab:long_traj_psnr}.
\end{itemize}
\input{Supplementary_Material/table/geonvs_notation_table}
\input{Supplementary_Material/table/supple_maintable}
\input{Supplementary_Material/table/supple_subtables}

\vspace{-4mm}
\section{\normalsize Additional Qualitative Results}
\label{sec:supple_qualitative}
\vspace{-2mm}
We provide additional qualitative results:
\begin{itemize}
    \item \textbf{Small-viewpoint set NVS} in~\cref{fig:supple_smallNVS_sample1,fig:supple_smallNVS_sample2,fig:supple_smallNVS_sample3,fig:supple_smallNVS_sample4}, corresponding to~\cref{tab:method_comparison}.
    \item \textbf{Large-viewpoint set NVS} in~\cref{fig:supple_largeNVS_sample1,fig:supple_largeNVS_sample2}, corresponding to~\cref{tab:large_viewpoint}.
    \item \textbf{GeoNVS with CameraCtrl} in~\cref{fig:supple_ablation_camctrl}, corresponding to~\cref{tab:ablation_lrms2}.
    \item \textbf{GeoNVS with geometry priors} in~\cref{fig:supple_ablation_lrm}, corresponding to~\cref{tab:ablation_lrms1}.
    \item \textbf{Trajectory NVS with ViPE~\cite{huang2025vipe} Reconstruction} in~\cref{fig:supple_trajNVS_sample1,fig:supple_trajNVS_sample2}, corresponding to~\cref{tab:long_traj_geo}.
    \item \textbf{Input-level injection vs Ours} in~\cref{fig:supple_ablation_trajnvs}, corresponding to~\cref{tab:geometric_consistency}.
\end{itemize}

\vspace{-4mm}
\section{\normalsize Limitations and Failure Cases}
\label{sec:supple_limitations}
\vspace{-2mm}
As shown in Fig.~\ref{fig:failure_cases}, GeoNVS exhibits two main failure modes. First, geometry inconsistency arises in thin structures such as poles and fences, where sparse Gaussian representations fail to accurately capture fine-grained geometry, propagating errors into the rendered novel-view features. Second, blurry textures appear in heavily occluded regions invisible from any reference view, where insufficient geometric guidance leaves the diffusion model to hallucinate plausible but blurry content. Incorporating uncertainty-aware Gaussian representations or inpainting-based priors for occluded regions remains a promising direction for future work.

\input{Supplementary_Material/figure/smallnvs_1}
\input{Supplementary_Material/figure/smallnvs_2}
\input{Supplementary_Material/figure/smallnvs_3}
\input{Supplementary_Material/figure/smallnvs_4}

\input{Supplementary_Material/figure/largenvs_1}
\input{Supplementary_Material/figure/largenvs_2}

\input{Supplementary_Material/figure/ablation_lrms}
\input{Supplementary_Material/figure/ablation_camctrl}

\input{Supplementary_Material/figure/trajnvs_1}
\input{Supplementary_Material/figure/trajnvs_2}
\input{Supplementary_Material/figure/ablation_trajnvs}

\input{Supplementary_Material/figure/limitations}

%% file: Supplementary_Material/figure/fusionblock.tex
\begin{figure}[t]
\centering
\includegraphics[width=0.97\linewidth]{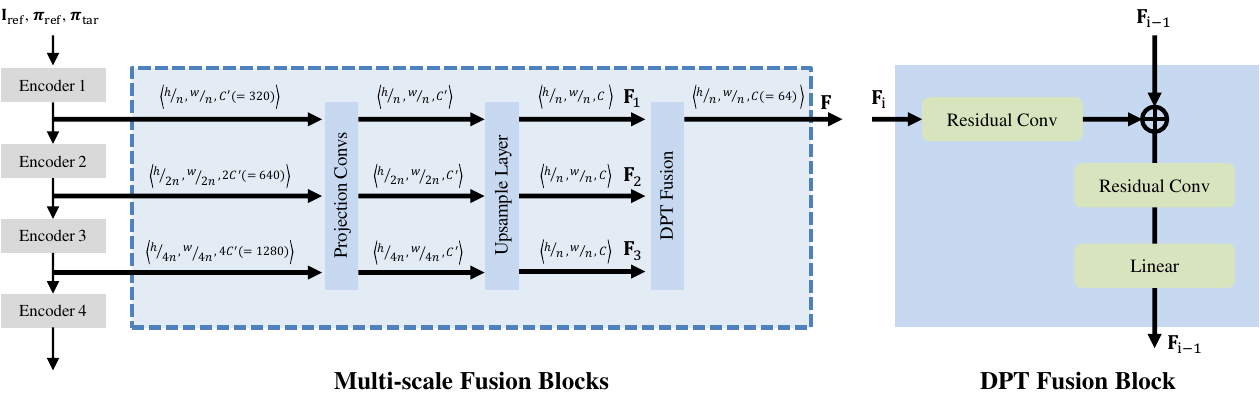}
\vspace{-2mm}
\caption{
\textbf{Multi-scale fusion module of GeoNVS.}}
\vspace{-3mm}
\label{fig:multi_scale_fusion}
\end{figure}

%% file: Supplementary_Material/figure/refinenet.tex
\begin{figure}[t]
\centering
\includegraphics[width=0.97\linewidth]{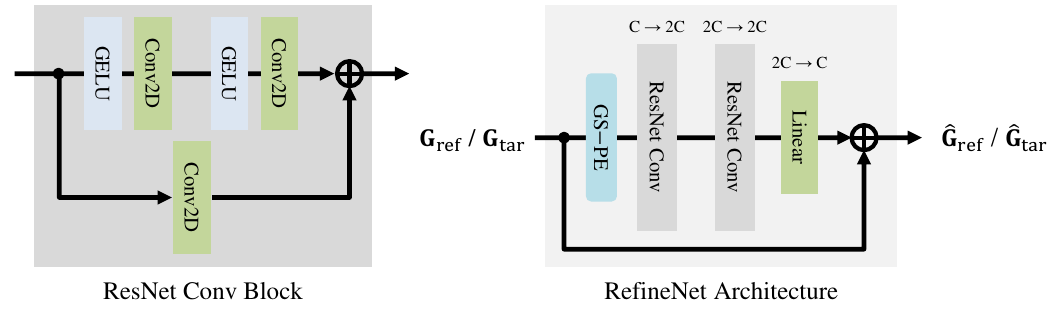}
\vspace{-2mm}
\caption{
\textbf{RefineNet module ($\mathcal{R}$) of GS-Adapter.}}
\vspace{-5mm}
\label{fig:refinenet_module}
\end{figure}

%% file: Supplementary_Material/figure/dataset_preprocess.tex
\begin{figure}[t]
\centering
\includegraphics[width=0.97\linewidth]{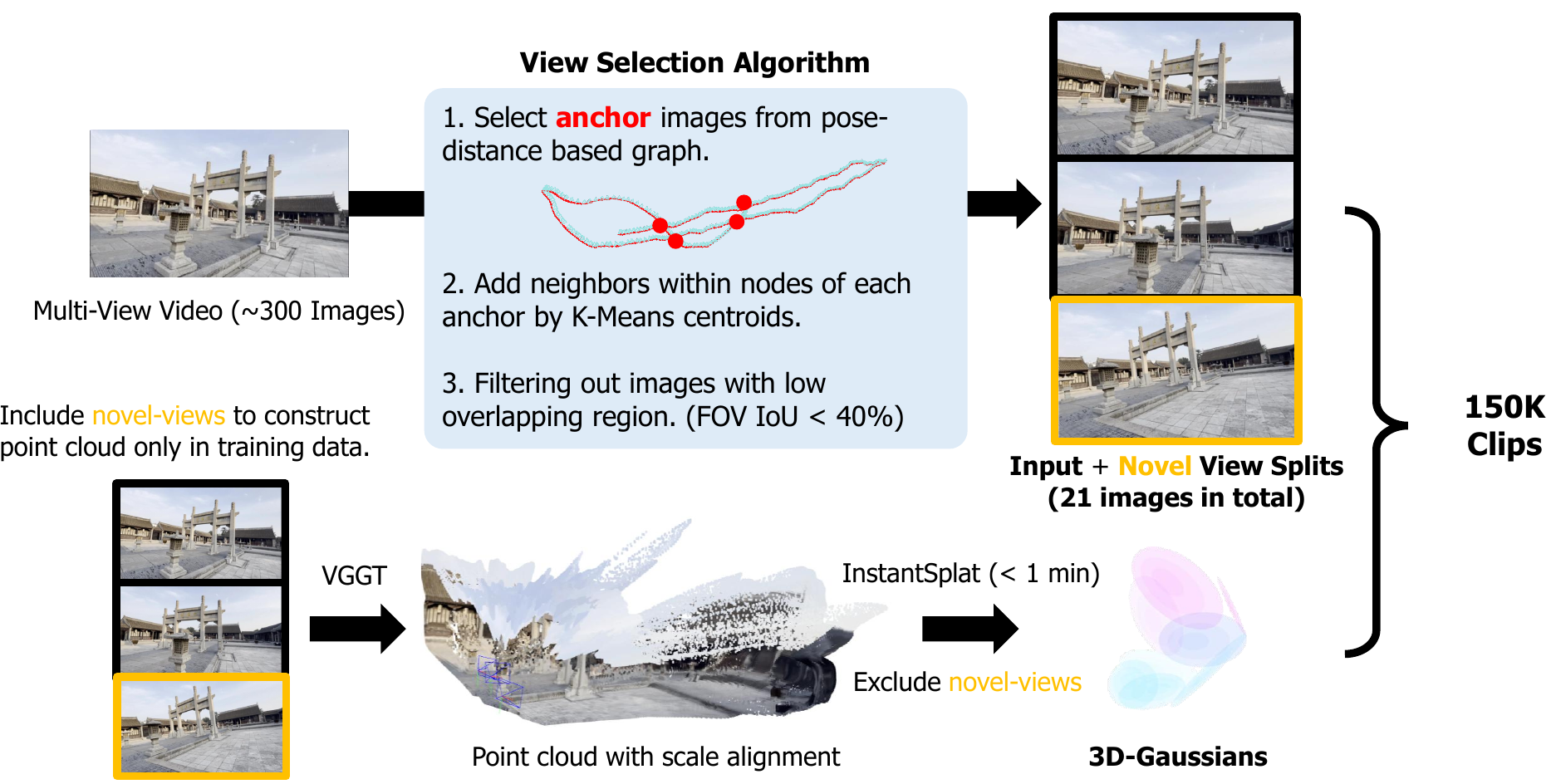}
\vspace{-2mm}
\caption{
\textbf{Pipeline to construct training dataset of GeoNVS.}}
\vspace{-5mm}
\label{fig:training_dataset_pipeline}
\end{figure}

%% file: Supplementary_Material/table/geonvs_notation_table.tex
\setlength{\aboverulesep}{0pt}
\setlength{\belowrulesep}{0pt}

\begin{table}[ht]
\centering
\scriptsize
\vspace{-3mm}
\caption{Notation Table for GeoNVS.}
\label{tab:notation}
\vspace{-3mm}
\begin{tabular}{M p{7.0cm}}

\toprule
\rowcolor{headerblue}
\multicolumn{1}{l}{\textcolor{white}{\textbf{Notation}}} &
\textcolor{white}{\textbf{Description}} \\
\midrule

\multicolumn{2}{l}{\textbf{Scene \& Camera}} \\[1pt]
\rowcolor{rowgray}
P, Q & Number of reference (input) views and target novel views \\
\mathbf{I}_\mathrm{ref}, \mathbf{I}_\mathrm{tar}
  & Reference view, target novel-view RGB images \\
\rowcolor{rowgray}
\boldsymbol{\pi}^{\mathrm{ref}}, \boldsymbol{\pi}^{\mathrm{tar}} & Camera poses of reference views and target views \\
s^{*} & Scale aligning predicted cameras to target cameras \\
\rowcolor{rowgray}
\mathbf{C}^{\mathrm{gt}},\ \mathbf{C}^{\mathrm{pred}}
  & Ground-truth and predicted camera centers \\[2pt]

\midrule
\multicolumn{2}{l}{\textbf{3D Gaussian Splatting}} \\[1pt]
\rowcolor{rowgray}
\mathcal{G} & Set of 3D Gaussians \\
N & Total number of Gaussian particles \\
\rowcolor{rowgray}
\boldsymbol{\mu} \in \mathbb{R}^{3} & 3D position of a Gaussian \\
\mathbf{r} \in \mathbb{R}^{4} & Rotation of a Gaussian (quaternion) \\
\rowcolor{rowgray}
\mathbf{s} \in \mathbb{R}^{3} & Scale of a Gaussian \\
\boldsymbol{\sigma} \in \mathbb{R}^{1} & Opacity of a Gaussian \\
\rowcolor{rowgray}
C(d,p) & Rendered color at pixel $p$ for view direction $d$ \\
c_{i}(d) & View-dependent color of the $i$-th Gaussian \\
\rowcolor{rowgray}
w_{i} & Rendering (blending) weight of the $i$-th Gaussian \\
\alpha_{i}
  & Product of opacity $\sigma_i$ and 2-D projected Gaussian density \\
\rowcolor{rowgray}
\mathrm{f}_{i} & Per-Gaussian feature vector \\[2pt]

\midrule
\multicolumn{2}{l}{\textbf{GS-Adapter \& Diffusion Features}} \\[1pt]
\rowcolor{rowgray}
n & VAE spatial downsampling factor \\
C & Channel dimension of diffusion features \\
\rowcolor{rowgray}
\mathbf{F}^{t} & Diffusion features at denoising timestep $t$ \\
\mathbf{F}^{t}_\mathrm{ref} \in
  \mathbb{R}^{P \times \frac{h}{n} \times \frac{w}{n} \times C}
  & Reference-view diffusion features at timestep $t$ \\
\rowcolor{rowgray}
\mathbf{F}^{t}_\mathrm{tar} \in
  \mathbb{R}^{Q \times \frac{h}{n} \times \frac{w}{n} \times C}
  & Novel-view diffusion features at timestep $t$ \\
\hat{\mathbf{F}}^{t} & Geometry-augmented diffusion features \\
\rowcolor{rowgray}
\hat{\mathbf{f}}_{i} \in \mathbb{R}^{C}
  & $\ell_2$-normalized uplifted feature of the $i$-th Gaussian \\
\mathbf{G}_\mathrm{tar} \in
  \mathbb{R}^{Q \times \frac{h}{n} \times \frac{w}{n} \times C}
  & Rasterized geometry-constrained novel-view features \\
\rowcolor{rowgray}
i^{*}(d,p) & Index of the most contributing Gaussian at pixel $(d,p)$ \\
\tilde{\mathbf{x}}(d,p) = (\tilde{x},\tilde{y},\tilde{z})
  & Normalized 3D center of $i^{*}(d,p)$ \\
\rowcolor{rowgray}
w^{*}(d,p) & Rendering weight of the dominant Gaussian $i^{*}(d,p)$ \\
\gamma(\cdot) & Sinusoidal positional encoding \\
\rowcolor{rowgray}
D' = C/4 & Encoding dimension for positional encoding \\
\omega_{k} = \omega_{0}^{-2k/D'} & Frequency for the $k$-th sinusoidal component \\
\rowcolor{rowgray}
\mathbf{G}'(d,p) & Positionally-encoded Gaussian feature (GS-PE applied) \\
\mathcal{R} & ResNet-based feature refinement network \\
\rowcolor{rowgray}
\hat{\mathbf{G}} = \mathcal{R}(\mathbf{G}') & Refined Gaussian features \\
\hat{\mathbf{G}}_\mathrm{ref} \in
  \mathbb{R}^{P \times \frac{h}{n} \times \frac{w}{n} \times C}
  & Refined reference-view Gaussian features \\
\rowcolor{rowgray}
\hat{\mathbf{G}}_\mathrm{tar}
  & Projected refined novel-view features \\
\mathbf{F}^{A}_\mathrm{tar} & Geometry-attended features via cross-attention \\
\rowcolor{rowgray}
\mathbf{W}_\mathrm{tar} \in [-1,1] & Pixel-wise confidence gate weight \\
\hat{\mathbf{F}}_\mathrm{tar} & Final geometry-augmented novel-view features \\[2pt]

\midrule
\multicolumn{2}{l}{\textbf{Loss Functions}} \\[1pt]
\rowcolor{rowgray}
\mathcal{L}_\mathrm{feat} & Feature consistency loss \\
\mathcal{L}_\mathrm{latent} & Latent space alignment loss \\

\midrule
\multicolumn{2}{l}{\textbf{Evaluation Metrics}} \\[1pt]
\rowcolor{rowgray}
\mathit{T}_\mathrm{err}\ [\mathrm{cm}] & Camera translation error \\
\mathit{R}_\mathrm{err}\ [\mathrm{deg}] & Camera rotation error \\
\rowcolor{rowgray}
\mathit{CD} & Chamfer Distance between reconstructed point clouds \\
\mathrm{PSNR}_\mathrm{V}
  & PSNR on co-visible regions w.r.t.\ reference views \\
\rowcolor{rowgray}
\mathrm{PSNR}_\mathrm{U} & PSNR on non-co-visible (unseen) regions \\

\bottomrule
\vspace{-4mm}
\end{tabular}
\end{table}

%% file: Supplementary_Material/table/supple_maintable.tex
\begin{table}[t]
\centering
\caption{\textbf{Small-viewpoint set NVS.} We report SSIM and LPIPS metrics corresponding to~\cref{tab:method_comparison} of the manuscript.
$P$ denotes the number of reference images.
For $P$=1, we sweep the unit length for camera normalization
following~\cite{zhou2025stable} to resolve scale ambiguity.
For each dataset, all combined (generative + geometry) methods use the top-performing feed-forward geometry model by PSNR as the geometry prior.}
\vspace{-5mm}

  \begin{subtable}[t]{\linewidth}
    \centering
    \caption{\textbf{SSIM $\uparrow$ on small-viewpoint set NVS.}}
    \label{tab:smallnvs_ssim}
    \resizebox{\columnwidth}{!}{%
    \setlength{\tabcolsep}{1.1pt}
    \renewcommand{\arraystretch}{1.4}
    \setlength{\aboverulesep}{2pt}
    \setlength{\belowrulesep}{2pt}
    \begin{tabular}{l *{22}{c}}
    \toprule
    \multirow{3}{*}{Method}
    & dataset
    & OO3D~\cite{wu2023omniobject3d}
    & \multicolumn{4}{c}{RE10K~\cite{zhou2018stereo}} 
    & \multicolumn{2}{c}{LLFF~\cite{mildenhall2019local}} 
    & \multicolumn{2}{c}{DTU~\cite{jensen2014large}}
    & \multicolumn{2}{c}{CO3D~\cite{reizenstein2021common}} 
    & \multicolumn{2}{c}{WRGBD~\cite{xia2024rgbd}} 
    & Mip360~\cite{barron2022mip} 
    & \multicolumn{2}{c}{DL3DV~\cite{ling2024dl3dv}} 
    & \multicolumn{2}{c}{T\&T~\cite{Knapitsch2017}} \\
    \cmidrule(lr){2-2}
    \cmidrule(lr){3-3}
    \cmidrule(lr){4-7}
    \cmidrule(lr){8-9}
    \cmidrule(lr){10-11}
    \cmidrule(lr){12-13}
    \cmidrule(lr){14-15}
    \cmidrule(lr){16-16}
    \cmidrule(lr){17-18}
    \cmidrule(lr){19-20}
    & split
    & S 
      & D & P & \multicolumn{2}{c}{R}
      & \multicolumn{2}{c}{R} 
      & \multicolumn{2}{c}{R} 
      & V & R 
      & S$_e$ & S$_h$ 
      & R 
      & S & L
      & \multicolumn{2}{c}{O} & Average \\
    \cmidrule(lr){2-2}
    \cmidrule(lr){3-3}
    \cmidrule(lr){4-4}
    \cmidrule(lr){5-5}
    \cmidrule(lr){6-7}
    \cmidrule(lr){8-9}
    \cmidrule(lr){10-11}
    \cmidrule(lr){12-12}
    \cmidrule(lr){13-13}
    \cmidrule(lr){14-14}
    \cmidrule(lr){15-15}
    \cmidrule(lr){16-16}
    \cmidrule(lr){17-17}
    \cmidrule(lr){18-18}
    \cmidrule(lr){19-20}
    & $P$
      & 3
      & 1 & 2 & 1 & 3
      & 1 & 3
      & 1 & 3
      & 1 & 3 
      & 3 & 6 
      & 6 
      & 6 & 9 
      & 2 & 3 & \\
    \midrule
    \multicolumn{20}{l}{\textbf{Feed-forward geometry models}} \\
    \multicolumn{2}{l}{MVSplat~\cite{chen2024mvsplat}} & 0.879 & 0.317 & \snd{0.868} & 0.364 & 0.870 & 0.003 & 0.323 & 0.140 & 0.409 & 0.237 & 0.390 & 0.406 & 0.393 & 0.239 & 0.379 & 0.427 & 0.333 & 0.377 & 0.409 \\
    \multicolumn{2}{l}{HiSplat~\cite{tang2024hisplat}} & 0.877 & 0.444 & \fst{0.872} & 0.540 & 0.902 & 0.01 & 0.384 & 0.149 & 0.467 & 0.286 & 0.434 & 0.434 & 0.419 & 0.290 & 0.419 & 0.457 & 0.360 & 0.468 & 0.456 \\
    \multicolumn{2}{l}{DepthSplat~\cite{xu2025depthsplat}} & \trd{0.897} & 0.468 & \snd{0.868} & 0.529 & \fst{0.918} & 0.02 & 0.332 & 0.262 & 0.485 & 0.337 & 0.492 & 0.441 & 0.434 & 0.303 & 0.465 & \trd{0.528} & 0.472 & 0.535 & 0.488 \\
    \multicolumn{2}{l}{VGGT~\cite{wang2025vggt} + InstantSplat~\cite{fan2024instantsplat}} & 0.892 & 0.499 & 0.798 & 0.561 & \snd{0.914} & 0.234 & \fst{0.672} & 0.215 & \trd{0.640} & 0.297 & 0.524 & 0.409 & 0.432 & \snd{0.362} & \snd{0.518} & \snd{0.571} & \fst{0.716} & \fst{0.807} & \trd{0.559} \\
    \multicolumn{2}{l}{Pi3~\cite{wang2025pi3} + InstantSplat~\cite{fan2024instantsplat}} & 0.857 & 0.392 & 0.807 & 0.452 & \trd{0.910} & 0.180 & 0.374 & 0.356 & 0.314 & 0.420 & 0.470 & 0.371 & 0.385 & 0.277 & 0.455 & 0.523 & 0.406 & 0.475 & 0.468 \\
    \midrule
    \multicolumn{20}{l}{\textbf{Generative models}} \\
    \multicolumn{2}{l}{MotionCtrl~\cite{wang2024motionctrl}} & - & 0.502 & - & 0.592 & - & \snd{0.314} & - & \trd{0.401} & - & 0.452 & - & - & - & - & - & - & - & - & 0.452 \\
    \multicolumn{2}{l}{CameraCtrl~\cite{he2025cameractrl}} & - & 0.491 & - & 0.578 & - & \trd{0.294} & - & 0.400 & - & 0.451 & - & - & - & - & - & - & - & - & 0.443 \\
    \multicolumn{2}{l}{ViewCrafter~\cite{yu2024viewcrafter}} & 0.861 & \trd{0.551} & - & \fst{0.693} & 0.647 & 0.283 & 0.253 & 0.370 & 0.274 & \snd{0.520} & 0.300 & 0.233 & 0.215 & 0.194 & 0.241 & 0.230 & 0.286 & 0.300 & 0.379 \\
    \multicolumn{2}{l}{SEVA~\cite{zhou2025stable}} & \fst{0.926} & 0.534 & 0.777 & 0.616 & 0.842 & \trd{0.294} & 0.467 & \snd{0.430} & \snd{0.659} & \trd{0.507} & \trd{0.551} & \snd{0.527} & \snd{0.522} & \trd{0.339} & 0.450 & 0.513 & 0.521 & 0.608 & \snd{0.560} \\
    \midrule
    \multicolumn{20}{l}{\textbf{Generative model + Geometry prior}} \\
    \multicolumn{2}{l}{MVSplat360~\cite{chen2024mvsplat360}} & 0.884 & 0.518 & 0.756 & 0.574 & 0.817 & \snd{0.314} & 0.419 & 0.366 & 0.517 & 0.459 & \snd{0.555} & \trd{0.514} & \trd{0.483} & 0.327 & 0.471 & 0.524 & 0.467 & 0.535 & 0.528 \\
    \multicolumn{2}{l}{GenFusion~\cite{wu2025genfusion}} & 0.876 & \snd{0.559} & 0.818 & \trd{0.618} & 0.864 & 0.271 & \trd{0.554} & 0.355 & 0.556 & 0.438 & 0.449 & 0.453 & 0.425 & 0.282 & 0.394 & 0.492 & \trd{0.630} & 0.689 & 0.540 \\
    \multicolumn{2}{l}{Difix3D~\cite{wu2025difix3d+}} & \trd{0.898} & 0.463 & \trd{0.857} & 0.528 & 0.903 & 0.230 & \snd{0.638} & 0.342 & 0.637 & 0.375 & 0.479 & 0.350 & 0.333 & 0.301 & \trd{0.481} & 0.519 & \snd{0.696} & \snd{0.779} & 0.545 \\
    \rowcolor[gray]{0.9}\multicolumn{2}{l}{GeoNVS (\textbf{Ours})} & \snd{0.904} & \fst{0.612} & 0.826 & \snd{0.680} & 0.895 & \fst{0.346} & 0.514 & \fst{0.480} & \fst{0.691} & \fst{0.550} & \fst{0.627} & \fst{0.607} & \fst{0.577} & \fst{0.383} & \fst{0.562} & \fst{0.635} & 0.603 & \trd{0.695} & \fst{0.622} \\
    \bottomrule
    \end{tabular}
    }
  \end{subtable}

  \vspace{2.5pt}

  \begin{subtable}[t]{\linewidth}
    \centering
    \caption{\textbf{LPIPS $\downarrow$ on small-viewpoint set NVS.}}
    \label{tab:smallnvs_lpips}
    \resizebox{\columnwidth}{!}{%
    \setlength{\tabcolsep}{1.1pt}
    \renewcommand{\arraystretch}{1.4}
    \setlength{\aboverulesep}{2pt}
    \setlength{\belowrulesep}{2pt}
    \begin{tabular}{l *{22}{c}}
    \toprule
    \multirow{3}{*}{Method}
    & dataset
    & OO3D~\cite{wu2023omniobject3d}
    & \multicolumn{4}{c}{RE10K~\cite{zhou2018stereo}} 
    & \multicolumn{2}{c}{LLFF~\cite{mildenhall2019local}} 
    & \multicolumn{2}{c}{DTU~\cite{jensen2014large}}
    & \multicolumn{2}{c}{CO3D~\cite{reizenstein2021common}} 
    & \multicolumn{2}{c}{WRGBD~\cite{xia2024rgbd}} 
    & Mip360~\cite{barron2022mip} 
    & \multicolumn{2}{c}{DL3DV~\cite{ling2024dl3dv}} 
    & \multicolumn{2}{c}{T\&T~\cite{Knapitsch2017}} \\
    \cmidrule(lr){2-2}
    \cmidrule(lr){3-3}
    \cmidrule(lr){4-7}
    \cmidrule(lr){8-9}
    \cmidrule(lr){10-11}
    \cmidrule(lr){12-13}
    \cmidrule(lr){14-15}
    \cmidrule(lr){16-16}
    \cmidrule(lr){17-18}
    \cmidrule(lr){19-20}
    & split
    & S 
      & D & P & \multicolumn{2}{c}{R}
      & \multicolumn{2}{c}{R} 
      & \multicolumn{2}{c}{R} 
      & V & R 
      & S$_e$ & S$_h$ 
      & R 
      & S & L
      & \multicolumn{2}{c}{O} & Average \\
    \cmidrule(lr){2-2}
    \cmidrule(lr){3-3}
    \cmidrule(lr){4-4}
    \cmidrule(lr){5-5}
    \cmidrule(lr){6-7}
    \cmidrule(lr){8-9}
    \cmidrule(lr){10-11}
    \cmidrule(lr){12-12}
    \cmidrule(lr){13-13}
    \cmidrule(lr){14-14}
    \cmidrule(lr){15-15}
    \cmidrule(lr){16-16}
    \cmidrule(lr){17-17}
    \cmidrule(lr){18-18}
    \cmidrule(lr){19-20}
    & $P$
      & 3
      & 1 & 2 & 1 & 3
      & 1 & 3
      & 1 & 3
      & 1 & 3 
      & 3 & 6 
      & 6 
      & 6 & 9 
      & 2 & 3 & \\
    \midrule
    \multicolumn{20}{l}{\textbf{Feed-forward geometry models}} \\
    \multicolumn{2}{l}{MVSplat~\cite{chen2024mvsplat}} & 0.152 & 0.643 & \snd{0.098} & 0.615 & 0.134 & 1.04 & 0.514 & 0.816 & 0.516 & 0.662 & 0.592 & 0.494 & 0.615 & 0.673 & 0.614 & 0.630 & 0.566 & 0.533 & 0.550 \\
    \multicolumn{2}{l}{HiSplat~\cite{tang2024hisplat}} & 0.155 & 0.522 & 0.110 & 0.484 & 0.094 & 1.02 & 0.479 & 0.783 & 0.483 & 0.645 & 0.581 & 0.496 & 0.610 & 0.691 & 0.627 & 0.650 & 0.612 & 0.474 & 0.529 \\
    \multicolumn{2}{l}{DepthSplat~\cite{xu2025depthsplat}} & 0.134 & 0.444 & \trd{0.100} & 0.414 & \snd{0.074} & 0.979 & 0.351 & 0.550 & 0.362 & 0.553 & 0.498 & \trd{0.431} & 0.493 & 0.552 & 0.487 & 0.477 & 0.332 & 0.235 & 0.415 \\
    \multicolumn{2}{l}{VGGT~\cite{wang2025vggt} + InstantSplat~\cite{fan2024instantsplat}} & 0.172 & 0.441 & 0.138 & 0.454 & \trd{0.078} & \trd{0.517} & \fst{0.153} & 0.779 & 0.340 & 0.608 & 0.530 & 0.696 & 0.623 & 0.618 & 0.466 & 0.497 & \snd{0.194} & \snd{0.127} & 0.413 \\
    \multicolumn{2}{l}{Pi3~\cite{wang2025pi3} + InstantSplat~\cite{fan2024instantsplat}} & 0.294 & 0.701 & 0.136 & 0.684 & 0.087 & 0.691 & 0.328 & 0.666 & 0.727 & 0.593 & 0.643 & 0.652 & 0.675 & 0.660 & 0.467 & 0.503 & 0.406 & 0.311 & 0.512 \\
    \midrule
    \multicolumn{20}{l}{\textbf{Generative models}} \\
    \multicolumn{2}{l}{MotionCtrl~\cite{wang2024motionctrl}} & - & 0.502 & - & 0.478 & - & 0.738 & - & 0.664 & - & 0.518 & - & - & - & - & - & - & - & - & 0.580 \\
    \multicolumn{2}{l}{CameraCtrl~\cite{he2025cameractrl}} & - & 0.514 & - & 0.492 & - & 0.610 & - & \trd{0.611} & - & 0.540 & - & - & - & - & - & - & - & - & 0.553 \\
    \multicolumn{2}{l}{ViewCrafter~\cite{yu2024viewcrafter}} & 0.211 & 0.405 & - & \snd{0.338} & 0.385 & 0.527 & 0.562 & 0.634 & 0.690 & \trd{0.425} & 0.727 & 0.715 & 0.727 & 0.675 & 0.649 & 0.680 & 0.542 & 0.516 & 0.553 \\
    \multicolumn{2}{l}{SEVA~\cite{zhou2025stable}} & \fst{0.062} & 0.400 & 0.131 & 0.402 & 0.095 & \snd{0.469} & \trd{0.212} & \snd{0.524} & \snd{0.210} & \snd{0.399} & \snd{0.333} & \snd{0.339} & \snd{0.331} & \snd{0.367} & \snd{0.305} & \snd{0.288} & 0.238 & 0.164 & \snd{0.293} \\
    \midrule
    \multicolumn{20}{l}{\textbf{Generative model + Geometry prior}} \\
    \multicolumn{2}{l}{MVSplat360~\cite{chen2024mvsplat360}} & 0.153 & 0.454 & 0.203 & 0.471 & 0.164 & 0.581 & 0.413 & 0.731 & 0.485 & 0.581 & 0.525 & 0.436 & \trd{0.487} & 0.556 & 0.380 & 0.410 & 0.445 & 0.348 & 0.435 \\
    \multicolumn{2}{l}{GenFusion~\cite{wu2025genfusion}} & 0.158 & 0.408 & 0.183 & \trd{0.384} & 0.159 & 0.554 & 0.301 & 0.702 & 0.430 & 0.621 & 0.616 & 0.507 & 0.577 & 0.655 & 0.537 & 0.520 & 0.278 & 0.223 & 0.434 \\
    \multicolumn{2}{l}{Difix3D~\cite{wu2025difix3d+}} & \trd{0.106} & 0.447 & \fst{0.097} & 0.422 & 0.083 & 0.520 & \fst{0.153} & 0.648 & \trd{0.240} & 0.544 & \trd{0.456} & 0.508 & 0.549 & \trd{0.425} & \trd{0.313} & \trd{0.326} & \fst{0.158} & \fst{0.101} & \trd{0.339} \\
    \rowcolor[gray]{0.9}\multicolumn{2}{l}{GeoNVS (\textbf{Ours})} & \snd{0.074} & \fst{0.325} & 0.113 & \fst{0.319} & \fst{0.071} & \fst{0.412} & \snd{0.208} & \fst{0.449} & \fst{0.174} & \fst{0.393} & \fst{0.279} & \fst{0.275} & \fst{0.290} & \fst{0.329} & \fst{0.245} & \fst{0.235} & \trd{0.225} & \trd{0.138} & \fst{0.253} \\
    \bottomrule
    \end{tabular}
    }
  \end{subtable}
\vspace{-6mm}
\label{tab:supple_smallnvs}
\end{table}

%% file: Supplementary_Material/table/supple_subtables.tex
\begin{table}[t]
\centering
\caption{\textbf{Large-viewpoint set NVS.}
The large-viewpoint set contains scenes with low view overlap between
reference and target views.
$P$ denotes the number of reference images.
We use DepthSplat~\cite{xu2025depthsplat} as the geometry prior for
both combined methods.}
\vspace{-5mm}

\begin{subtable}[t]{0.48\linewidth}
  \centering
  \caption{\textbf{SSIM $\uparrow$ results.}}
  \vspace{-3mm}
  \label{tab:large_viewpoint_ssim}
  \resizebox{\linewidth}{!}{%
  \setlength{\tabcolsep}{1.0pt}
  \renewcommand{\arraystretch}{1.3}
  \setlength{\aboverulesep}{2pt}
  \setlength{\belowrulesep}{2pt}
  \begin{tabular}{l c c c c c c c c c}
  \toprule
  \multirow{3}{*}{Method}
    & dataset
    & CO3D~\cite{reizenstein2021common}
    & \multicolumn{2}{c}{WRGBD~\cite{xia2024rgbd}}
    & \multicolumn{2}{c}{Mip360~\cite{barron2022mip}}
    & \multicolumn{2}{c}{DL3DV~\cite{ling2024dl3dv}} \\
  \cmidrule(lr){2-2}\cmidrule(lr){3-3}
  \cmidrule(lr){4-5}\cmidrule(lr){6-7}\cmidrule(lr){8-9}
    & split & R & \multicolumn{2}{c}{S$_h$} & \multicolumn{2}{c}{R} & \multicolumn{2}{c}{S} & Average \\
  \cmidrule(lr){2-2}\cmidrule(lr){3-3}
  \cmidrule(lr){4-5}\cmidrule(lr){6-7}\cmidrule(lr){8-9}
    & $P$ & 1 & 1 & 3 & 1 & 3 & 1 & 3 \\
  \midrule
  \multicolumn{6}{l}{\textbf{Geometry prior}} \\
  \multicolumn{2}{l}{DepthSplat~\cite{xu2025depthsplat}} & 0.170 & 0.171 & \trd{0.361} & 0.054 & \trd{0.283} & 0.067 & \snd{0.408} & 0.216 \\
  \midrule
  \multicolumn{6}{l}{\textbf{Generative models}} \\
  \multicolumn{2}{l}{MotionCtrl~\cite{wang2024motionctrl}}  & \snd{0.461} & \snd{0.412} & - & \fst{0.284} & - & \trd{0.318} & - & \trd{0.369} \\
  \multicolumn{2}{l}{CameraCtrl~\cite{he2025cameractrl}}    & \trd{0.456} & \trd{0.382} & - & \trd{0.249} & - & 0.307 & - & 0.349 \\
  \multicolumn{2}{l}{ViewCrafter~\cite{yu2024viewcrafter}}  & 0.384 & - & 0.231 & 0.222 & 0.190 & 0.271 & 0.229 & 0.255 \\
  \multicolumn{2}{l}{SEVA~\cite{zhou2025stable}}            & 0.428 & \trd{0.382} & \snd{0.455} & 0.235 & \snd{0.318} & \fst{0.543} & \trd{0.390} & \snd{0.393} \\
  \midrule
  \multicolumn{6}{l}{\textbf{Generative model + Geometry prior}} \\
  \multicolumn{2}{l}{GenFusion~\cite{wu2025genfusion}}      & 0.346 & 0.302 & 0.370 & 0.190 & 0.253 & 0.262 & \trd{0.390} & 0.302 \\
  \multicolumn{2}{l}{Difix3D~\cite{wu2025difix3d+}}        & 0.182 & 0.170 & 0.337 & 0.077 & 0.248 & 0.089 & 0.388 & 0.213 \\
  \rowcolor[gray]{0.9}
  \multicolumn{2}{l}{GeoNVS (\textbf{Ours})}               & \fst{0.534} & \fst{0.442} & \fst{0.507} & \snd{0.276} & \fst{0.352} & \snd{0.370} & \fst{0.480} & \fst{0.430} \\
  \bottomrule
  \end{tabular}
  }
\end{subtable}
\hfill
\begin{subtable}[t]{0.48\linewidth}
  \centering
  \caption{\textbf{LPIPS $\downarrow$ results.}}
  \vspace{-3mm}
  \label{tab:large_viewpoint_lpips}
  \resizebox{\linewidth}{!}{%
  \setlength{\tabcolsep}{1.0pt}
  \renewcommand{\arraystretch}{1.3}
  \setlength{\aboverulesep}{2pt}
  \setlength{\belowrulesep}{2pt}
  \begin{tabular}{l c c c c c c c c c}
  \toprule
  \multirow{3}{*}{Method}
    & dataset
    & CO3D~\cite{reizenstein2021common}
    & \multicolumn{2}{c}{WRGBD~\cite{xia2024rgbd}}
    & \multicolumn{2}{c}{Mip360~\cite{barron2022mip}}
    & \multicolumn{2}{c}{DL3DV~\cite{ling2024dl3dv}} \\
  \cmidrule(lr){2-2}\cmidrule(lr){3-3}
  \cmidrule(lr){4-5}\cmidrule(lr){6-7}\cmidrule(lr){8-9}
    & split & R & \multicolumn{2}{c}{S$_h$} & \multicolumn{2}{c}{R} & \multicolumn{2}{c}{S} & Average \\
  \cmidrule(lr){2-2}\cmidrule(lr){3-3}
  \cmidrule(lr){4-5}\cmidrule(lr){6-7}\cmidrule(lr){8-9}
    & $P$ & 1 & 1 & 3 & 1 & 3 & 1 & 3 \\
  \midrule
  \multicolumn{6}{l}{\textbf{Geometry prior}} \\
  \multicolumn{2}{l}{DepthSplat~\cite{xu2025depthsplat}} & 0.797 & 0.756 & 0.577 & 0.807 & 0.609 & 0.828 & 0.538 & 0.702 \\
  \midrule
  \multicolumn{6}{l}{\textbf{Generative models}} \\
  \multicolumn{2}{l}{MotionCtrl~\cite{wang2024motionctrl}}  & \trd{0.616} & 0.723 & - & 0.817 & - & 0.720 & - & 0.719 \\
  \multicolumn{2}{l}{CameraCtrl~\cite{he2025cameractrl}}    & 0.643 & \trd{0.676} & - & 0.701 & - & 0.648 & - & 0.667 \\
  \multicolumn{2}{l}{ViewCrafter~\cite{yu2024viewcrafter}}  & 0.659 & - & 0.729 & \trd{0.647} & 0.676 & \trd{0.602} & 0.645 & 0.660 \\
  \multicolumn{2}{l}{SEVA~\cite{zhou2025stable}}            & \snd{0.583} & \snd{0.579} & \snd{0.427} & \snd{0.618} & \snd{0.432} & \snd{0.543} & \trd{0.388} & \snd{0.510} \\
  \midrule
  \multicolumn{6}{l}{\textbf{Generative model + Geometry prior}} \\
  \multicolumn{2}{l}{GenFusion~\cite{wu2025genfusion}}      & 0.715 & 0.703 & 0.648 & 0.738 & 0.663 & 0.657 & 0.547 & 0.667 \\
  \multicolumn{2}{l}{Difix3D~\cite{wu2025difix3d+}}        & 0.785 & 0.749 & \trd{0.506} & 0.744 & \trd{0.477} & 0.749 & \snd{0.382} & \trd{0.627} \\
  \rowcolor[gray]{0.9}
  \multicolumn{2}{l}{GeoNVS (\textbf{Ours})}               & \fst{0.504} & \fst{0.556} & \fst{0.390} & \fst{0.585} & \fst{0.386} & \fst{0.510} & \fst{0.333} & \fst{0.466} \\
  \bottomrule
  \end{tabular}
  }
\end{subtable}
\vspace{-4mm}
\label{tab:supple_large_viewpoint}
\end{table}

\begin{table}[t]
\centering
\caption{\textbf{Long trajectory.}
The long trajectory NVS measures the long-term generation ability of
generative NVS. $P$ denotes the number of reference images.
We use VGGT~\cite{wang2025vggt} as the geometry prior for our method.}
\vspace{-3mm}

\begin{subtable}[t]{0.48\linewidth}
  \centering
  \caption{\textbf{SSIM $\uparrow$ results.}}
  \vspace{-3mm}
  \label{tab:long_traj_ssim}
  \resizebox{\linewidth}{!}{%
  \setlength{\tabcolsep}{1.1pt}
  \renewcommand{\arraystretch}{1.3}
  \setlength{\aboverulesep}{2pt}
  \setlength{\belowrulesep}{2pt}
  \begin{tabular}{l *{10}{c}}
  \toprule
  \multirow{2}{*}{Method}
    & dataset
    & \multicolumn{3}{c}{Mip360~\cite{barron2022mip}}
    & \multicolumn{3}{c}{DL3DV~\cite{ling2024dl3dv}}
    & \multicolumn{3}{c}{T\&T~\cite{Knapitsch2017}} \\
  \cmidrule(lr){2-2}\cmidrule(lr){3-5}
  \cmidrule(lr){6-8}\cmidrule(lr){9-11}
    & $P$ & 3 & 6 & 9 & 3 & 6 & 9 & 2 & 3 & 6 \\
  \midrule
  \multicolumn{2}{l}{DepthSplat~\cite{xu2025depthsplat}} & 0.277 & 0.292 & 0.310 & \trd{0.415} & \trd{0.464} & \trd{0.490} & 0.474 & 0.530 & 0.552 \\
  \multicolumn{2}{l}{VGGT~\cite{wang2025vggt}}           & \snd{0.305} & \snd{0.351} & \snd{0.383} & \snd{0.466} & \snd{0.515} & \snd{0.537} & \fst{0.718} & \fst{0.809} & \fst{0.860} \\
  \multicolumn{2}{l}{SEVA~\cite{zhou2025stable}}         & \trd{0.302} & \trd{0.326} & \trd{0.343} & 0.384 & 0.442 & 0.479 & \trd{0.534} & \trd{0.614} & \trd{0.721} \\
  \rowcolor[gray]{0.9}
  \multicolumn{2}{l}{\textbf{Ours}}                     & \fst{0.341} & \fst{0.372} & \fst{0.395} & \fst{0.485} & \fst{0.559} & \fst{0.607} & \snd{0.628} & \snd{0.712} & \snd{0.763} \\
  \bottomrule
  \end{tabular}
  }
\end{subtable}
\hfill
\begin{subtable}[t]{0.48\linewidth}
  \centering
  \caption{\textbf{LPIPS $\downarrow$ results.}}
  \vspace{-3mm}
  \label{tab:long_traj_lpips}
  \resizebox{\linewidth}{!}{%
  \setlength{\tabcolsep}{1.1pt}
  \renewcommand{\arraystretch}{1.3}
  \setlength{\aboverulesep}{2pt}
  \setlength{\belowrulesep}{2pt}
  \begin{tabular}{l *{10}{c}}
  \toprule
  \multirow{2}{*}{Method}
    & dataset
    & \multicolumn{3}{c}{Mip360~\cite{barron2022mip}}
    & \multicolumn{3}{c}{DL3DV~\cite{ling2024dl3dv}}
    & \multicolumn{3}{c}{T\&T~\cite{Knapitsch2017}} \\
  \cmidrule(lr){2-2}\cmidrule(lr){3-5}
  \cmidrule(lr){6-8}\cmidrule(lr){9-11}
    & $P$ & 3 & 6 & 9 & 3 & 6 & 9 & 2 & 3 & 6 \\
  \midrule
  \multicolumn{2}{l}{DepthSplat~\cite{xu2025depthsplat}} & \trd{0.644} & \trd{0.593} & \trd{0.560} & 0.547 & 0.494 & \trd{0.464} & 0.376 & 0.301 & 0.259 \\
  \multicolumn{2}{l}{VGGT~\cite{wang2025vggt}}           & 0.659 & 0.633 & 0.645 & \trd{0.466} & \trd{0.469} & 0.488 & \fst{0.193} & \fst{0.126} & \trd{0.124} \\
  \multicolumn{2}{l}{SEVA~\cite{zhou2025stable}}         & \snd{0.479} & \snd{0.417} & \snd{0.378} & \snd{0.400} & \snd{0.315} & \snd{0.262} & \trd{0.241} & \trd{0.175} & \snd{0.112} \\
  \rowcolor[gray]{0.9}
  \multicolumn{2}{l}{\textbf{Ours}}                     & \fst{0.341} & \fst{0.370} & \fst{0.329} & \fst{0.343} & \fst{0.253} & \fst{0.206} & \snd{0.197} & \snd{0.137} & \fst{0.105} \\
  \bottomrule
  \end{tabular}
  }
\end{subtable}
\vspace{-4mm}
\label{tab:supple_long_traj}
\end{table}

%% file: Supplementary_Material/figure/smallnvs_1.tex
\begin{figure}[t]
\centering
\subfloat[\textbf{Qualitative results.} Given sparse \textcolor{red}{reference-view} images, novel-view images are synthesized at \textcolor{blue}{target viewpoints}.\label{fig:qual_smallNVS_1}]{
    \includegraphics[width=1.0\textwidth]{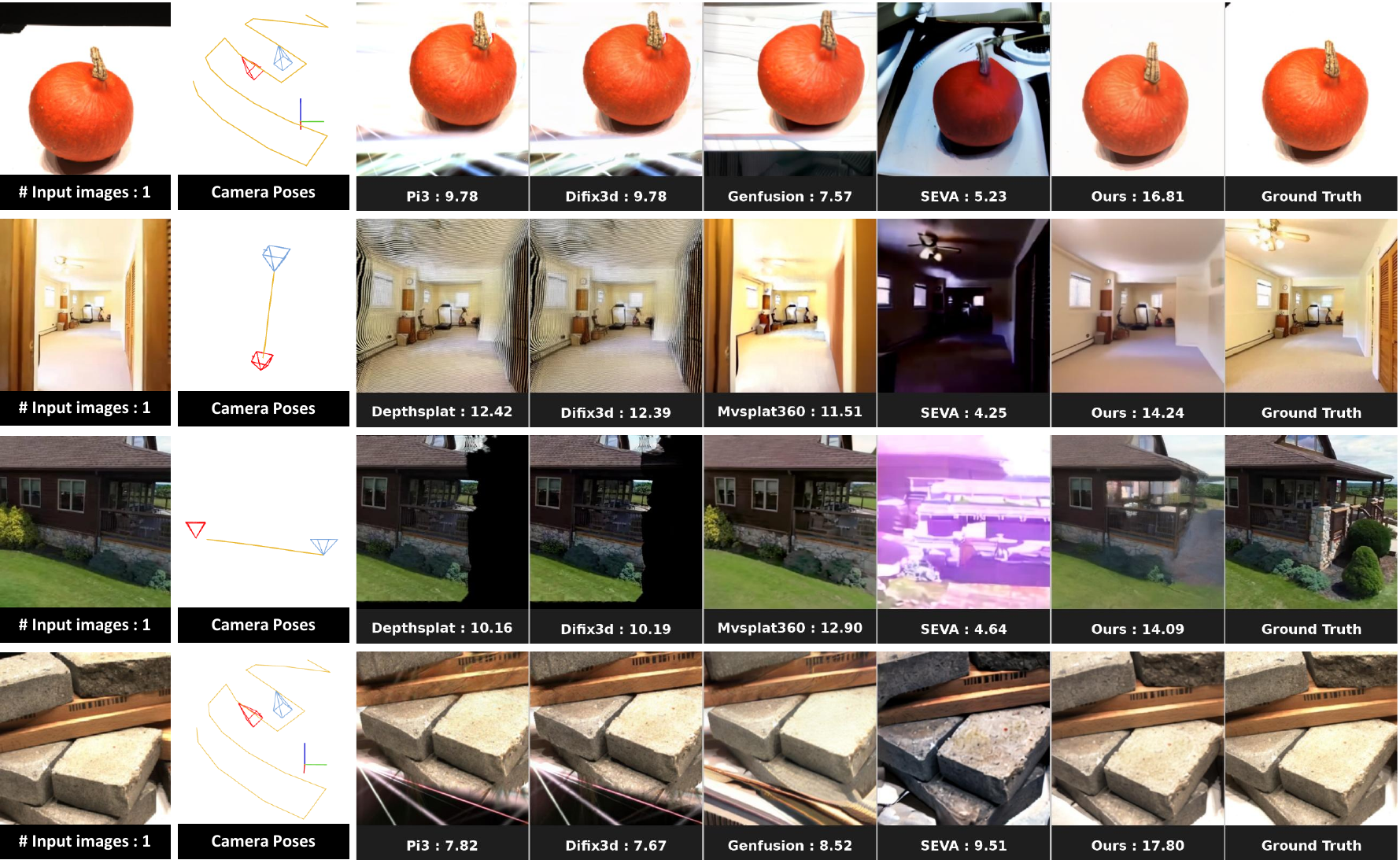}
}\\ 
\vspace{1mm}
\subfloat[\textbf{Qualitative results.} Given sparse \textcolor{red}{reference-view} images, novel-view images are synthesized at \textcolor{blue}{target viewpoints}.\label{fig:qual_smallNVS_2}]{
\vspace{1mm}
    \includegraphics[width=1.0\textwidth]{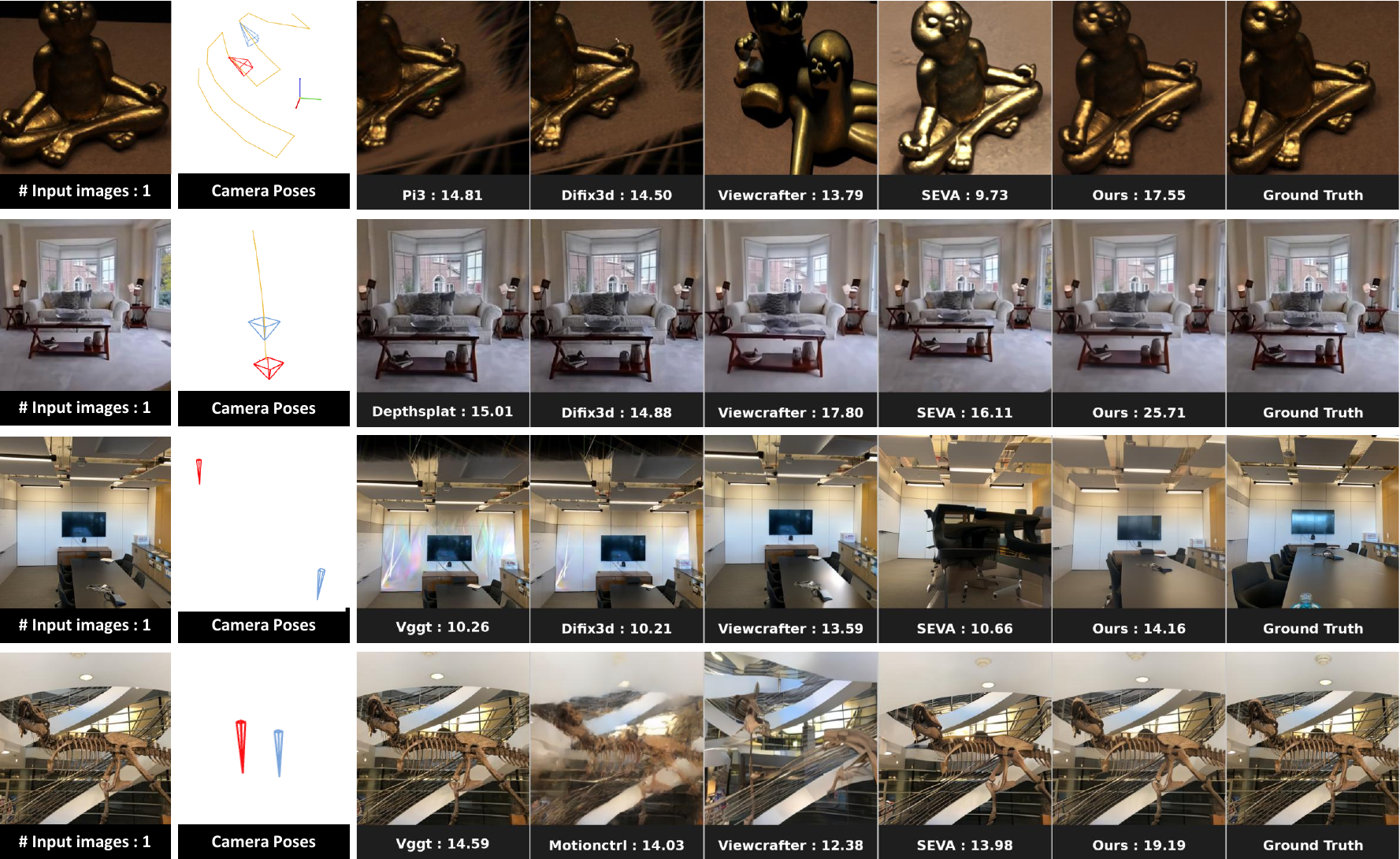}
}\\
\vspace{2mm}
\caption{\textbf{Qualitative results of small-viewpoint NVS.}}
\label{fig:supple_smallNVS_sample1}
\vspace{-4mm}
\end{figure}

%% file: Supplementary_Material/figure/smallnvs_2.tex
\begin{figure}[t]
\centering
\subfloat[\textbf{Qualitative results.} Given sparse \textcolor{red}{reference-view} images, novel-view images are synthesized at \textcolor{blue}{target viewpoints}.\label{fig:qual_smallNVS_3}]{
    \includegraphics[width=1.0\textwidth]{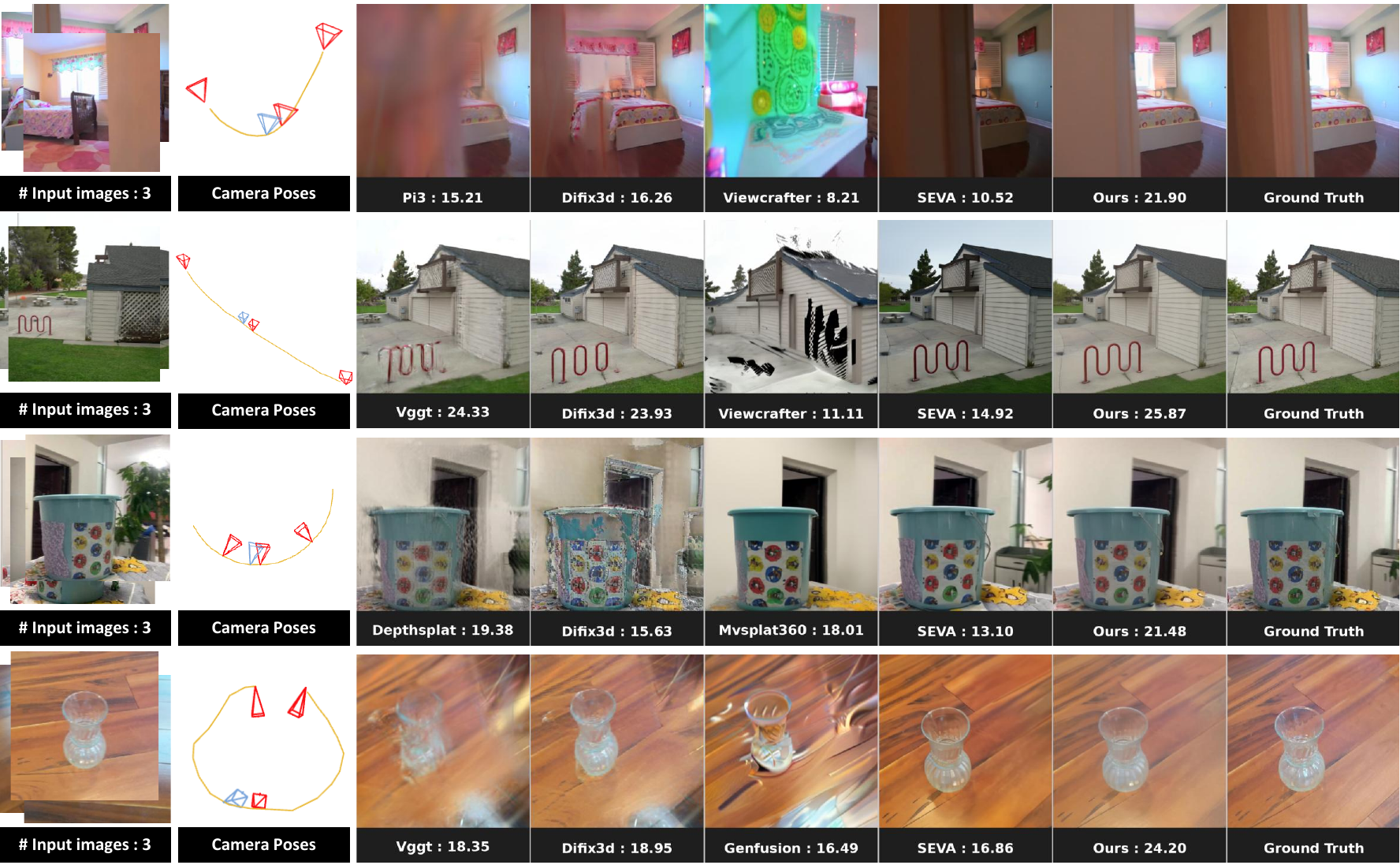}
}\\ 
\vspace{1mm}
\subfloat[\textbf{Qualitative results.} Given sparse \textcolor{red}{reference-view} images, novel-view images are synthesized at \textcolor{blue}{target viewpoints}.\label{fig:qual_smallNVS_4}]{
\vspace{1mm}
    \includegraphics[width=1.0\textwidth]{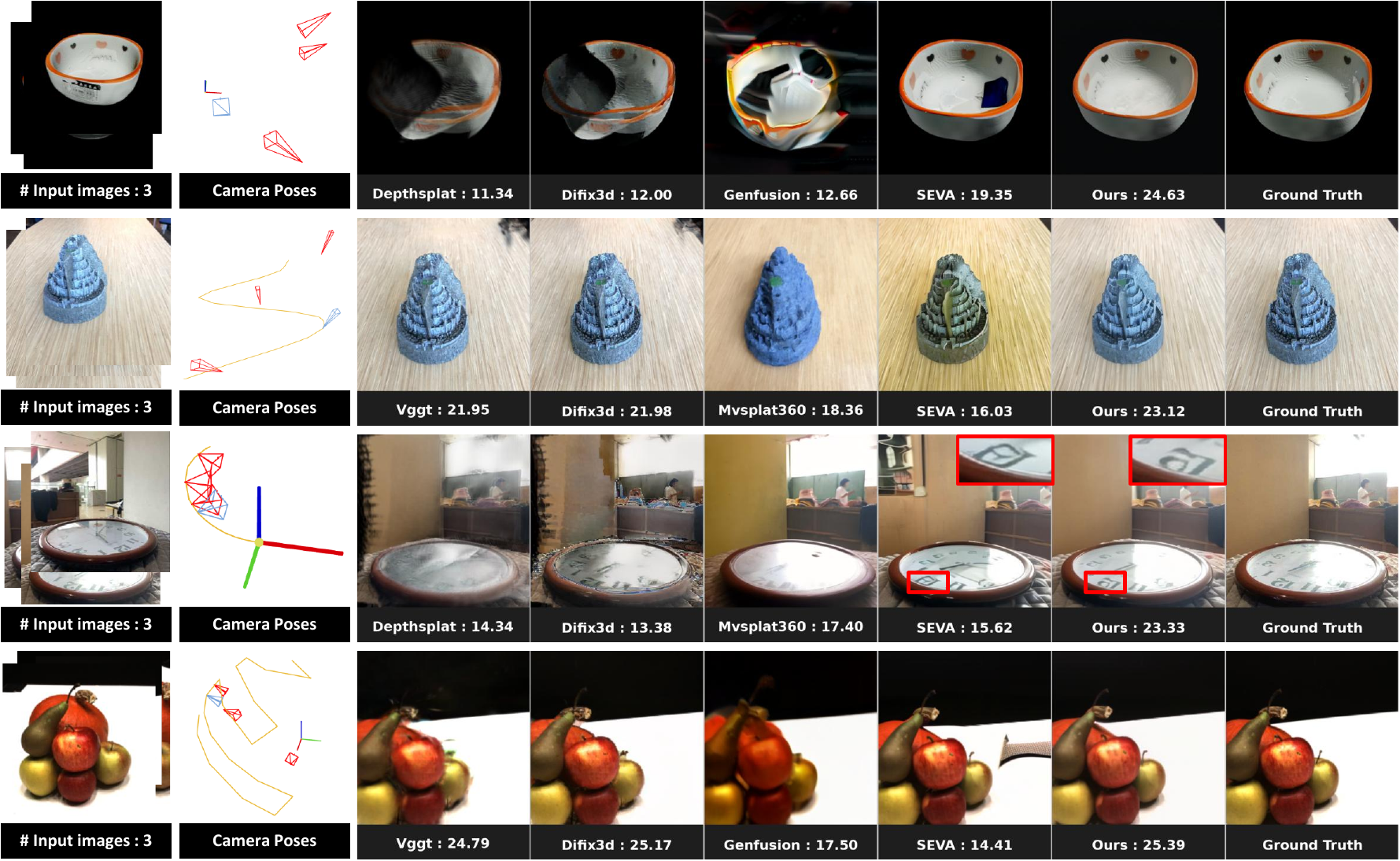}
}\\
\vspace{2mm}
\caption{\textbf{Qualitative results of small-viewpoint NVS.}}
\label{fig:supple_smallNVS_sample2}
\vspace{-4mm}
\end{figure}

%% file: Supplementary_Material/figure/smallnvs_3.tex
\begin{figure}[t]
\centering
\subfloat[\textbf{Qualitative results.} Given sparse \textcolor{red}{reference-view} images, novel-view images are synthesized at \textcolor{blue}{target viewpoints}.\label{fig:qual_smallNVS_5}]{
    \includegraphics[width=1.0\textwidth]{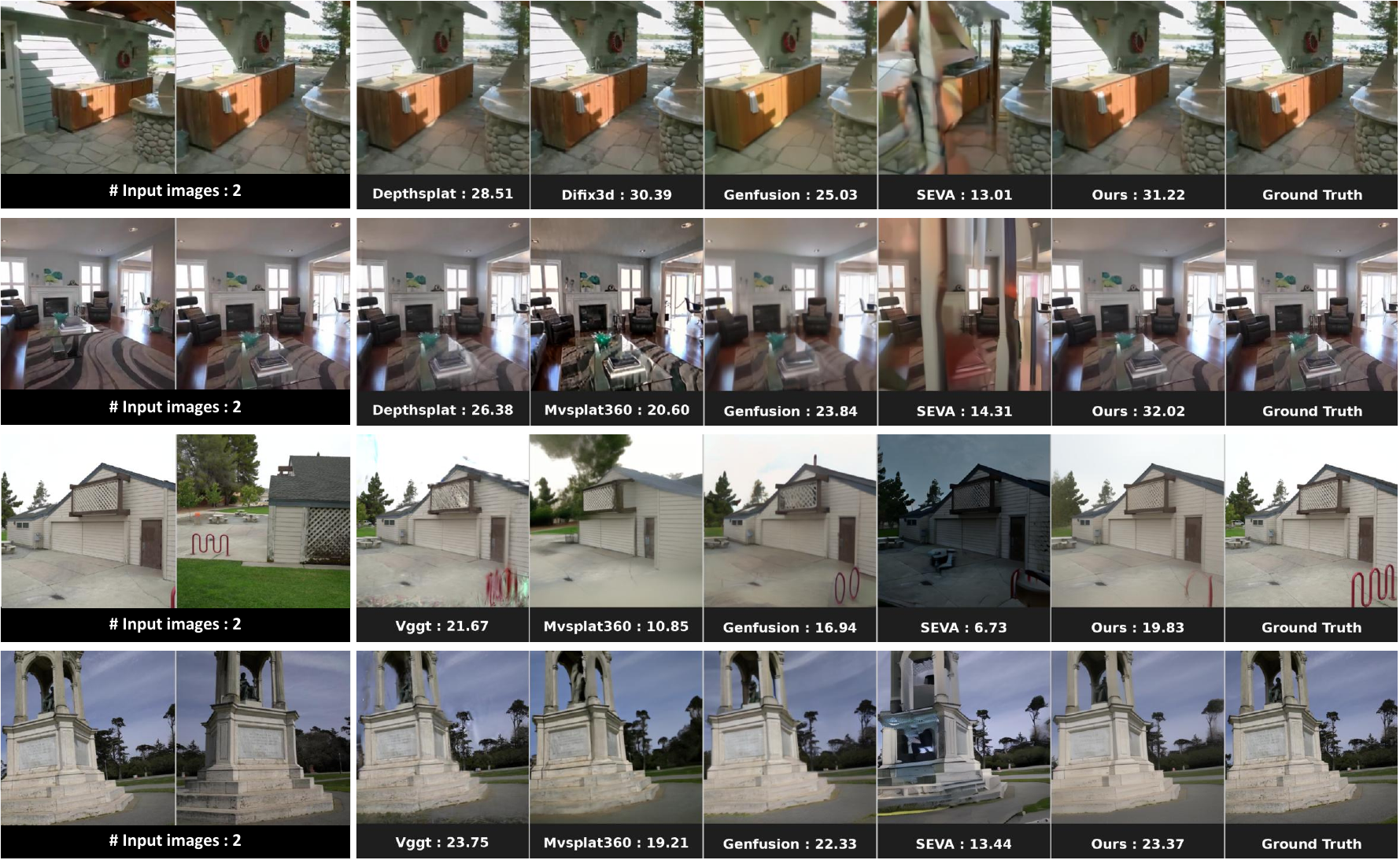}
}\\ 
\vspace{1mm}
\subfloat[\textbf{Qualitative results.} Given sparse \textcolor{red}{reference-view} images, novel-view images are synthesized at \textcolor{blue}{target viewpoints}.\label{fig:qual_smallNVS_6}]{
\vspace{1mm}
    \includegraphics[width=1.0\textwidth]{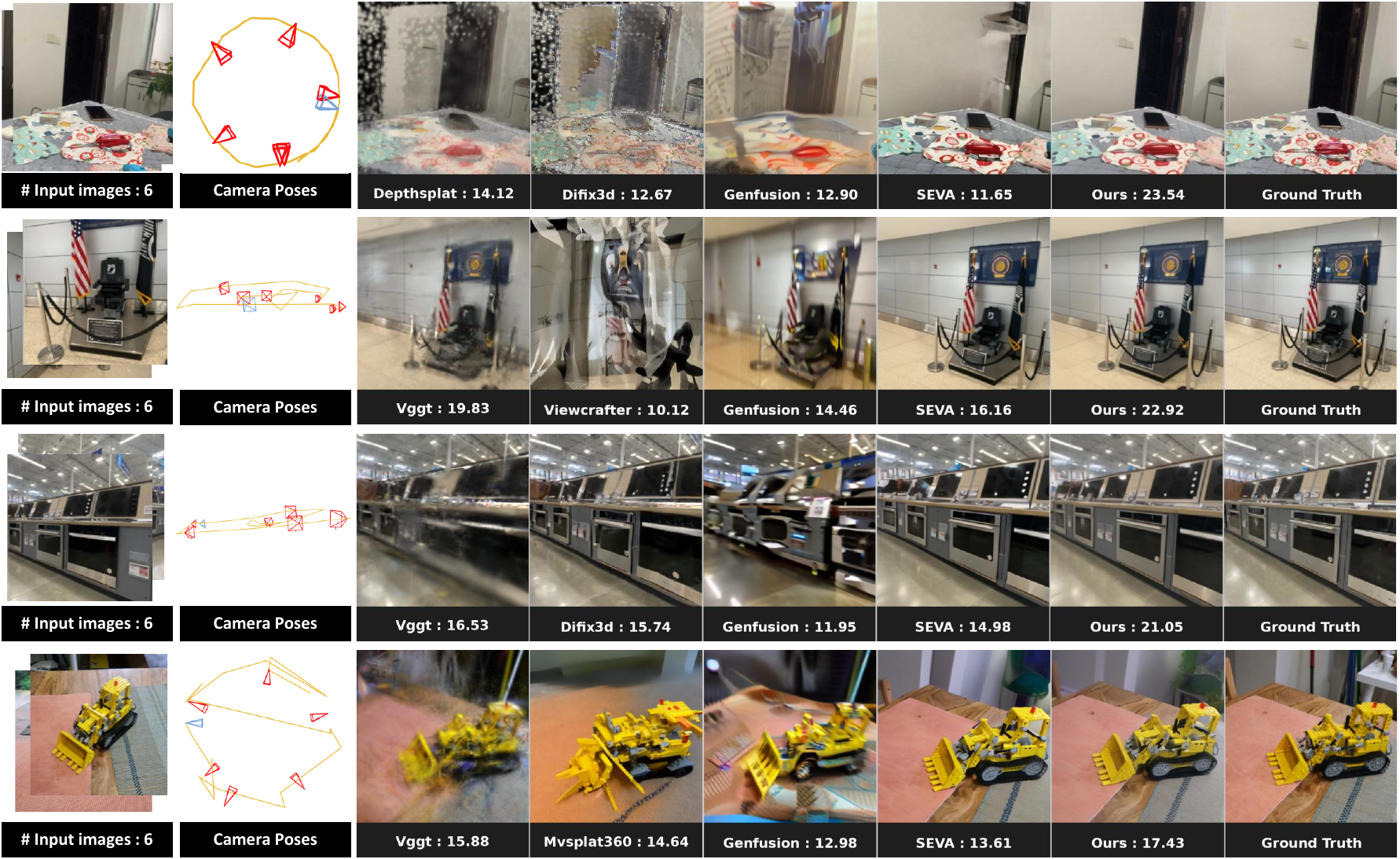}
}\\
\vspace{2mm}
\caption{\textbf{Qualitative results of small-viewpoint NVS.}}
\label{fig:supple_smallNVS_sample3}
\vspace{-4mm}
\end{figure}

%% file: Supplementary_Material/figure/smallnvs_4.tex
\begin{figure}[t]
\centering
\subfloat[\textbf{Qualitative results.} Given sparse \textcolor{red}{reference-view} images, novel-view images are synthesized at \textcolor{blue}{target viewpoints}.\label{fig:qual_smallNVS_7}]{
    \includegraphics[width=1.0\textwidth]{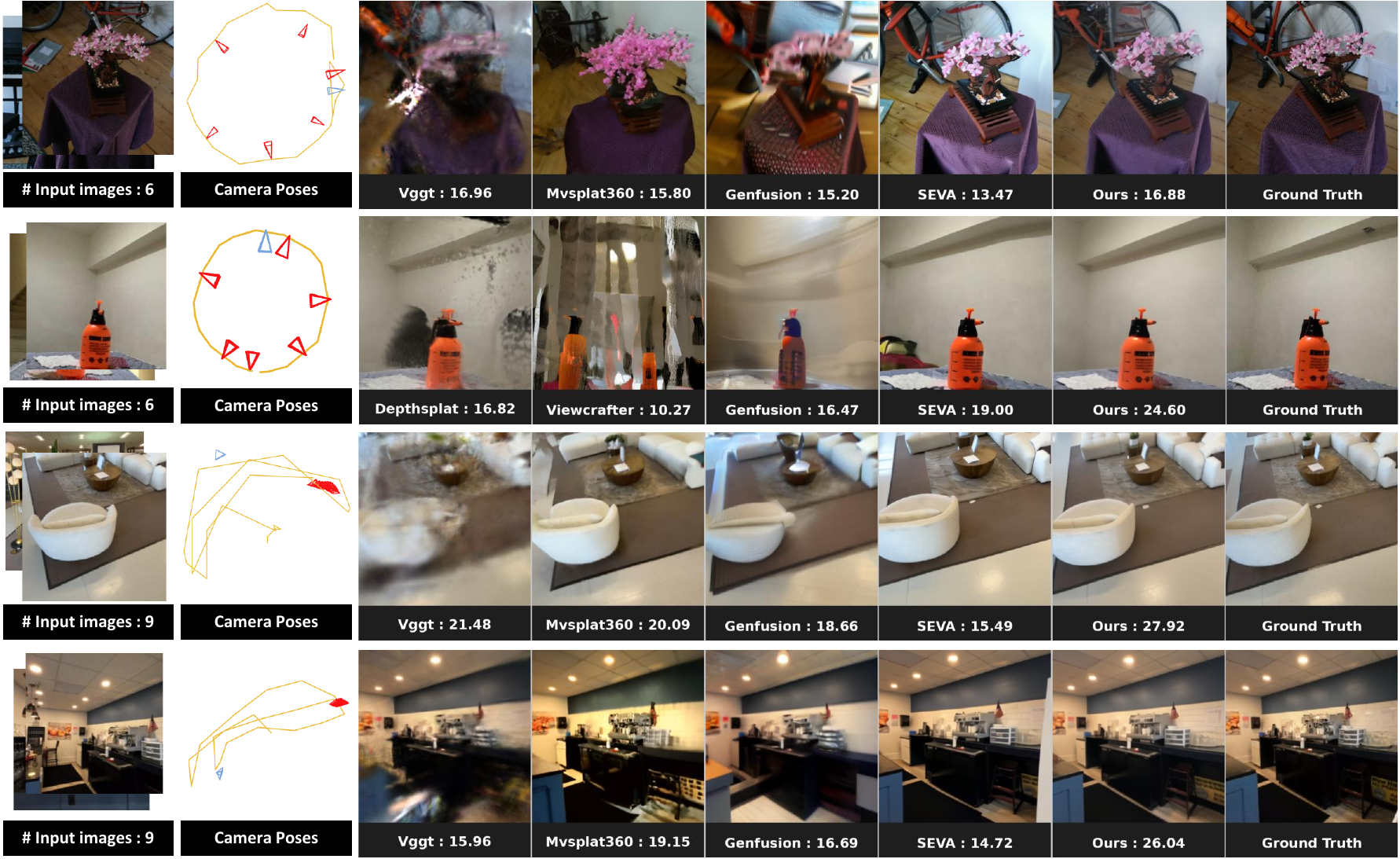}
}\\ 
\vspace{1mm}
\subfloat[\textbf{Qualitative results.} Given sparse \textcolor{red}{reference-view} images, novel-view images are synthesized at \textcolor{blue}{target viewpoints}.\label{fig:qual_smallNVS_8}]{
\vspace{1mm}
    \includegraphics[width=1.0\textwidth]{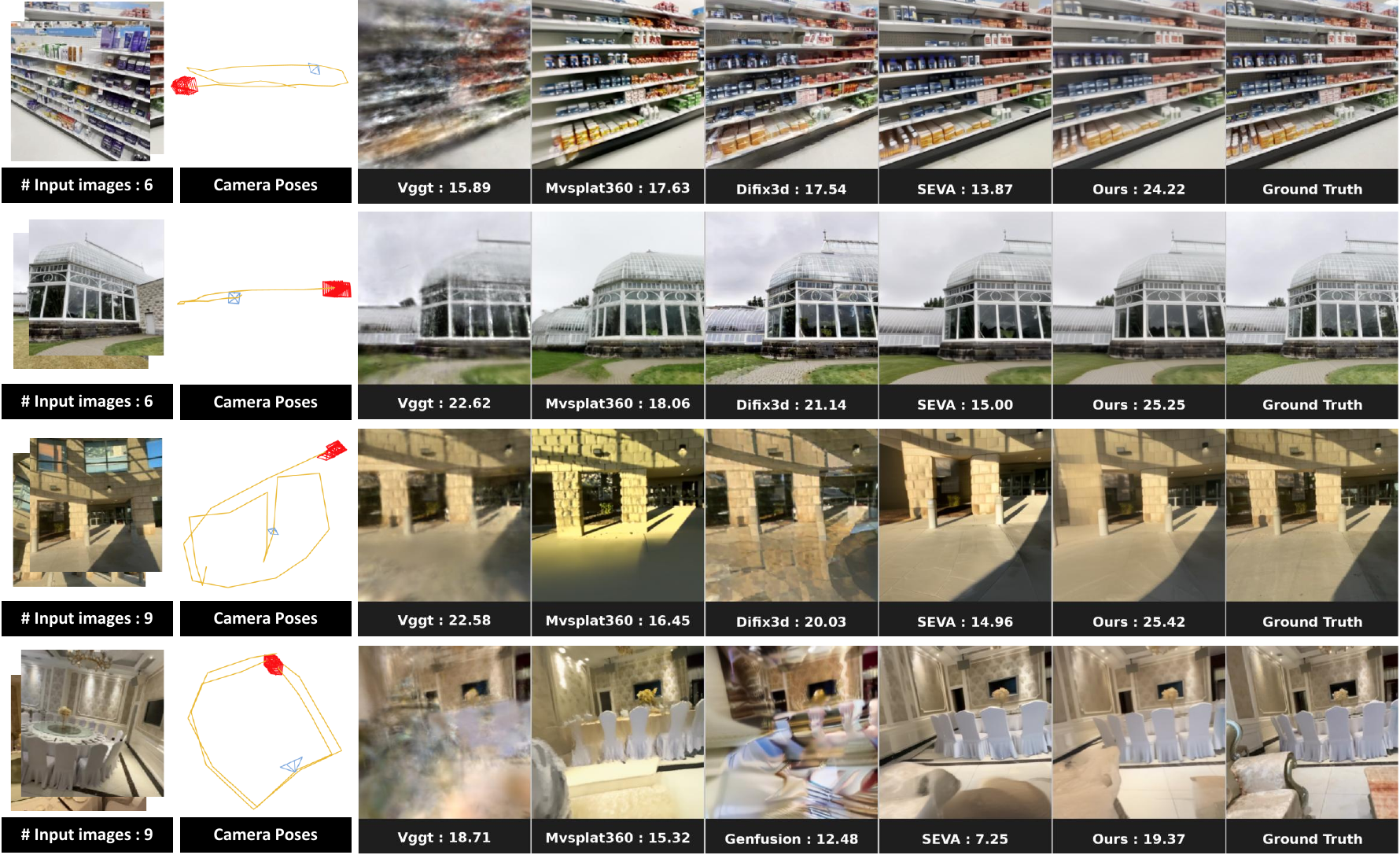}
}\\
\vspace{2mm}
\caption{\textbf{Qualitative results of small-viewpoint NVS.}}
\label{fig:supple_smallNVS_sample4}
\vspace{-4mm}
\end{figure}

%% file: Supplementary_Material/figure/largenvs_1.tex
\begin{figure}[t]
\centering
\subfloat[\textbf{Qualitative results.} Given sparse \textcolor{red}{reference-view} images, novel-view images are synthesized at \textcolor{blue}{target viewpoints}.\label{fig:qual_largeNVS_1}]{
    \includegraphics[width=1.0\textwidth]{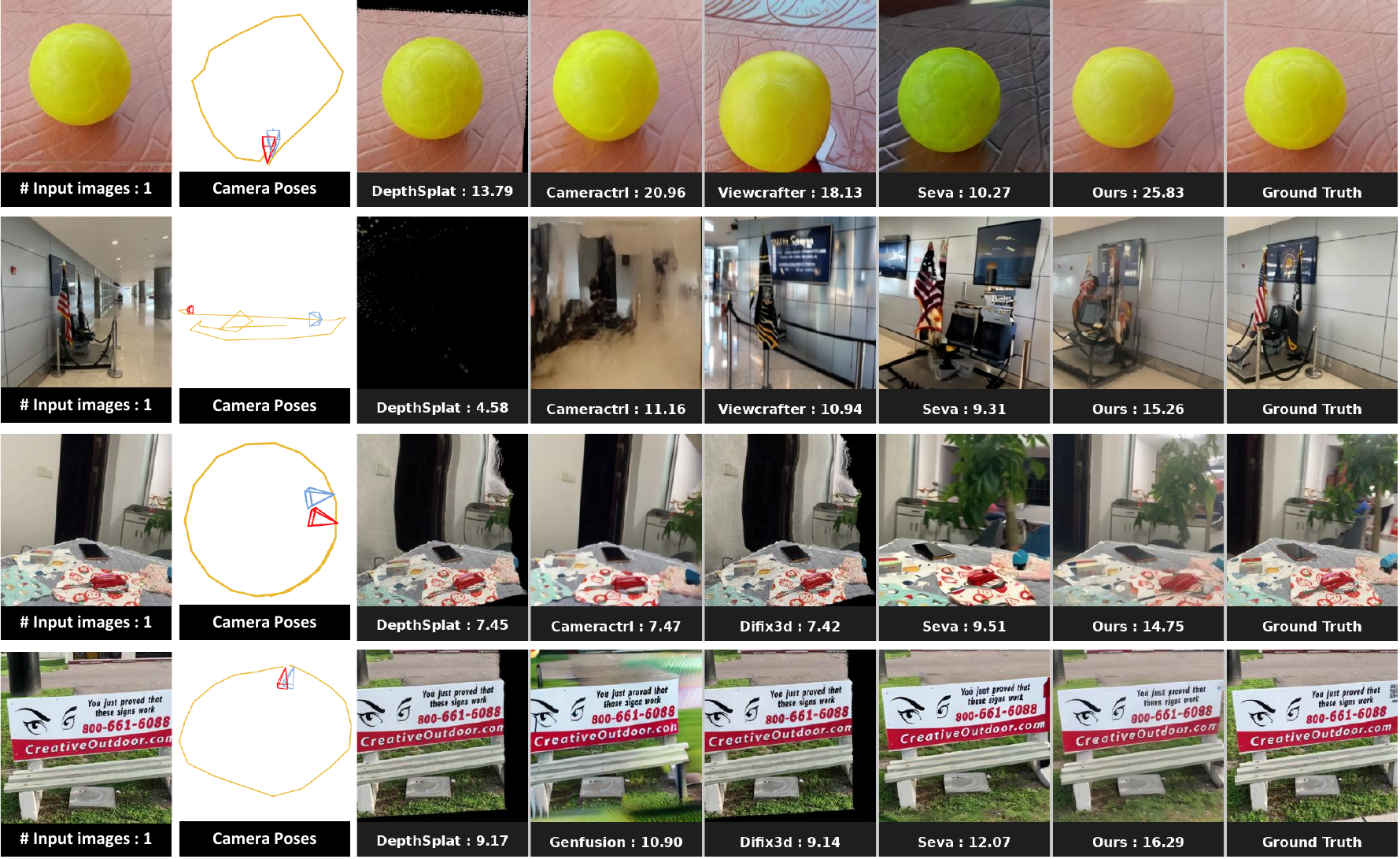}
}\\ 
\vspace{1mm}
\subfloat[\textbf{Qualitative results.} Given sparse \textcolor{red}{reference-view} images, novel-view images are synthesized at \textcolor{blue}{target viewpoints}.\label{fig:qual_largeNVS_2}]{
\vspace{1mm}
    \includegraphics[width=1.0\textwidth]{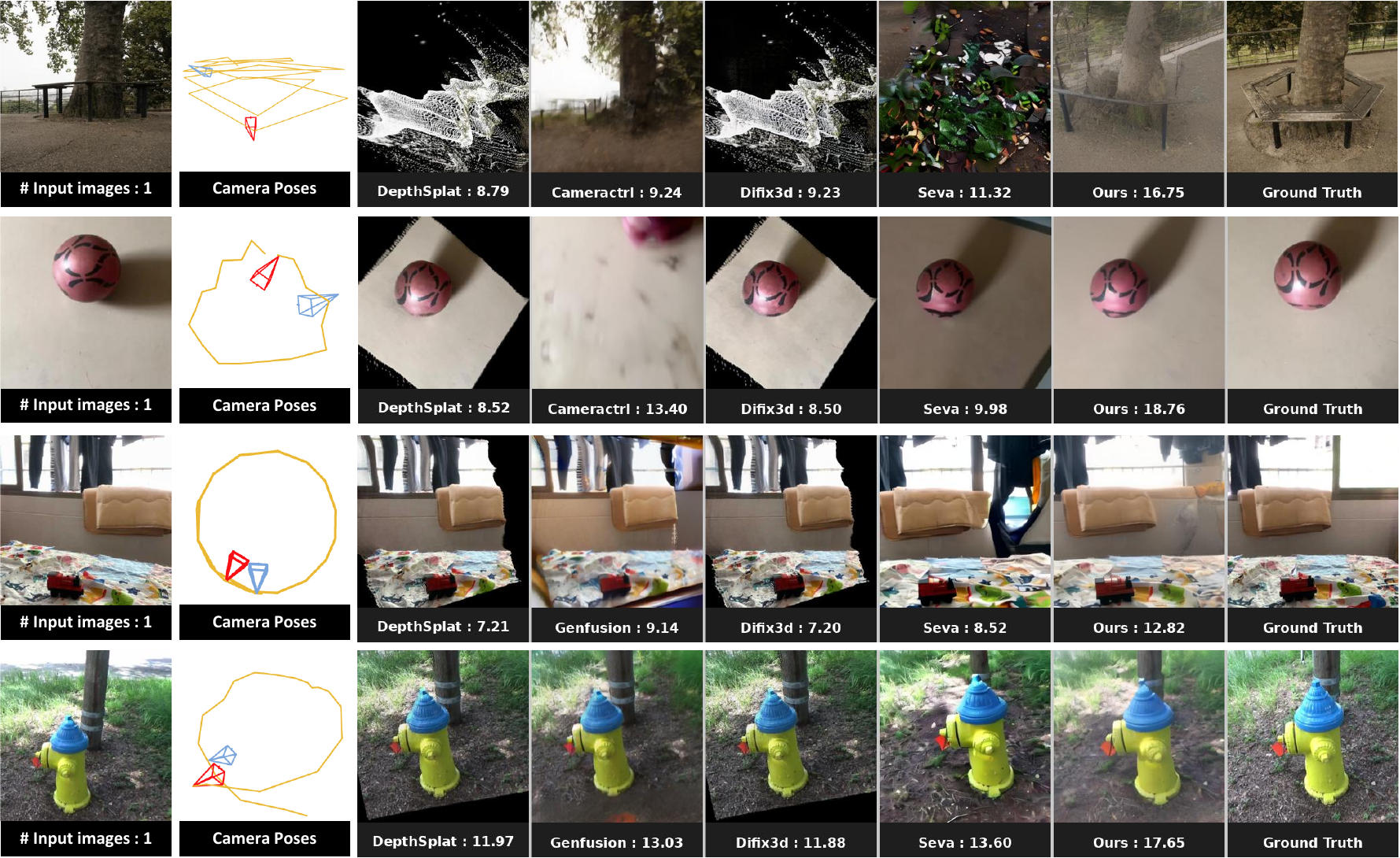}
}\\
\vspace{2mm}
\caption{\textbf{Qualitative results of large-viewpoint NVS.}}
\label{fig:supple_largeNVS_sample1}
\vspace{-4mm}
\end{figure}

%% file: Supplementary_Material/figure/largenvs_2.tex
\begin{figure}[t]
\centering
\subfloat[\textbf{Qualitative results.} Given sparse \textcolor{red}{reference-view} images, novel-view images are synthesized at \textcolor{blue}{target viewpoints}.\label{fig:qual_largeNVS_3}]{
    \includegraphics[width=1.0\textwidth]{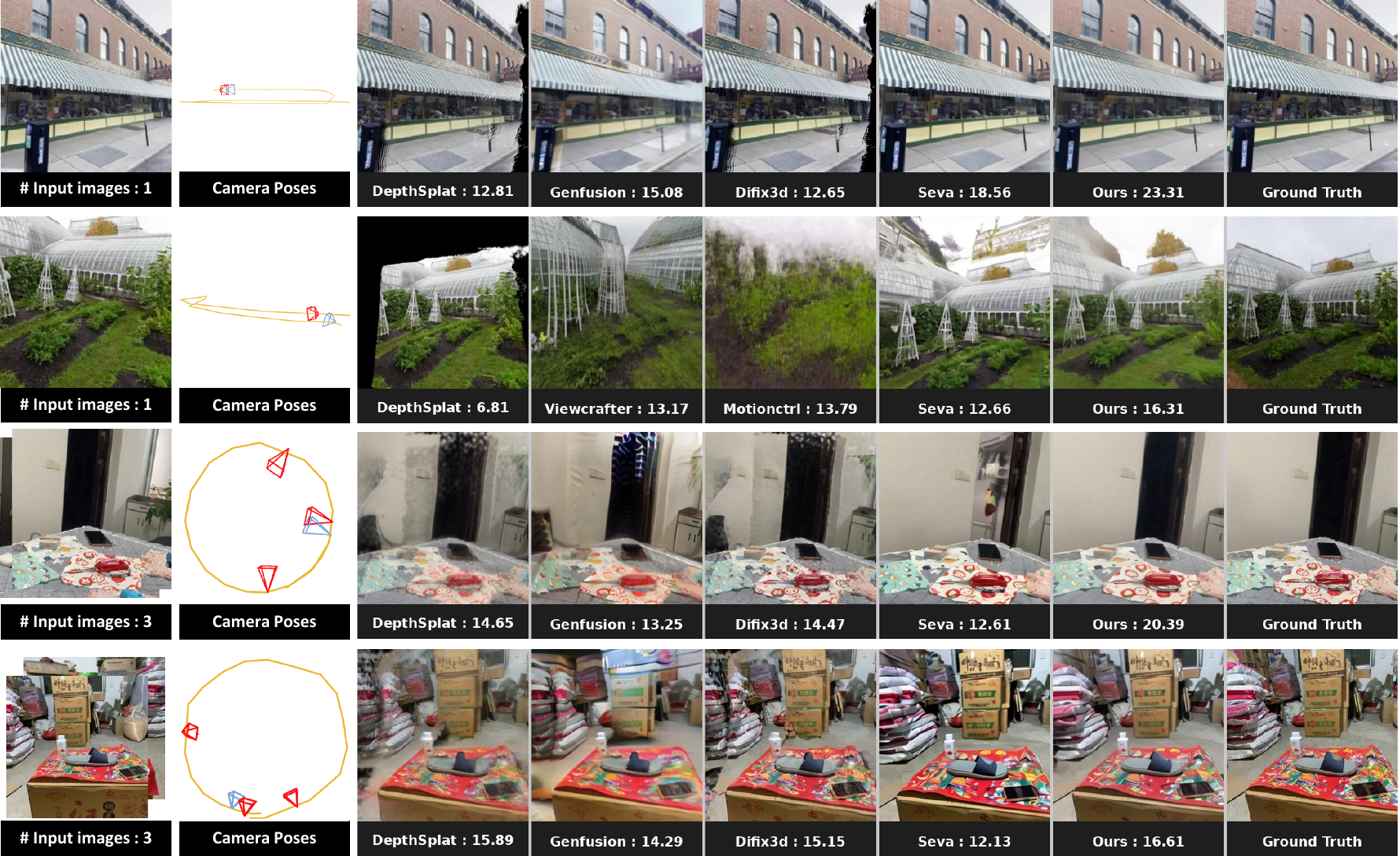}
}\\ 
\vspace{1mm}
\subfloat[\textbf{Qualitative results.} Given sparse \textcolor{red}{reference-view} images, novel-view images are synthesized at \textcolor{blue}{target viewpoints}.\label{fig:qual_largeNVS_4}]{
\vspace{1mm}
    \includegraphics[width=1.0\textwidth]{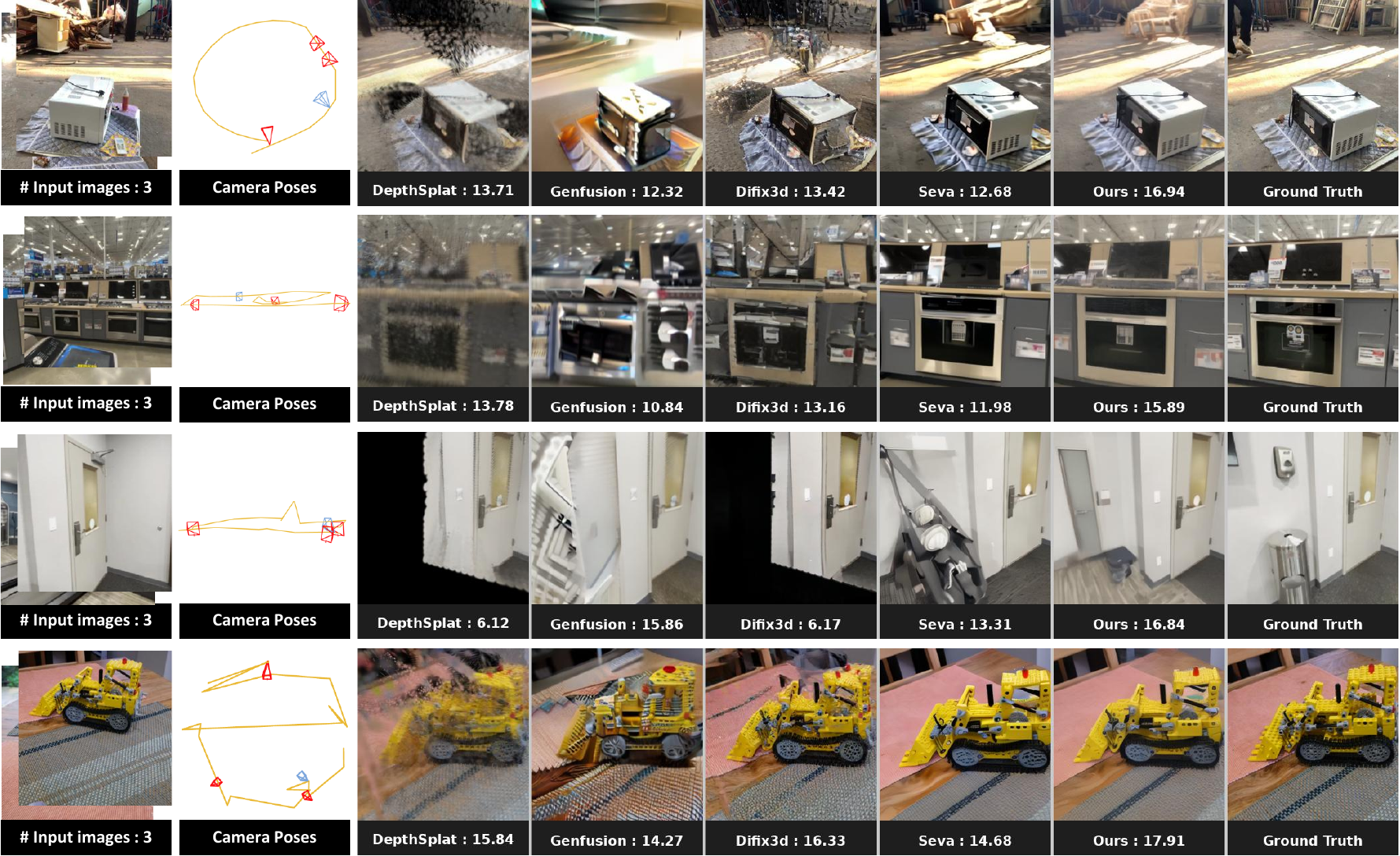}
}\\
\vspace{2mm}
\caption{\textbf{Qualitative results of large-viewpoint NVS.}}
\label{fig:supple_largeNVS_sample2}
\vspace{-4mm}
\end{figure}

%% file: Supplementary_Material/figure/ablation_lrms.tex
\begin{figure}[t]
\centering
\subfloat[\textbf{GeoNVS with Naïve Fusion.} Naïve fusion heavily relies on the geometry prior quality and fails when the prior is corrupted.\label{fig:qual_ablation_lrm_1}]{
    \includegraphics[width=1.0\textwidth]{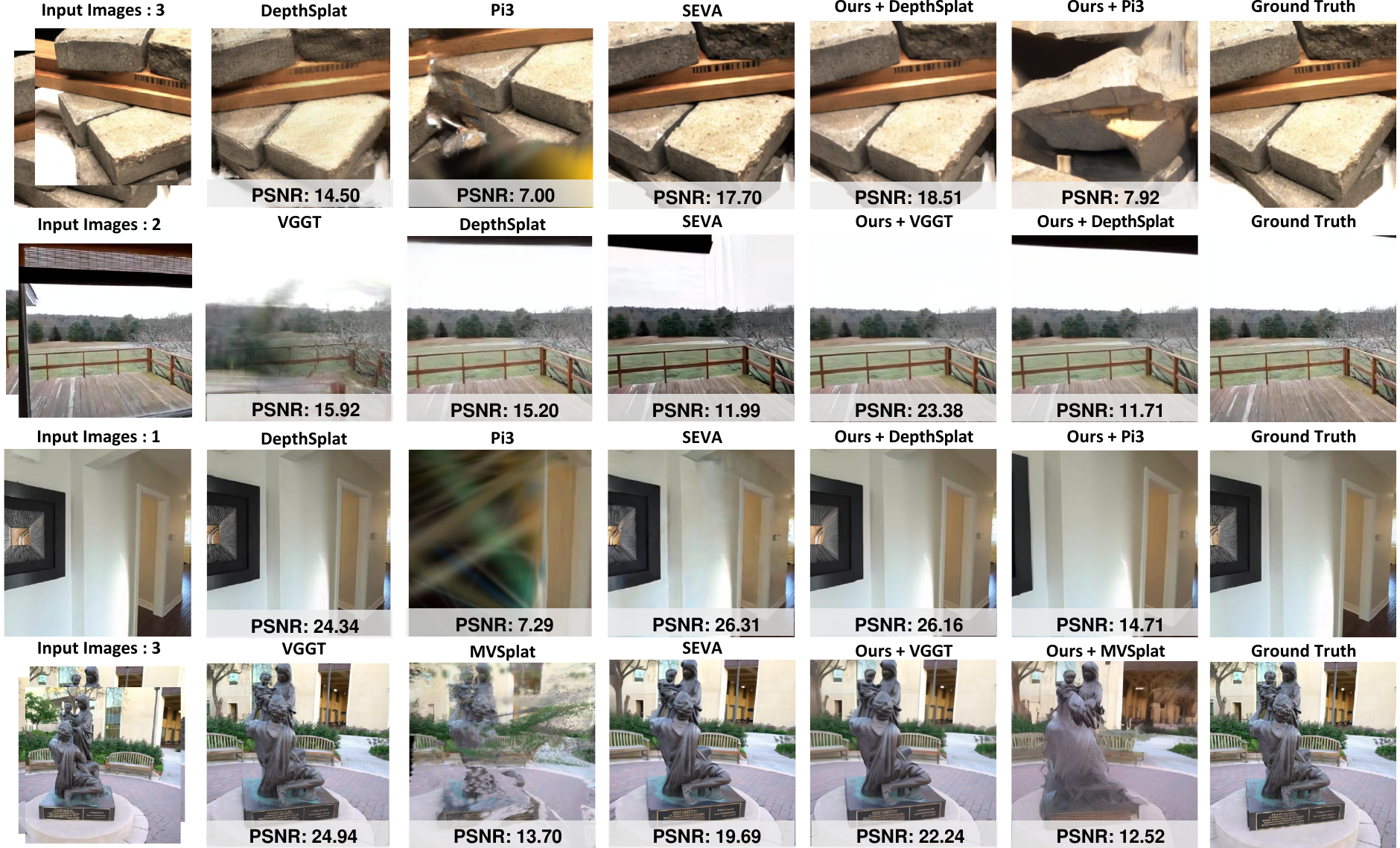}
}\\ 
\vspace{1mm}
\subfloat[\textbf{GeoNVS with Adaptive Fusion.} Adaptive Fusion substantially mitigates this sensitivity, achieving robust and consistent results regardless of the geometry prior quality.\label{fig:qual_ablation_lrm_2}]{
\vspace{1mm}
    \includegraphics[width=1.0\textwidth]{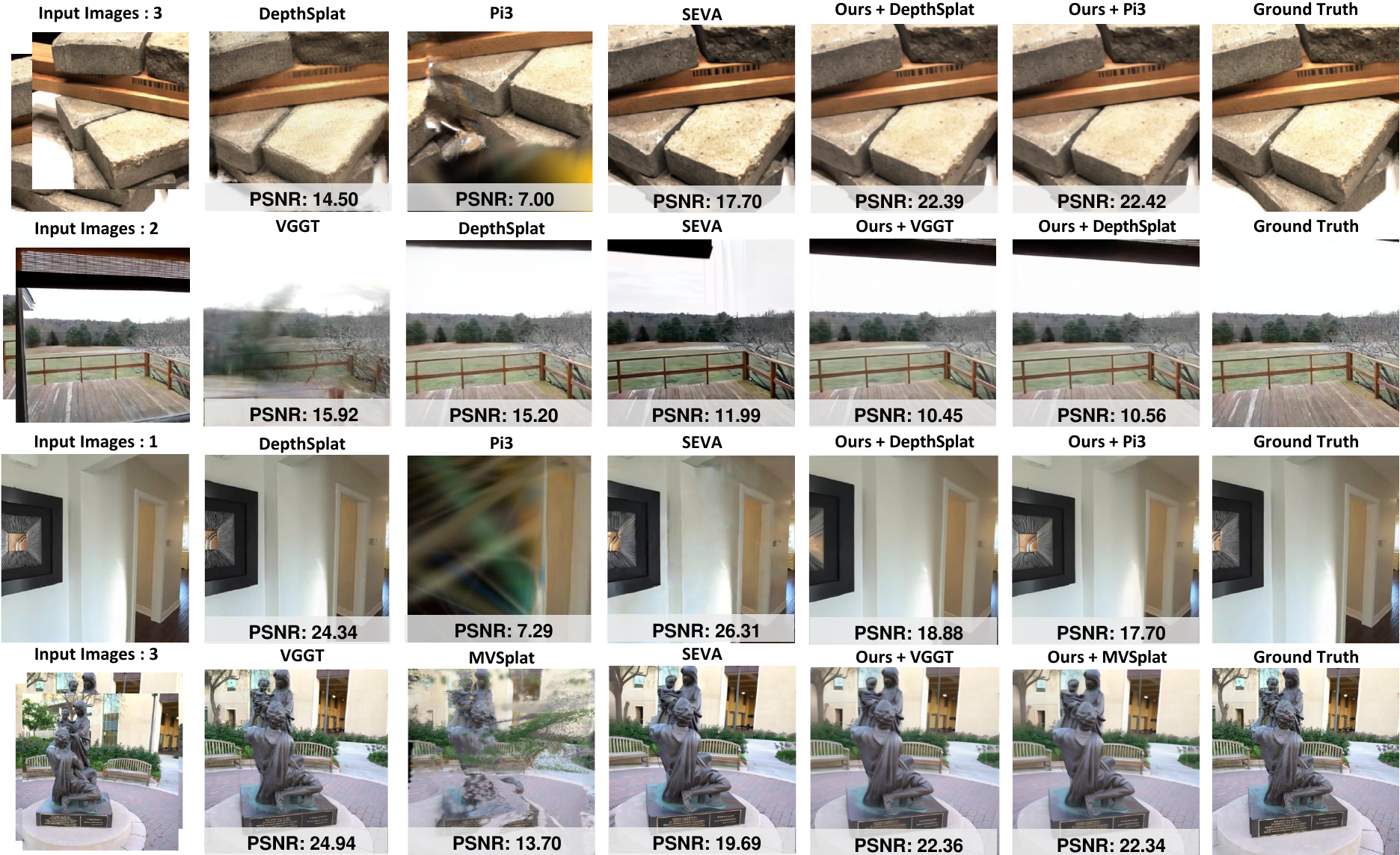}
}\\
\vspace{2mm}
\caption{\textbf{Ablation results with various geometry prior in~\cref{tab:ablation_lrms1}.}}
\label{fig:supple_ablation_lrm}
\vspace{-4mm}
\end{figure}

%% file: Supplementary_Material/figure/ablation_camctrl.tex
\begin{figure}[t]
\centering
\subfloat[\textbf{GeoNVS with CameraCtrl~\cite{he2025cameractrl}.} Given sparse \textcolor{red}{reference-view} images, novel-view images are synthesized at \textcolor{blue}{target viewpoints}. \label{fig:qual_ablation_camctrl_1}]{
    \includegraphics[width=1.0\textwidth]{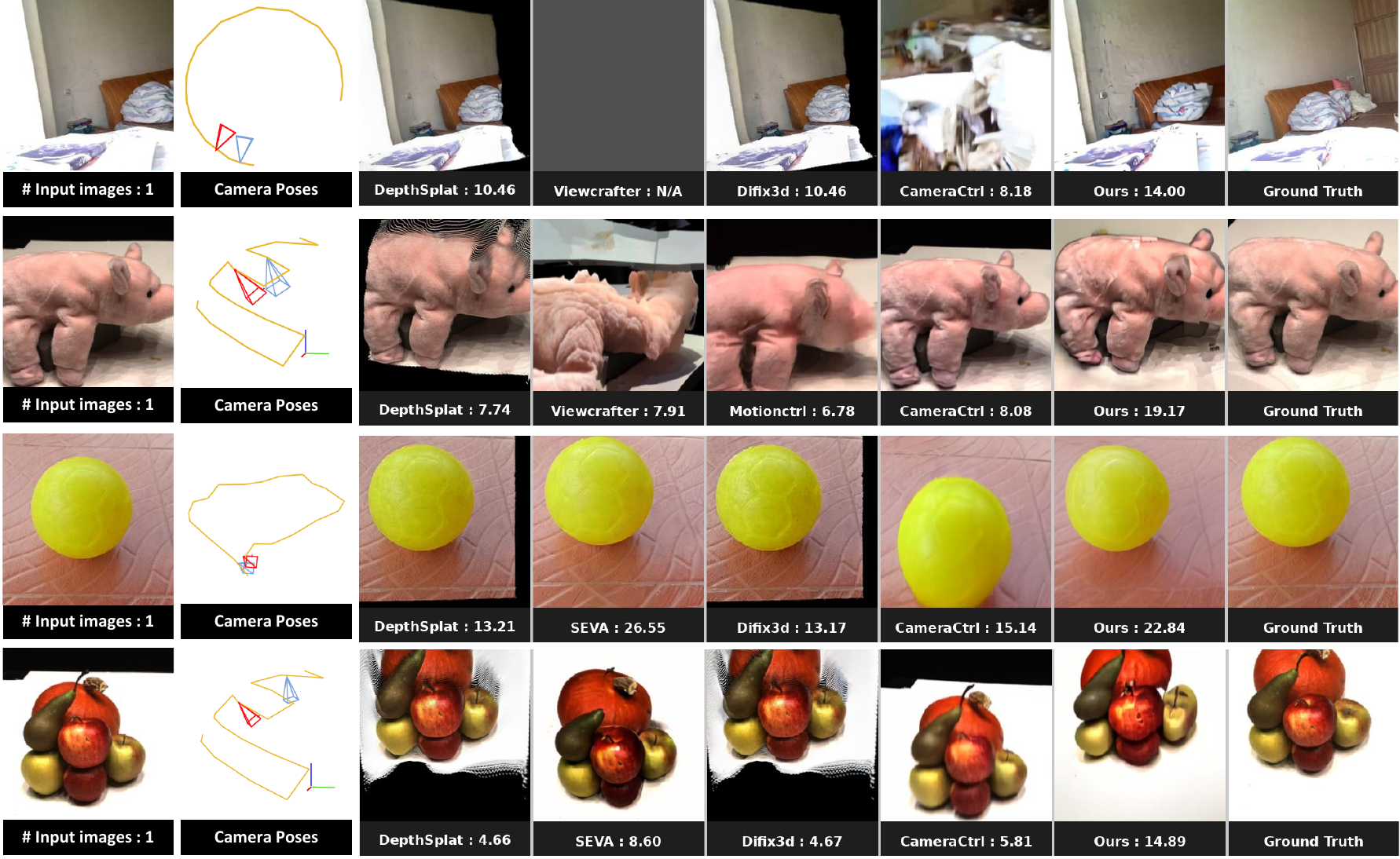}
}\\ 
\vspace{1mm}
\subfloat[\textbf{GeoNVS with CameraCtrl~\cite{he2025cameractrl}.} Given sparse \textcolor{red}{reference-view} images, novel-view images are synthesized at \textcolor{blue}{target viewpoints}.\label{fig:qual_ablation_camctrl_2}]{
\vspace{1mm}
    \includegraphics[width=1.0\textwidth]{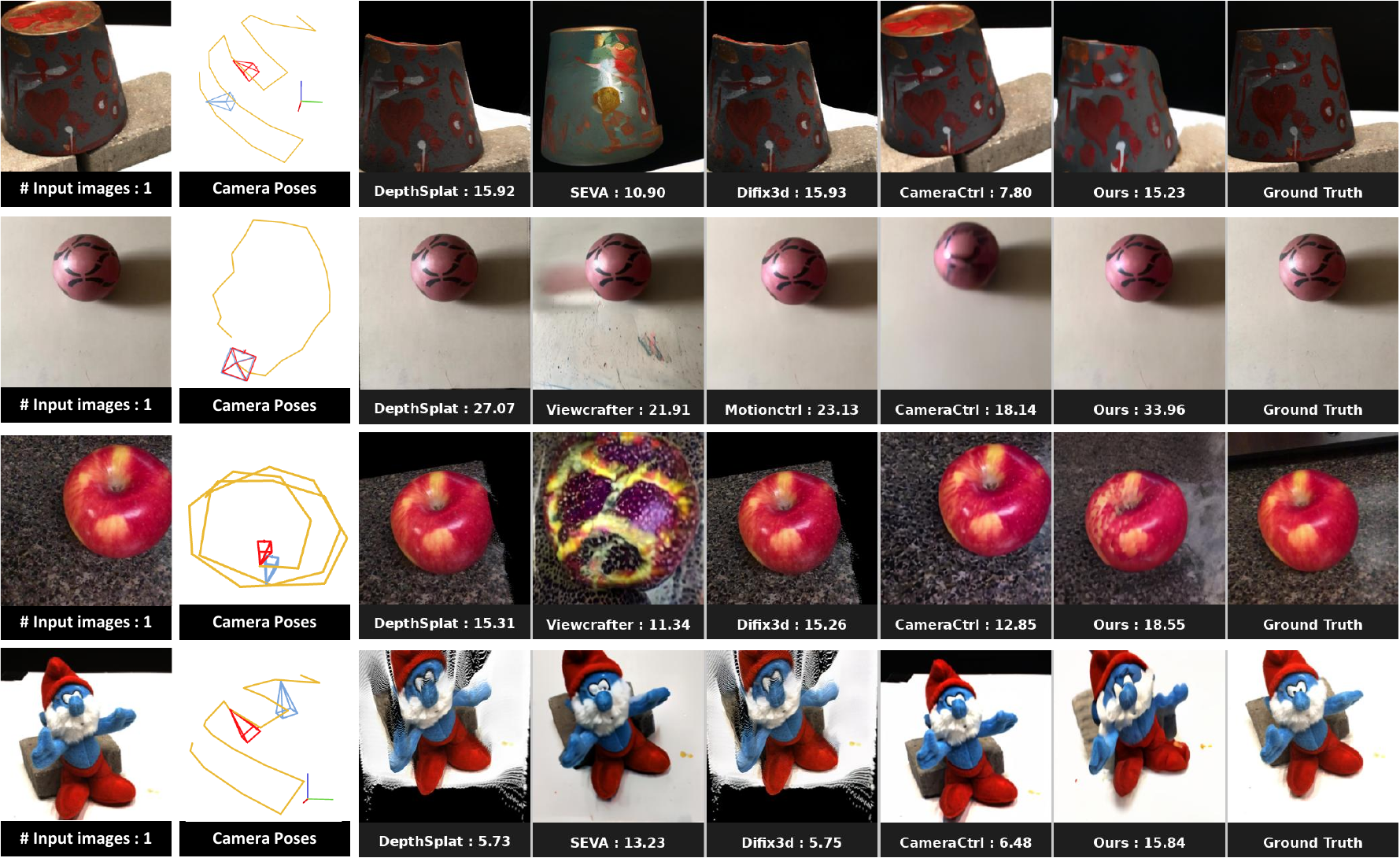}
}\\
\vspace{2mm}
\caption{\textbf{Ablation results of GeoNVS with CameraCtrl in~\cref{tab:ablation_lrms2}.}}
\label{fig:supple_ablation_camctrl}
\vspace{-4mm}
\end{figure}

%% file: Supplementary_Material/figure/trajnvs_1.tex
\begin{figure}[t]
\centering
\subfloat[\textbf{Qualitative results of GeoNVS.} 3D reconstruction and {\color{SkyBlue} camera trajectory} estimated from the generated video, showing the improved geometric fidelity of GeoNVS.\label{fig:qual_trajNVS_1}]{
    \includegraphics[width=1.0\textwidth]{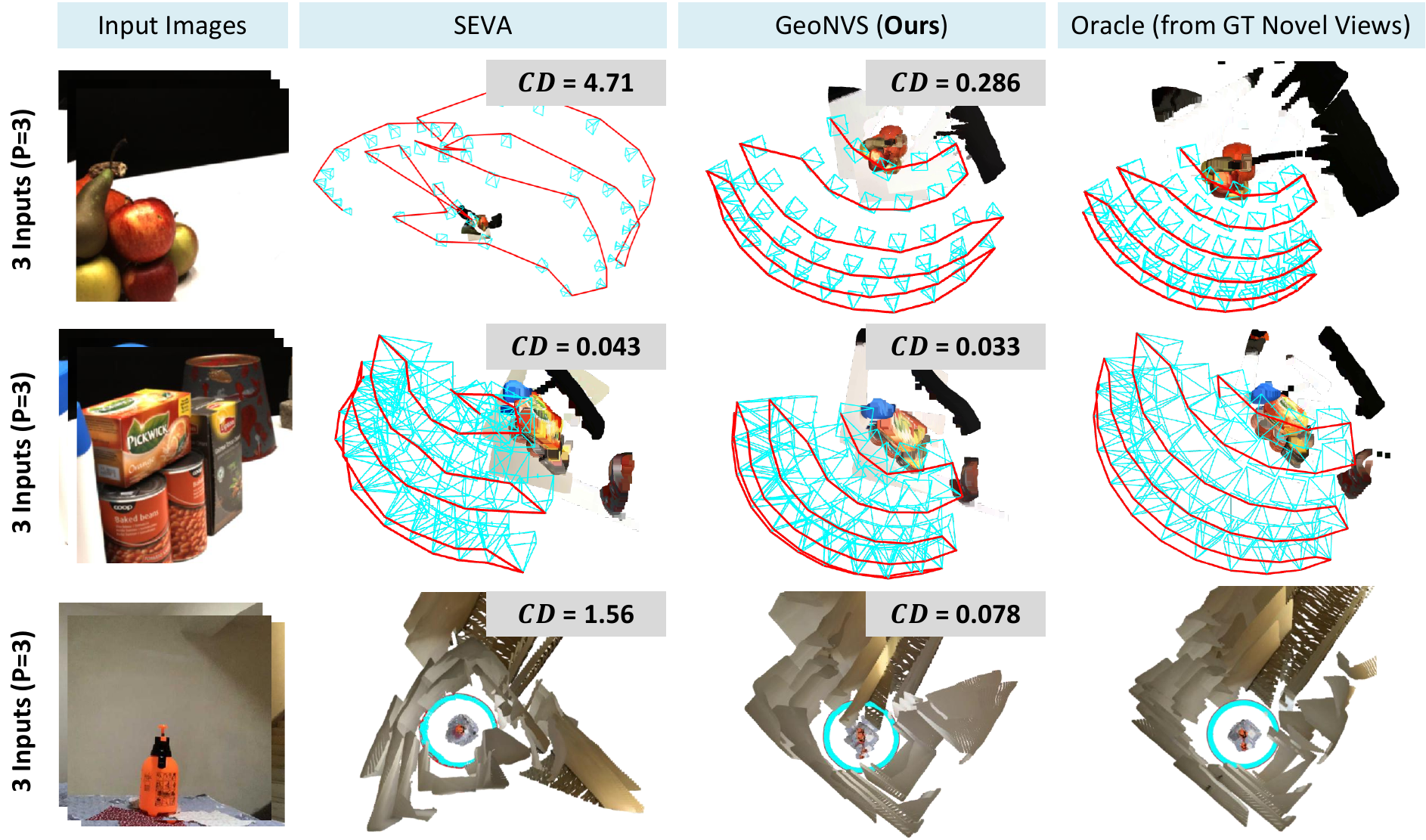}
}\\ 
\vspace{1mm}
\subfloat[\textbf{Qualitative results of GeoNVS.} 3D reconstruction and {\color{SkyBlue} camera trajectory} estimated from the generated video, showing the improved geometric fidelity of GeoNVS.\label{fig:qual_trajNVS_2}]{
\vspace{1mm}
    \includegraphics[width=1.0\textwidth]{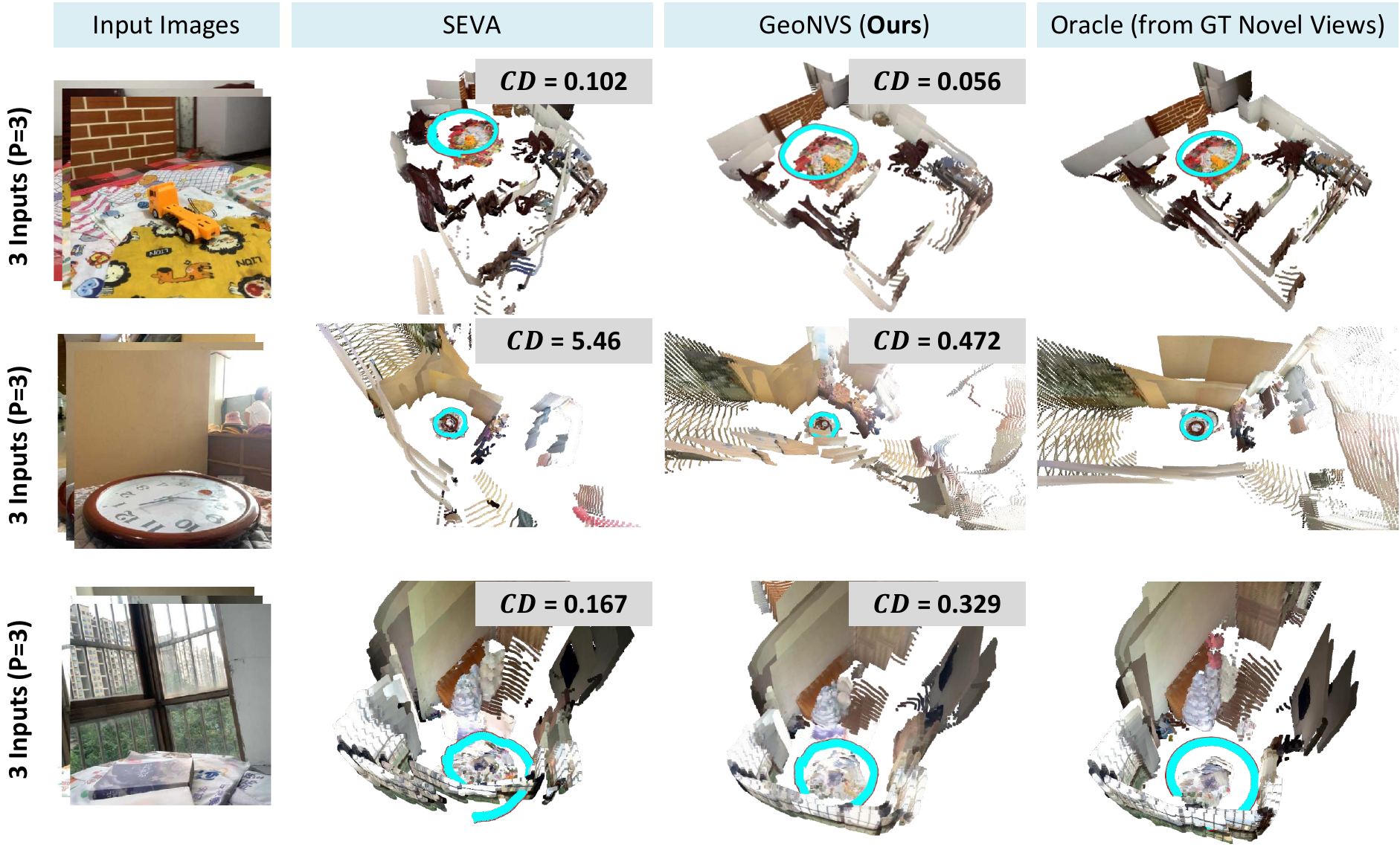}
}\\
\vspace{2mm}
\caption{\textbf{Qualitative results of trajectory NVS.}}
\label{fig:supple_trajNVS_sample1}
\vspace{-4mm}
\end{figure}

%% file: Supplementary_Material/figure/trajnvs_2.tex
\begin{figure}[t]
\centering
\subfloat[\textbf{Qualitative results of GeoNVS.} 3D reconstruction and {\color{SkyBlue} camera trajectory} estimated from the generated video, showing the improved geometric fidelity of GeoNVS.\label{fig:qual_trajNVS_3}]{
    \includegraphics[width=1.0\textwidth]{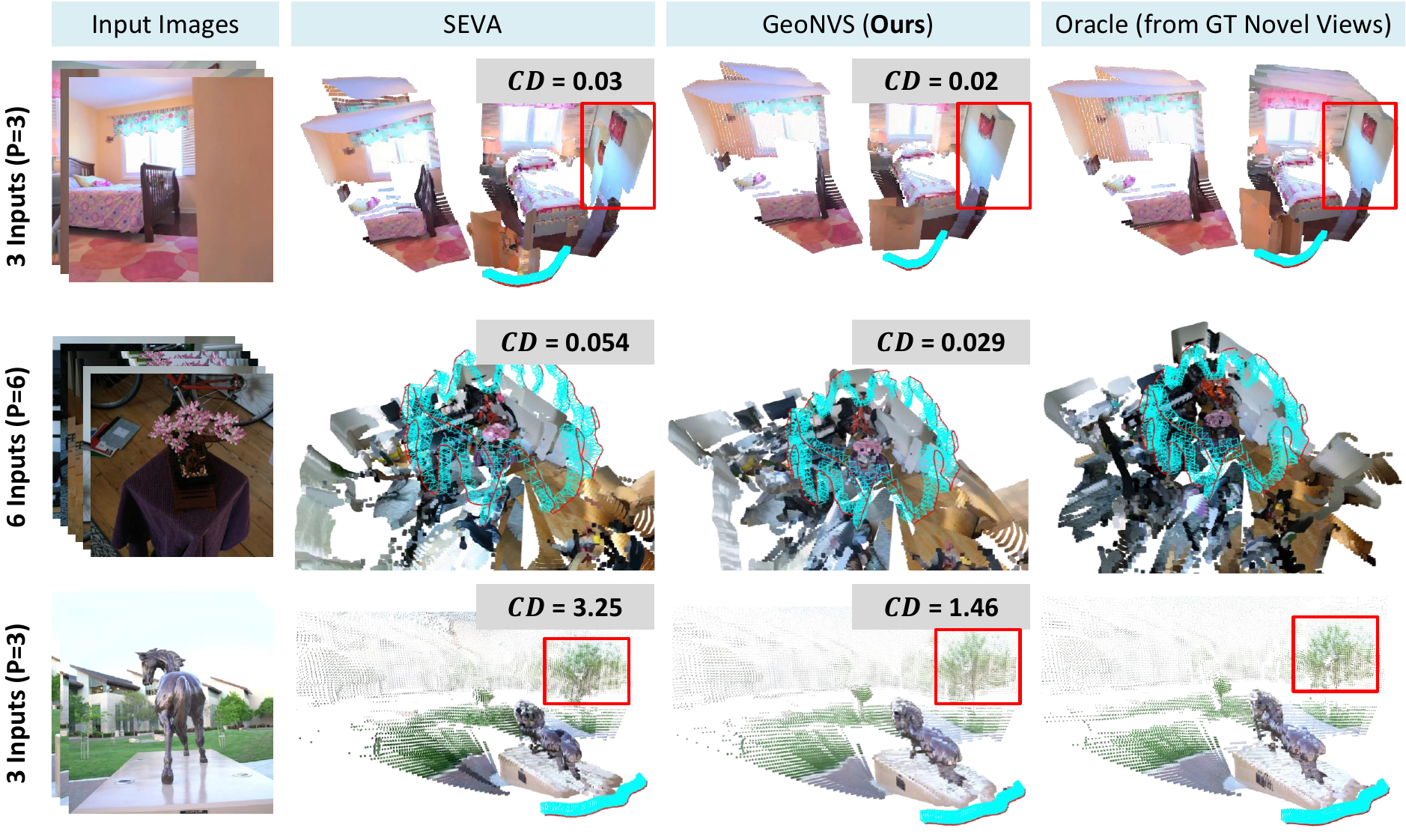}
}\\ 
\vspace{1mm}
\subfloat[\textbf{Qualitative results of GeoNVS.} 3D reconstruction and {\color{SkyBlue} camera trajectory} estimated from the generated video, showing the improved geometric fidelity of GeoNVS.\label{fig:qual_trajNVS_4}]{
\vspace{1mm}
    \includegraphics[width=1.0\textwidth]{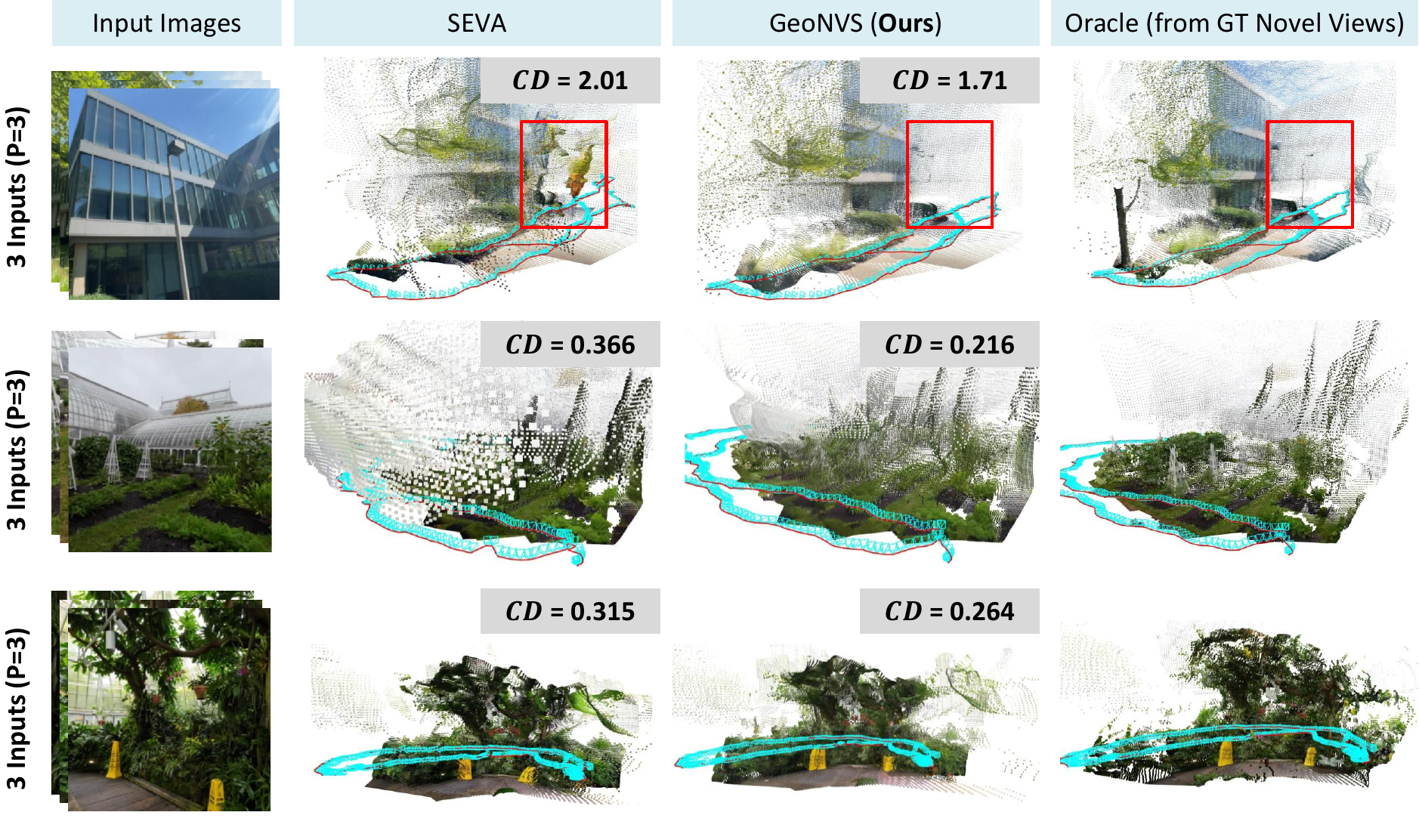}
}\\
\vspace{2mm}
\caption{\textbf{Qualitative results of trajectory NVS.}}
\label{fig:supple_trajNVS_sample2}
\vspace{-4mm}
\end{figure}

%% file: Supplementary_Material/figure/ablation_trajnvs.tex
\begin{figure}[t]
\centering
\subfloat[\textbf{GeoNVS compared to input-level injection.} 3D reconstruction and {\color{SkyBlue} camera trajectory} estimated from the generated video.\label{fig:qual_ablation_trajnvs_1}]{
    \includegraphics[width=1.0\textwidth]{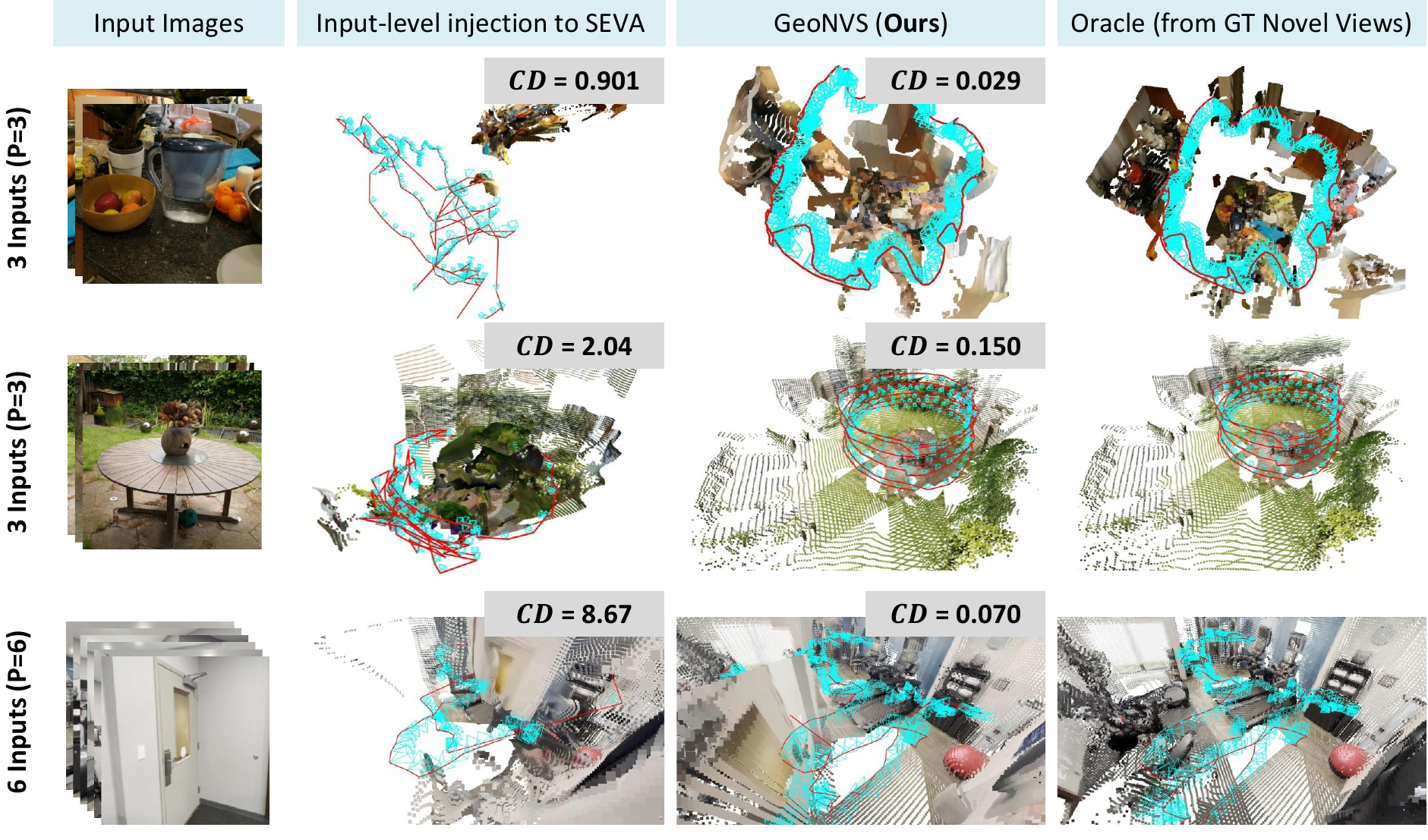}
}\\ 
\vspace{1mm}
\subfloat[\textbf{GeoNVS compared to difix3d~\cite{wu2025difix3d+}.} 3D reconstruction and {\color{SkyBlue} camera trajectory} estimated from the generated video.\label{fig:qual_ablation_trajnvs_2}]{
\vspace{1mm}
    \includegraphics[width=1.0\textwidth]{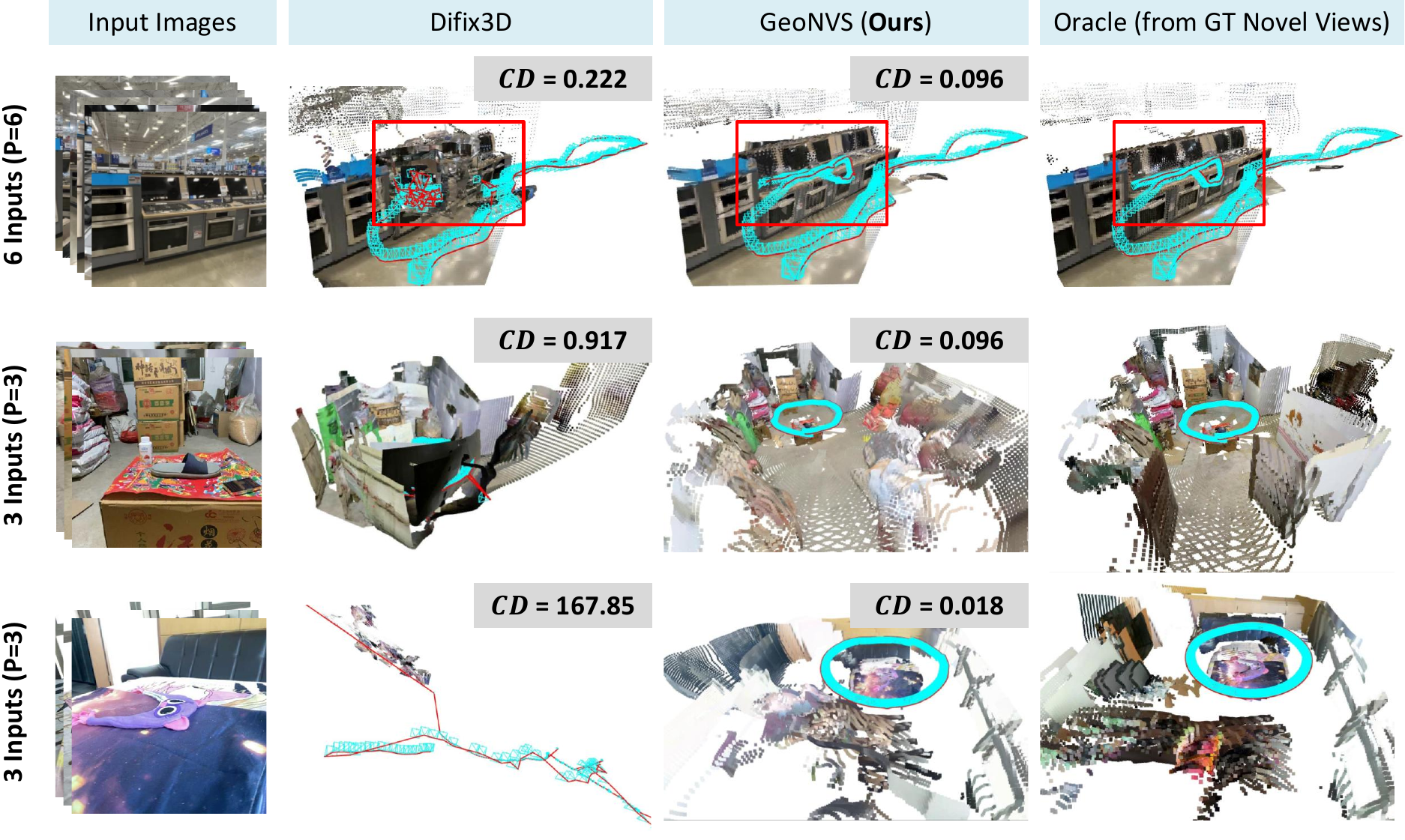}
}\\
\vspace{2mm}
\caption{\textbf{Qualitative results compared to input-level methods in~\cref{tab:geometric_consistency}.}}
\label{fig:supple_ablation_trajnvs}
\vspace{-4mm}
\end{figure}

%% file: Supplementary_Material/figure/limitations.tex
\begin{figure}[t]
\centering
\includegraphics[width=0.97\linewidth]{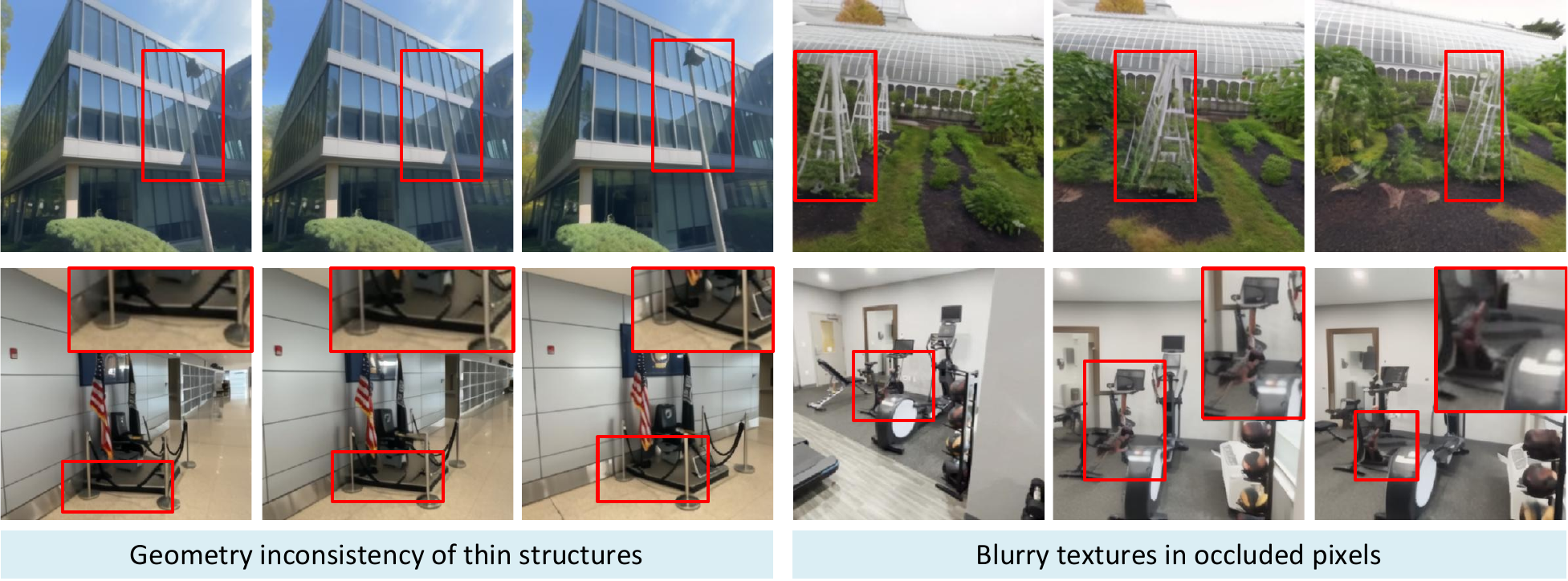}
\vspace{-2mm}
\caption{
\textbf{Limitations and Failure Cases of GeoNVS.} GeoNVS exhibits 
geometry inconsistency in thin structures (\textit{left}), where sparse 
Gaussian representations struggle to capture fine-grained geometry, and 
blurry textures in occluded regions (\textit{right}), where the geometry 
prior provides insufficient guidance for unobserved areas.}
\vspace{-5mm}
\label{fig:failure_cases}
\end{figure}